\documentclass[twoside,11pt]{article}
\usepackage{jair, theapa, rawfonts}
\usepackage{xcolor}
\usepackage{booktabs}
\usepackage{graphicx}
\usepackage{amssymb}
\usepackage{amsmath}
\usepackage{algorithm}
\usepackage{algorithmic}
\usepackage{subfig}
\usepackage{url}
\usepackage[title]{appendix}

\jairheading{75}{2022}{213-260}{02/2022}{09/2022}
\ShortHeadings{On Efficient Reinforcement Learning for Full-length Game of StarCraft II}
{Liu, Pang, Meng, Wang, Yu, \& Lu}
\firstpageno{213}

\begin{document}

\title{On Efficient Reinforcement Learning for Full-length \\ Game of StarCraft II}


\author{\name Ruo-Ze Liu \email liuruoze@163.com \\
       \name Zhen-Jia Pang \email pangzj@lamda.nju.edu.cn \\
       \name Zhou-Yu Meng \email misskanagi@gmail.com \\
       \name Wenhai Wang \email wangwenhai362@163.com \\
       \name Yang Yu \email yuy@nju.edu.cn \\
       \name Tong Lu \email lutong@nju.edu.cn \\
	   \addr \textit{(Corresponding author)} \\
       \addr National Key Laboratory for Novel Software Technology, \\
       Nanjing University, Nanjing 210023, China}


\maketitle

\begin{abstract}
	StarCraft II (SC2) poses a grand challenge for reinforcement learning (RL), of which the main difficulties include huge state space, varying action space, and a long time horizon. In this work, we investigate a set of RL techniques for the full-length game of StarCraft II. We investigate a hierarchical RL approach, where the hierarchy involves two. One is the extracted macro-actions from experts' demonstration trajectories to reduce the action space in an order of magnitude. The other is a hierarchical architecture of neural networks, which is modular and facilitates scale. We investigate a curriculum transfer training procedure that trains the agent from the simplest level to the hardest level. We train the agent on a single machine with 4 GPUs and 48 CPU threads. On a 64x64 map and using restrictive units, we achieve a win rate of 99\% against the difficulty level-1 built-in AI. Through the curriculum transfer learning algorithm and a mixture of combat models, we achieve a 93\% win rate against the most difficult non-cheating level built-in AI (level-7). In this extended version of the paper, we improve our architecture to train the agent against the most difficult cheating level AIs (level-8, level-9, and level-10). We also test our method on different maps to evaluate the extensibility of our approach. By a final 3-layer hierarchical architecture and applying significant tricks to train SC2 agents, we increase the win rate against the level-8, level-9, and level-10 to 96\%, 97\%, and 94\%, respectively. Our codes and models are all open-sourced now at~\url{https://github.com/liuruoze/HierNet-SC2}.
	
	To provide a baseline referring the AlphaStar for our work as well as the research and open-source community, we reproduce a scaled-down version of it, mini-AlphaStar (mAS). The latest version of mAS is 1.07, which can be trained using supervised learning and reinforcement learning on the raw action space which has 564 actions. It is designed to run training on a single common machine, by making the hyper-parameters adjustable and some settings simplified. We then can compare our work with mAS using the same computing resources and training time. By experiment results, we show that our method is more effective when using limited resources. The inference and training codes of mini-AlphaStar are all open-sourced at~\url{https://github.com/liuruoze/mini-AlphaStar}. We hope our study could shed some light on the future research of efficient reinforcement learning on SC2 and other large-scale games.

\end{abstract}

\section{Introduction}
\label{Introduction}
In recent years, reinforcement learning~\cite{sutton1998introduction} (RL) has developed rapidly in many domains. The combination of deep neural networks (DNN) and reinforcement learning emerges deep reinforcement learning (DRL), which has solved many games. The game of Go has been considered conquered after AlphaGo~\cite{silver2016mastering} and AlphaGo Zero~\cite{silver2017mastering}. Many Atari games are nearly solved using DQN~\cite{mnih2013playing} and follow-up methods. Various mechanical control problems, such as robotic arms~\cite{levine2016end} and self-driving vehicles~\cite{shalev2016safe}, have also made significant progress by DRL methods. However, RL algorithms are still challenging to use in large-scale problems. Moreover, agents cannot learn to handle problems as smartly and efficiently as humans. To further improve the ability of RL, more complex environments like strategic games have attracted interest from researchers.

StarCraft (SC) is one of such new environments for exploring RL's ability. Some pioneers have begun to explore RL on it, e.g., DeepMind~\cite{vinyals2017starcraft}, FAIR~\cite{tian2017elf}, and Alibaba~\cite{peng2017multiagent}. From the perspective of reinforcement learning, StarCraft is a challenging problem. Firstly, it is an imperfect information game. Players can only see a small region through a local camera. There is also a large fog of war on the map. Secondly, StarCraft's state space and action space are huge. The map of StarCraft is much larger than that of Go. There are hundreds of units or buildings of which each type may correspond to different actions, making the overall action space huge. Thirdly, the full-length game of StarCraft usually often lasts 10 minutes to 30 minutes. Each player needs to make thousands of decisions in this period. Finally, StarCraft is a multi-agent game, which may need cooperation and competition. The combination of these issues makes StarCraft a grand challenge for reinforcement learning.

Most previous agents on StarCraft are based on manual rules and scripts. Some works use imitation learning~\cite{behavioralCloning} to learn macro-management, like~\cite{justesen2017learning}. Some works use reinforcement learning to learn micro-management, such as~\cite{usunier2016episodic}, which handles specific problems like local battles in StarCraft. However, there are few works about full-length games on SC. StarCraft II (SC2), the successor of SC, is used as a new game for exploring RL. The learning platform based on SC2 is called SC2LE~\cite{vinyals2017starcraft} (StarCraft II Learning Environment). The full-length game result on SC2LE given by DeepMind~\cite{vinyals2017starcraft} shows that the state-of-the-art A3C algorithm~\cite{mnih2016asynchronous} did not achieve even one victory against the easiest level-1 built-in AI, which illustrates its difficulty. 

In this paper, we explore a set of RL techniques for the full-length game of SC2. In the method section, we present the hierarchical RL method we investigated in this paper, which uses several levels of abstractions to make the intractable large-scale RL problems easier to handle. An effective training algorithm tailed to the architecture is also investigated. After that, we give the experiments in the full-length SC2 game on a 64x64 map. Then, we discuss the impacts of different architectures, reward design, and settings of curriculum learning. Experimental results achieved in several difficult levels of full-length games on SC2LE illustrate the effectiveness of our method.

Note that a shorter conference version of this manuscript appeared on~\cite{pang2019starcraft}, in which the challenge of training against cheating level AIs (level-8, level-9, and level-10) was not addressed. In this manuscript, we train a new agent on these cheating levels using three improvements: First, we expand the two-layer hierarchical network to a three-layer one. Second, we use the win/lose outcome as a reward instead of the designed reward functions. Third, we train from scratch against the cheating level built-in AIs. We find the latter two improvements are important for stable training due to the mutual impacts of modules in the architecture and the randomness of the cheating level AIs. We also train on two other different maps, \textit{Flat64} and \textit{Simple96}, to test the extensibility of our method. Finally, by a final 3-layer hierarchical architecture and applying significant tricks to train SC2 agents, we increase the final win rate against the cheating level built-in AIs (level-8, level-9, and level-10) to 96\%, 97\%, and 94\%, respectively.

Several other methods using RL to handle the full-length game of SC2 have been published after our conference version. We give comparisons with them in this manuscript. We discuss and compare our work with TStarBots~\cite{Sun2018tsbot}, showing we use less prior knowledge and fewer computing resources than TStarBots while achieving similar performance. A recent state-of-the-art agent, the AlphaStar, has already beat professional players, which achieves the Grand-Master level on SC2. However, its RL setting is simplified compared to our work, and its success also lies in substantial computing resources and much human knowledge. AlphaStar consists of 4 main parts: deep neural network, supervised learning, reinforcement learning, and multi-agent league mechanism. Its network contains three encoders, one core part, and six heads. We will give a detailed analysis of why the A3C algorithm fails in the first experiments in SC2LE while a similar algorithm~\cite{Espeholt2018vtace} succeeds in AlphaStar. 

AlphaStar's parameter number is about 40M, while ours is about 50k. AlphaStar needs 12000 CPU cores, 384 TPUs, 971K replays, and 44 days to train its model, which is very costly. Due to these differences, directly comparing our results with theirs is unfair. Therefore, in this manuscript, we reproduce a scaled-down version of it, which is called mini-AlphaStar (mAS). The mAS follows AlphaStar architecture and is trainable on one common machine by making the hyper-parameters smaller and settings simpler. We then compare our hierarchical approach with mAS using the same computing resources and training time. We show that our work is more efficient and more effective when using limited resources. To our best knowledge, mAS is the first minimal reproduction of the AlphaStar, designed to run training on one common machine. The mAS may provide a useful baseline on future research and the open-source community. The main contributions of this paper are as follows: 
\begin{itemize}
    \item We investigate a hierarchical architecture that makes large-scale SC2 problems easier to handle. We also present an effective training algorithm for this architecture.
    \item We study in detail the impact of different training settings on our architecture. Experiment results on SC2LE show that our method achieves comparative results.
    \item In the extended version, we train an effective agent on cheating difficulty levels and two other maps through an improved hierarchical architecture.
    \item We give a detailed discussion of the differences between our works and AlphaStar. To be compared with it, we reproduce a scaled-down version of AlphaStar, called mini-AlphaStar. 
	\item We present a complete comparison between our work and the mini-AlphaStar. Experiment results show that our hierarchical method is more effective when using limited resources.
\end{itemize}

The remaining parts of the paper are organized as follows. Section~\ref{section: Background} introduces background knowledge and related works of our method. We present our method in Section~\ref{section: Methodology} and experiments in Section~\ref{section: Experiments}. As the extensions to our conference paper, introduction and comparison experiments of mini-AlphaStar are presented in Section~\ref{section: Introduction of mini-AlphaStar}, and the additional experiments against cheating level built-in AIs are provided in Section~\ref{section: Additional Experiments}. Section~\ref{section: Conclusion} concludes our paper.

\section{Background} \label{section: Background}
In this section, we first give the definitions of reinforcement learning. Then we discuss previous hierarchical RL methods. After that, previous studies on StarCraft are analyzed. Finally, we present the discussion of AlphaStar.

\subsection{Reinforcement Learning}
Consider a finite-horizon Markov Decision Process (MDP), which can be specified as a 6-tuple:
\begin{equation}
   M=\langle S,A,P(.),R(.),\gamma, T \rangle
\end{equation}
$ S $ is the state space, and $ s \in S $ is one state in the state space. $ A $ is the action space, and $ a \in A $ is an action. $ P(.)=\Pr(s' | s, a) $ represents probability distribution of the next state $ s' $ over $ S $ when an agent chooses an action $ a $ in a state $ s $. $ R(.)=R(s, a) $ represents the instant reward gained from the environment. $ \gamma $ is the reward discount factor. $ T $ is the max length of the time horizon.

Let policy $ \pi $ be the mapping or distribution from $ S $ to $ A $. If $ \pi(s) $ is a mapping, when it is in state $ s $, the agent selects the action $ a=\pi(s) $. This $ \pi(s) $ is called a deterministic policy. On the contrary, if $ \pi(s) $ is a distribution, when the agent is in state $ s $, the agent samples the action $ a\sim\pi(s) $. Under that situation, $ \pi(s) $ is called a stochastic policy. If a policy does not change over time, it is a stationary policy. Otherwise, it is a non-stationary policy. Assuming the agent, which uses policy $ \pi $ and starts from state $ s_0 $, chooses action $ a_0 $ and gains reward $ r_0=R(s_0,a_0) $. It then transforms to the next state $ s_1 $ according to the distribution $ \Pr( .| s, a) $. The process is repeated and generates a sequence $ \tau $ as below:
\begin{equation}
    \tau = s_0, a_0, r_0, s_1, a_1, r_1,\ldots
    \label{eqn:d_i}
\end{equation}
Suppose there is a state $ s_{end} $, in which an agent will stop. The process from $ s_0 $ to $ s_{end} $ is called one episode. The sequence $\tau$ in the episode is called a trajectory of the agent. For the finite-horizon problem, the agent also ends the exploration when the time step exceeds $ T $. Common RL algorithms require hundreds or thousands of episodes to learn a policy. In one episode, the discounted cumulative reward got by the agent is defined as:
\begin{equation}
    G = r_0 + \gamma r_1 + \gamma^2 r_2 + \cdots
    \label{eqn:r_i}
\end{equation}
$ G $ is called the return of this episode. A typical RL algorithm aims to find an optimal policy to maximize the expected return.
\begin{equation}
    \pi^* = \operatorname*{argmax}_{\pi}\mathbb{E}_{\pi}[\sum_{t=0}^{T} \gamma^t R(s_t,a_t)]
\end{equation}

\subsection{Hierarchical Reinforcement Learning}
When the dimension of the state space is huge, the states that need to be explored exhibits exponential growth, which is called the curse of dimensionality in RL. Hierarchical reinforcement learning (HRL) handles the curse of dimensionality by decomposing a complex problem into several sub-problems and solving each sub-problem one by one. There are some traditional HRL algorithms. For example, Option~\cite{sutton1999between} makes abstraction for actions. MaxQ~\cite{Dietterich1999maxq} splits up the problem by decomposition of the value function. ALISP~\cite{David2002alisp} provides a safe state abstraction method that maintains hierarchical optimality. Although these algorithms can better handle the curse of dimensionality problems, they mostly need to be manually defined, which is time-consuming and laborious. Nevertheless, another advantage of the HRL algorithm is that the resolution of time is reduced so that the credit assignment problem over a long time horizon can be better handled.

Other HRL algorithms have been proposed in recent years. Option-Critic~\cite{bacon2017option} is a method using the theorem of gradient descent to learn options and the corresponding policies simultaneously, which reduces the effort of manual designing options. However, the automatically learned options do not perform as well as non-hierarchical algorithms on certain tasks. FeUdalNetwork~\cite{vezhnevets2017feudal} designs a hierarchical architecture that includes a Manager module and a Worker module and proposes a gradient transfer strategy to learn parameters of the Manager and Worker in an end-to-end way. However, due to its complexity, its hyper-parameters are hard-to-tune. MLSH (Meta Learning Shared Hierarchies)~\cite{frans2017meta} proposes a hierarchical learning approach based on meta-learning, which enhances the learning ability to transfer to new tasks through sub-policies learned in multiple tasks. MLSH has achieved better results on some tasks than the PPO~\cite{schulman2017proximal} algorithm. However, because its setting is multi-tasking, it can not be straightforward to apply to our environment.

The hierarchical framework we propose in this paper consists of two levels of abstraction: the abstraction of the action space and the abstraction of the architecture. We use the method of data mining to mine the macro actions from the experts' trajectories. It saves the effort of manual design and ensures the effectiveness of the learned macro actions. We refer to the FeUdalNetwork and design the top layer controller and the lower layer executer in architecture design. For the sake of simplicity, we also draw on the ideas of MLSH (using different buffers for different layers and updating them separately) to design our training algorithm. Our framework can learn an effective policy more efficiently through these two abstractions. Another advantage of our hierarchical framework is that it is modular and thus easy to scale.

\subsection{StarCraft II}
Games are ideal environments for RL research. Reinforcement learning problems on real-time strategy (RTS) games are far more complicated than Go due to their complexity of states, diversity of actions, and a long time horizon. Traditionally, research on RTS games is based on searching and planning approaches~\cite{ontanon2013survey}. In recent years, RL algorithms have also been investigated in RTS. One of the most famous RTS research environments is StarCraft. Previous works on StarCraft mainly focused on local battles or part-length games, getting states directly from the game engine, and giving command actions to the game engine. For example, Usunier~\cite{usunier2016episodic} presented a heuristic reinforcement learning algorithm combining exploration in the space of policy and back-propagation. Peng~\cite{peng2017multiagent} introduced BiCNet based on multi-agent reinforcement learning combined with actor-critic. Although these methods have achieved good results, they are only effective for part-length games. ELF~\cite{tian2017elf} provided a framework for efficient training and a mini-RTS platform for RL research. ELF also gave a baseline of the A3C~\cite{mnih2016asynchronous} algorithm in a full-length game of mini-RTS. However, because the setting of mini-RTS is relatively simple, there is still a distance from the complexity of StarCraft. 

SC2LE~\cite{vinyals2017starcraft} is a new reinforcement learning environment based on StarCraft II (SC2). SC2 is the successor of StarCraft. In StarCraft, the game engine gives the location information for each unit. On the contrary, in SC2LE, the location information needs to be perceived from the image of the game screen. Meanwhile, to simulate the hand movements of humans, SC2LE changed the action space to the click actions of the mouse. These two changes significantly increase the difficulty of RL training. The benchmark result given by DeepMind shows that the A3C algorithm~\cite{mnih2016asynchronous} does not achieve even one victory against the easiest level-1. In addition to a full-length game, SC2LE also provides several mini-games for research. Zambaldi~\cite{zambaldi2018relational} proposed a relation-based RL algorithm, which achieved good results on these mini-games. However, its results on the full-length game are not reported. TStarBots~\cite{Sun2018tsbot} achieved good results on the full-length game of SC2, which is close to our work. However, our work uses less prior knowledge and fewer computing resources. We will discuss these in Section~\ref{subsection:Comparison with TStarBots}.

\subsection{AlphaStar}
Recently, AlphaStar, which can beat professional players, made big progress on the research of RL for SC2. However, we discovered that AlphaStar also has some shortcomings. E.g., it has abandoned the original settings of SC2LE, which is for simulating human actions. That is, AlphaStar doesn't use the ``human action space" (HAS) but instead the ``raw actions space" (RAS). RAS was originally the one that only the game engine could use. Besides, AlphaStar also gets more information about units from the game engine. These changes decrease the learning difficulties for RL by a large margin. We will discuss them in detail in Section~\ref{subsection:Comparison with AlphaStar}. 

We argue that AlphaStar has not perfectly solved the problem of SC2, for these 4 reasons: 1. AlphaStar uses too much human knowledge to build its agent, e.g., it uses many human replays to do supervised learning (SL) and uses replays as rewards to guide the exploration of the agents in RL training; 2. Due to reason 1, AlphaStar has not created any new tactics other than the human ones originally envisaged by the game community, which can be seen in the analysis in~\cite{liu2021rethinkAS}; 3. Like the training of Go~\cite{silver2016mastering,silver2017mastering}, AlphaStar's training needs a lot of computing resources. Moreover, it requires even more training resources~\cite{AlphaStarNature} than AlphaGo, which is not acceptable for small companies or research institutions. 4. AlphaStar is hard to be replicated. Until now, there are only a few reproductions of it that achieve similar results to theirs.

After AlphaStar, there are some follow-up works. The famous two are TStarBotsX~\cite{Han2020TX} and SCC~\cite{Wang2021SCC}. The former trains a high-level Zerg agent, using similar technologies as AlphaStar, but it added some rules to its training scheme and the training costs many machines. The latter provides a powerful Terran agent that can defeat two professional human players, but it also costs a lot of resources, and its codes are not open-sourced. There are now two open-source works using similar architectures as AlphaStar: DI-Star~\cite{opendilab2021distar} and SC2IL~\cite{metataro2021SC2IL}. The former uses the most similar architecture as AlphaStar, but its early version did not provide training codes. In contrast, since our first versions, we have provided all training codes. The latter provides training codes, but it only contains the supervised learning part. In contrast, our work contains both supervised learning and reinforcement learning parts meanwhile.

Note that this work share some similarities with our more recent work, TG~\cite{liu2021ThoughtGame}. The two ones both focus on efficient reinforcement learning on SC, and choose built-in AI as opponents. However, there are significant differences between the two. The TG concentrates on model-based RL and proposes using a hand-designed abstract model and transfer learning to train an agent. On the contrary, this paper's work focuses on hierarchal reinforcement learning and provides a novel hierarchal architecture to train the agent in an end-to-end manner. The hierarchical approach here has also achieved better results than TG's (e.g., the win rate against level-10 built-in AI is 94\% while TG's is 90\%). Moreover, unlike the method in TG, the hierarchical approach does not have the burden of hand designing an abstract model.

\section{Methodology} \label{section: Methodology}
This section introduces our hierarchical architecture and the generation of macro-actions first. Then the training algorithm of the architecture is given. Lastly, we discuss the reward design and the curriculum learning setting used in our method.

\subsection{Hierarchical Architecture} \label{subsection: Hierarchical Architecture}
StarCraft is a large-scale problem for RL, meaning it's hard to learn a single policy to win in the game. From the hierarchical perspective, we usually need to achieve multiple goals, e.g., earn our resources as much as possible, beat the enemy in the battle, and so on. We can split the big problem into several smaller ones. Each one can be seen as a single RL problem and handled by a sub-policy.

\begin{figure}[h]
	\begin{minipage}[t]{\linewidth}
		\centering
		\includegraphics[width=0.95\textwidth]{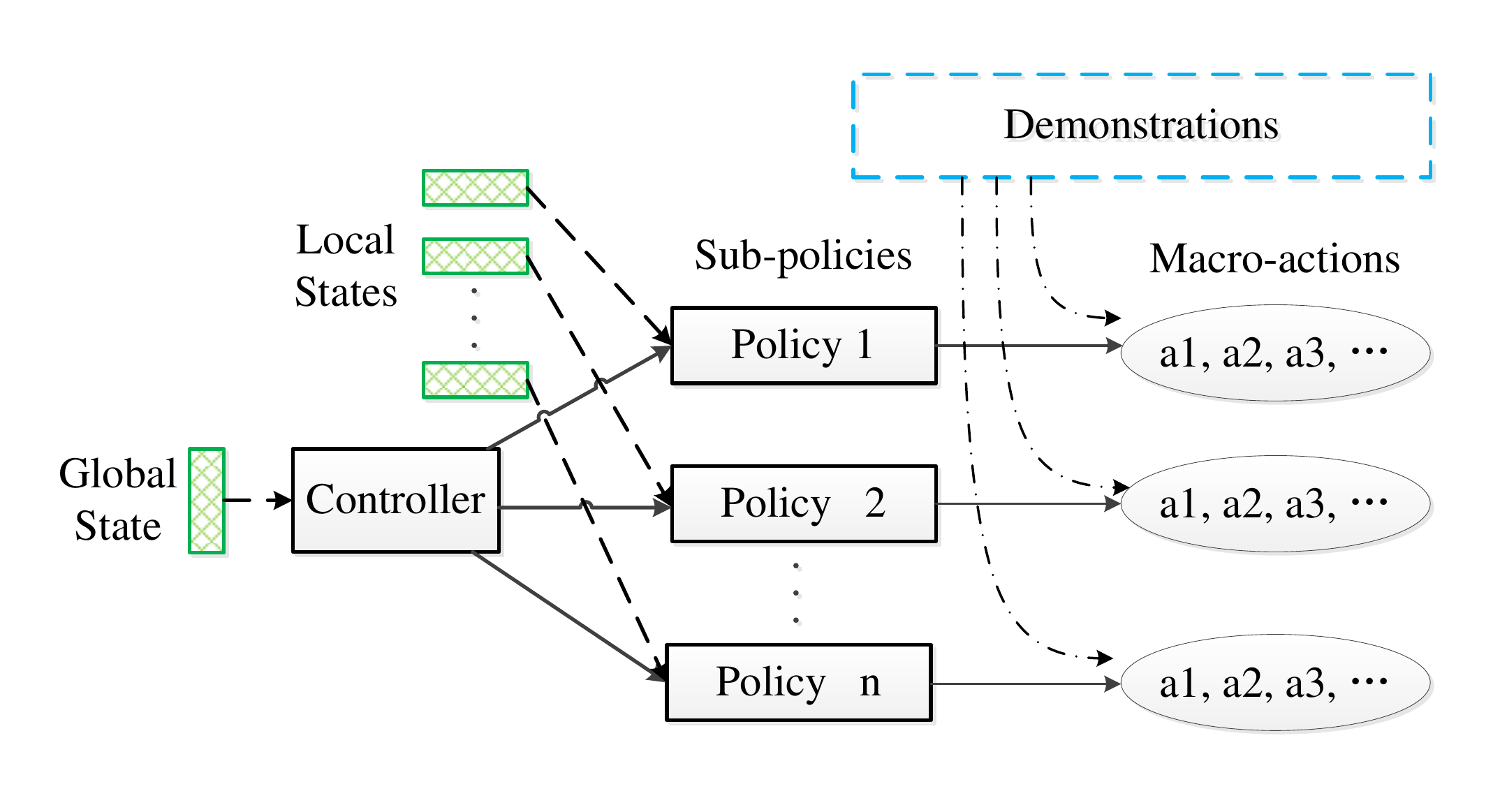}
		\caption{Overall Architecture. The explanation is in Section~\ref{subsection: Hierarchical Architecture}.}
		\label{fig:Arch}
		\end{minipage}
\end{figure}

Our hierarchical architecture is illustrated in Fig.~\ref{fig:Arch}. There are two types of policies running in different timescales. The controller decides to choose a sub-policy based on current global observation in every long time interval, and the sub-policy picks a macro action in every short time interval.

For further illustration, we use $\Pi$ to represent the controller. $S_c$ and $A_c$ are its state and action space, respectively. It is noted that $S_c$ denotes the global state~\footnote{In this partial observability game, the ``global state" is not the whole state without fog-of-war.} and is small. $A_c=\{1, 2, ..., n\}$ means the index of the sub-policy. Assuming there are $n$ sub-policies in the pool, we use $\langle \pi_1, \pi_2, ..., \pi_n \rangle$ to represent them. The state space and action space of the $i$th sub-policy are defined as $S_i$ and $A_i$, respectively. Meanwhile, $R_i$ is its reward function. Besides, we have a time interval $K$. It means that the controller chooses a sub-policy in every $K$ time units and the chosen sub-policy makes a decision in every time unit. Now, we can deeply go through the whole process.

At time $t_c$, the controller gets its own global observation $s^c_{t_c}$ (the superscript $c$ means the ``controller''). Then it will choose a sub-policy $i$ based on its state like below:
\begin{equation}
    a_{t_c}^c =\Pi(s^c_{t_c}), \quad s^c_{t_c} \in S_c, \quad a^c_{t_c} \in A_c
\end{equation}
Now the controller will wait for $K$ time units and the $i$th sub-policy (note that $i = a^c_{t_c}$ meaning the index of the sub-policy) begins to make its move. Assume its current time is $t_i$, and its local observation is $s^i_{t_i}$ (the superscript $i$ means the $i$th sub-policy), it gets the macro-action $a^i_{t_i} = \pi_i(s^i_{t_i})$. After the $i$th sub-policy doing the macro-action $a^i_{t_i}$ in the game, it will get the reward $r^i_{t_i} = R_i(s^i_{t_i}, a^i_{t_i})$ and its next local observation $s^i_{t_i + 1}$. The tuple $ (s^i_{t_i}, a^i_{t_i}, r^i_{t_i}, s^i_{t_i + 1})$ will be stored in its local buffer $D_i$ for future training. After $K$ moves, it will return to the controller and wait for the next chance.

The high-level controller gets the return of the chosen sub-policy $\pi_i$ and computes the reward of its action $a_{t_c}^c$ as follows:
\begin{equation}
    r_{t_c}^c = r^i_{t_i} + r^i_{t_i+1} + ... + r^i_{t_i+K-1}
\end{equation}
Also, the controller will get its next global state $s^c_{t_c+1}$ and the tuple $ (s^c_{t_c}, a^c_{t_c}, r^c_{t_c}, s^c_{t_c + 1})$ will be stored in its local buffer $D_c$. Now the time is $t_c + 1$, and the controller will make a next choice based on its current global observation. 

From the above, we can see some advantages to our hierarchical architecture. Firstly, each sub-policy and the high-level controller have different state spaces. The controller only needs the global information to make high-level decisions. The global state space $S_c$ is a small part of the overall state space $S$, e.g., the number of features in $S_c$ is nearly $1/3$ of the number of features in $S$. Also, a sub-policy responsible for combat is more focused on the local state space $S_i$ related to battle. Such a hierarchical structure can split the original huge state space into a plurality of subspaces corresponding to different policy networks. Secondly, the hierarchical structure can also split the tremendous action space $A$ into smaller ones. The sub-policies with different functions will have their own action space $A_i$, making the training easier. Thirdly, the hierarchical architecture can effectively reduce the size of execution steps. Since the control network calls a sub-network at every fixed time interval $ K $, the total execution step size of the high-level network becomes $ T/K $ step. The execution step size of each sub-policy will also be reduced because they do not need to be executed for the entire time horizon. Last but not least, the hierarchical architecture makes the design of the reward functions easier, e.g., different sub-policies may have different reward targets. Therefore, we can design different reward functions for different sub-policies and benefit from the specific reward functions.

We use stochastic policies for the controller and each sub-policies. Also, because we include the time-like feature (the ``game-loop'' in SC2) in the state for these policies, they are all non-stationary policies. Time interval $K$ is set to $8$ in this work. It is worth noting that we can continue to extend the layering under the sub-policy, making the two-layer architecture into three-layer or more. This change does not affect the overall training because each layer updates its replay buffer independently. In the following experimental sections, we mainly used a two-layer architecture. However, in the section of additional experiments, we use the three-layer hierarchical architecture when training the agent against the cheating difficulty level AIs.

\subsection{Generation of Macro-actions}
In StarCraft, the original action space $A$ is tremendous. Human players always need to do a sequence of actions to achieve one simple purpose. For example, if we want to build a building in the game, we have to select a worker, order it to build the building in a specific position, and make it come back after it is finished. The sequences of the actions for some simple purposes are stored in minds of human players. So we try to generate a macro-action space $A^{\eta}$ which is obtained through data mining from trajectories of experts. We then replace the original action space $A$ with the macro-action space $A^{\eta}$, which will improve the learning efficiency and testing speed. 

PrefixSpan~\cite{pei2001prefixspan} is a sequential pattern mining algorithm. It mines the complete set of potential patterns by using the idea of prefix-projection. For large databases, PefixSpan reduces the effort of candidate subsequence generation and the size of the projected database, thus providing more efficient processing. Therefore, we choose it as the data mining algorithm for mining the macro actions here.

The generation process of macro-actions is as follows:
\begin{itemize}
\item Firstly, we collect some expert trajectories which are sequences of operations $ a \in A $ from game replays.
\item Secondly, we use a PrefixSpan algorithm to mine the relationship of each operation and combine the related operations to be a sequence, called $a^{seq}$ of which max length is $C$. We then construct a set containing all these sequences, called $A^{seq}$ which is defined as $ A^{seq} =\{\, a^{seq} = (a_0, a_1, a_2, \cdots, a_i) \mid a_i \in A \text{ and } i \leq C, \,\} $.
\item Thirdly, we sort this set $ A^{seq} $ by $\operatorname*{frequency}(a^{seq})$ in game replays.
\item Fourthly, we remove duplicated and meaningless ones, remain the top $K$ ones. Meaningless refers to the sequences like repeated selection or camera movement.
\item Finally, the reduced set is marked as the newly generated macro-action space $A^{\eta}$.
\end{itemize}

Given hundreds of sequences, PrefixSpan will find the patterns whose frequency of occurrence is beyond the min\_spport threshold. Before inputting to the PrefixSpan, we analyze the replays and divide them into fragments (every 5 seconds for one fragment). For actions in these fragments, we store them as sequences: one sequence corresponds to one fragment.

We then use PrefixSpan to mine the patterns from these sequences. PrefixSpan outputs the patterns and their frequency. We output the top 75 frequent patterns but with the below restrictions: 1. There should be no repeated actions in this pattern. 2. The length of the pattern should be larger than two. 3. The first action in the pattern should be the select type (e.g., select army or select point).

After the mining, we postprocess these patterns: 1. If this pattern contains more than one select action, we remove it; 2. If the selected unit types in this pattern cannot match the subsequent command actions, we remove them. We then output the filtered patterns as the macro actions we will use in the subsequent experiments.

Using the macro-action space $A^{\eta}$, our MDP problem is now reduced to a simpler one, which is defined as:
\begin{equation}
    M =\langle S,A^{\eta}, P(.), R(.), \gamma, T \rangle,
\end{equation}
Meanwhile, the MDP problem of each sub-policy is also reduced to a new one like:
\begin{equation}
    M_{i} =\langle S_{i}, A_{i}^{\eta}, P(.), R_i(.), \gamma, T_{i} \rangle, \text{ for } i = 1 \text{ to } n
\end{equation}
For simplicity, though each sub-policy uses the macro-actions, in the following section we use $A_{i}$ to represent it, not $A_{i}^{\eta}$. Using our macro actions, we reduced the number of actions from hundreds to dozens, thus the training becomes easier.

\subsection{Training Algorithm}
The training algorithm of the hierarchical architecture is shown in Algorithm~\ref{alg: train} and can be summarized as follows. Firstly we initialize the controller and sub-policies. Then we run a total of $Z$ iterations. In each iteration, we run the game for $M$ episodes. We will clear all the replay buffers at the beginning of each iteration. In each episode, we collect the trajectories of the controller and sub-policies. We use the replay buffers to update the parameters of the controller and sub-policies at the end of each iteration.

The update algorithm we use is PPO~\cite{schulman2017proximal}. Entropy's loss was added to the PPO's loss calculation to encourage exploration. Therefore, our loss formula is as follows:
\begin{equation}
    L_t(\theta) = \hat{\mathbb{E}}_t[-L_t^\text{clip}(\theta) + c_1 L_t^\text{vf}(\theta) - c_2 S[\pi_\theta](s_t) ]
\end{equation}
where $c_1, c_2$ are the coefficients we need to tune, and $ S $ denotes an entropy bonus. $L_t^\text{clip}(\theta)$ is defined as follows:
\begin{equation}
    L_t^\text{clip}(\theta) = \hat{\mathbb{E}}_{t}[\text{min}(r_t(\theta)\hat{A}_t, \text{clip}(r(\theta), 1-\epsilon, 1+\epsilon)\hat{A}_t]
\end{equation}
\begin{equation}
    L_t^\text{vf}(\theta) = \hat{\mathbb{E}}_{t}[ (r(s_t, a_t) + \hat{V}_t(s_t) - \hat{V}_t(s_{t+1}))^2 ]
\end{equation}
where $r_t(\theta)= \frac{\pi_\theta(a_t|s_t)}{\pi_{\theta_{old}}(a_t|s_t)}$, $\hat{A}_t$ is computed by a truncated version of generalized advantage estimation.

Note that we have also tested imitation learning to train an agent by expert replays. The trained agent can beat the level-1 built-in AI. However, we later find that we can directly train our agent using RL without imitation learning, so we will only present the RL training results here.

\begin{algorithm}[t]
\caption{The proposed HRL training algorithm} %
{\bf Input:} Number of sub-policies $N$, time interval $K$, reward function $R_1, R_2, ..., R_n$, max episodes $M$, max iteration steps $Z$, max game steps $T$,
\begin{algorithmic}[1]
\STATE initialize replay buffer $\langle D_c, D_1, D_2, ..., D_n\rangle$, controller policy $\Pi_\phi$, and each sub-policy $\pi_{\theta_i}$
\FOR{$z=1$ to $Z$}
    \STATE clear $\langle D_c, D_1, D_2, ..., D_n\rangle$
    \FOR{$m=1$ to $M$}
        \STATE clear $\langle \tau_c, \tau_1, \tau_2, ..., \tau_n\rangle$
        \FOR{$t=0$ to $T$}
            \IF {$t$ mod $K == 0$}
            	\IF {$s^c_{t_c-1}$ is not None}
            		\STATE $\tau_c \gets \tau_c \cup \{(s^c_{t_c-1}, a^c_{t_c-1}, r^c_{t_c-1}, s^c_{t_c})\}$
				\ENDIF
            	\STATE $a^c_{t_c} \gets \Pi_\phi(s^c_{t_c}), \quad r^c_{t_c} \gets 0$
            	\STATE $j \gets a^c_{t_c}$
            \ENDIF
            \STATE collect experience by using $\pi_{\theta_j}$
            \STATE $\tau_j \gets \tau_j  \cup \{(s_{t}^j, a_{t}^j, R_j(s_{t}^j, a_{t}^j), s_{t+1}^j\}$
            \STATE $r^c_{t_c} \gets R_j(s_{t}^j, a_{t}^j) + r^c_{t_c} $
        \ENDFOR
        \STATE $D_c \gets D_c \cup \tau_c$
	    \FOR{$i=1$ to $N$}
             \STATE $D_i \gets D_i \cup \tau_i$
        \ENDFOR
    \ENDFOR
    \STATE use $D_c$ to update $\phi$ to maximize expected return 
    \FOR{$i=1$ to $N$}
        \STATE use $D_i$ to update $\theta_i$ to maximize expected return
    \ENDFOR
\ENDFOR
\end{algorithmic}
\label{alg: train}
\end{algorithm}

\subsection{Reward Design}
The reward has a significant impact on reinforcement learning~\cite{rewardShaping}. There are usually two types of rewards we can use. One is the dense reward that can be got during the game, and the other is the sparse reward which is only shown on a few specific states. Commonly, dense reward gives more positive or negative feedback to the agent and can often help agents learn faster and better than the sparse reward.

There are three types of rewards we explore in this paper. The win/loss reward (or called outcome reward) is a ternary 1 (win) / 0 (tie) / -1 (loss) received at the end of a game. The score reward is the Blizzard score get from the game engine. The designed reward is our designed reward function.

It's hard for agents to learn the game using win/loss reward. Blizzard scores can be seen as a dense reward. We will show that using this score as a reward can not help the agent get more chances to win in the experiment section. We have designed reward functions for the sub-policies which combine dense reward and sparse reward. 

Our reward design draws on the fact that human beings can focus on not only the results but also the performance in the middle of the process. They are as follows:
\begin{enumerate}
	\item We collect numerous replays of games from the experts (the same we used to generate macro-actions).
	\item We calculate the average number for each unit type $i$ built at the end state $z$ of all the replays, called $\bar{N^i_z}$.
	\item In each step $j$ of the game, we calculate the difference between each number of units and buildings in each of the two continuous steps of $j$ and $ j+1 $, which is that $d^i_{s_{j}} = N^i_{s_{j+1}} - N^i_{s_{j}} $.
	\item We calculate the reward based on the rule: if the number $N^i_{s_{j+1}}$ is less than the expert's $\bar{N^i_z}$, add the reward $d^i_{s_{j}}$, otherwise, minus the reward $d^i_{s_{j}}$.  We then sum all the rewards for all units and buildings and all the time steps.
	\item At the end of the game, if the episode lasts for $M$ minutes, we give the end state a reward using weight $ -\alpha $ multiplied by the $M$, which encourages the agent to win fastly.
	\item We give the end state a result reward which is to multiply the outcome (-1, 0, or 1) of the game by a weight $\beta$ which is let the agent also focus on the results. 
\end{enumerate}
The value of the parameters $\bar{N^i_z}$, $\alpha$, and $\beta$ are provided in Appendix~\ref{append:Reward Functions}. We find these rewards seem to be effective for training, which will be shown in the experiment section. 

The designed reward functions need much human knowledge to develop. On the contrary, the win/loss reward is straightforward and provides the actual learning results we want the agent to get. Can we only use the win/loss reward to train our agent? This question is not answered in the conference version of our paper. However, it is solved in this extended version. In Section~\ref{section: Additional Experiments} of additional experiments, we provide how to use only the win/loss reward to train an agent that can beat against the cheating level built-in AIs.

\subsection{Curriculum Learning}
Curriculum learning~\cite{curriculumLearning} is an effective technique that can be applied to RL~\cite{curriculumLearningRL}. It designs a curriculum from easy to hard. Agents can learn from this curriculum to improve their abilities. We use the idea of curriculum learning here to help our agents train on difficulty levels.

SC2 includes 10 difficulty levels of built-in AI which are all crafted by rules and scripts. From level-1 to level-10, the built-in AI's ability is constantly improving. Starting from level-3, the level of the built-in AI is close to that of ordinary human players. Starting with level-8, built-in AI begins to use cheating techniques. In the most difficult three levels (8, 9, 10), the built-in AIs get more resources and visions by cheating and thus are very difficult to beat. Training in higher difficulty levels gives less positive feedback, making it difficult for agents to learn from scratch. In this work, we first trained our agent in lower difficulty levels, then transferred the agent to a higher difficulty level using the pre-trained model as the initial model, following the idea of curriculum learning.

However, when the pre-trained agent transfers to high difficulty levels, we find that if the controller and all sub-policies are still updated at the same time, the training is sometimes unstable due to the mutual influence of different networks. We have devised a strategy to update the controller and sub-policies alternatively in response to this situation. We found that this method can make the training more stable, and the winning rate for high difficulty levels can rise steadily.

\section{Experiments} \label{section: Experiments}
In this section, we present the experiment results on SC2LE. We first introduce the experimental settings about the hierarchical architecture and three combat models. Then the details of the experiments are shown. Lastly, we discuss the results and analyze the possible causes.

\subsection{Setting} \label{subsection: Setting}
The setting of our architecture is as follows: a controller selects one sub-policy every $8$ seconds, and the sub-policy performs macro-actions every $1$ second. In the setup, we have two sub-policies in the sub-policy pool. One sub-policy controls the construction of buildings and the production of units in the base, called the base network. The other sub-policy is responsible for battle, called battle policy. This sub-policy has three different models, which are explained later.

\begin{figure}[h]
	\begin{minipage}[t]{\linewidth}
		\centering
		\includegraphics[width=0.85\textwidth]{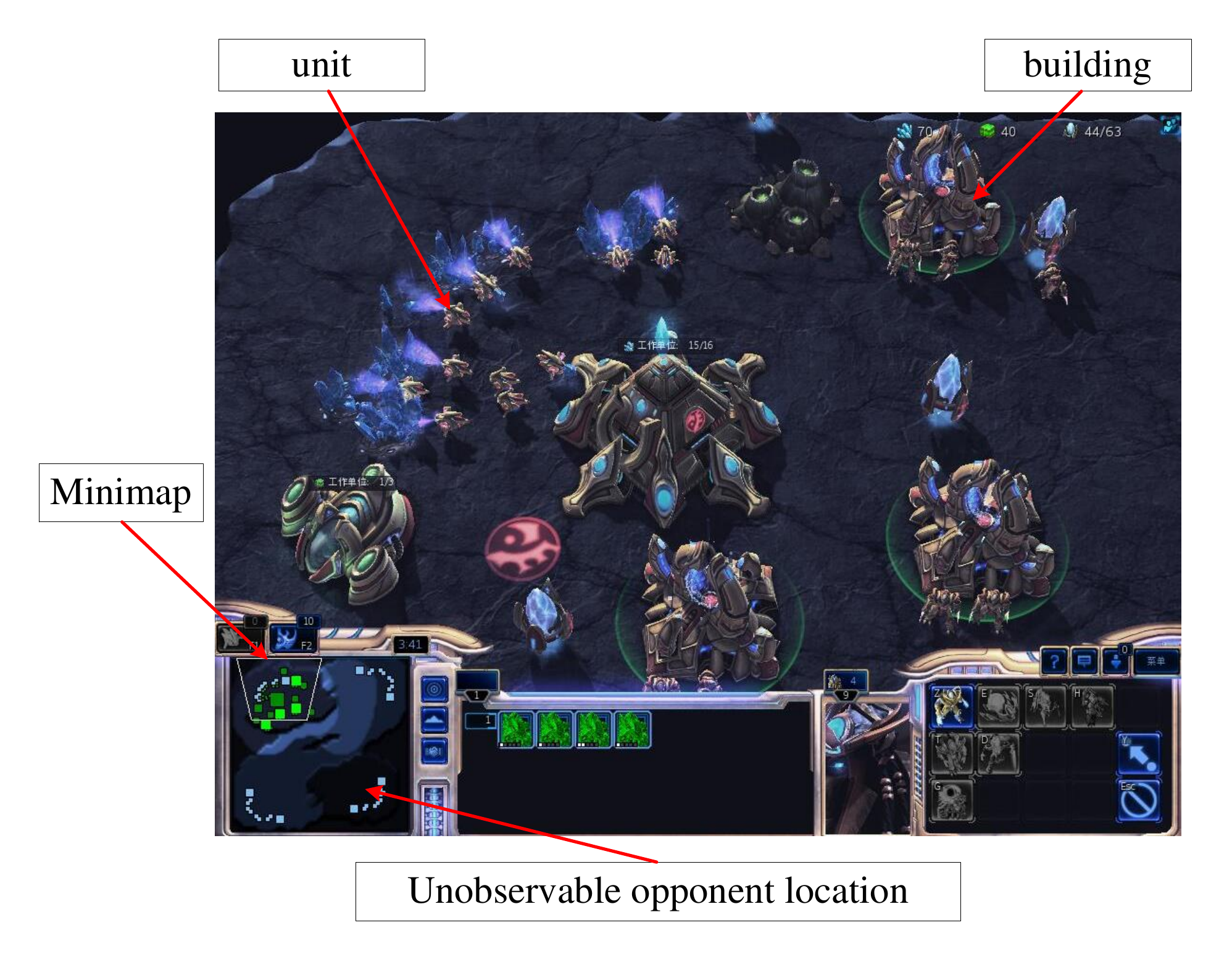}
		\caption{Screenshot of StarCraft II. See Section~\ref{subsection: Setting} for details.}
		\label{fig:pic}
	\end{minipage}
\end{figure}

A full-length game of SC2 in the 1v1 mode is as follows: Firstly, two players spawn on different random locations in the map. Secondly, they accumulate resources, construct buildings, and produce units. Thirdly, they attack and kill each other's units and buildings. Finally, the one who destroys the other's all buildings wins. Fig.~\ref{fig:pic} shows a screenshot of the running game. The Python interface for SC2LE is called PySC2. We use the 3.16.1 version of SC2, the first support version by PySC2.

For simplicity, we set our agent's race to \textit{Protoss} and built-in AI's race to \textit{Terran}. Our algorithms could generalize to the setting of any race against any race. The map we used is a 64x64 map called \textit{Simple64} in SC2LE. We set the maximum length of each game to 18000 game steps (about 15 minutes). For simplicity, our agent will not build any bases at sub-minerals and only uses two basic military units, which are \textit{Zealot} and \textit{Stalker}.

We use a single machine with 4 GPUs and 48 CPU cores for training. Our program has 10 workers, and each worker has 5 threads, meaning we have 50 SC2 environments running at the same time. The workers share the same global architecture and parameters. In each iteration, we run 100 full-length games. Therefore, each worker would collect data of 10 game episodes. It usually takes about 4 to 8 minutes for one iteration. When the worker has collected enough data, it will compute the gradients. When gradients of all workers are computed, they would be gathered and be used to update the overall networks. In our setup, the battle policy network has three different settings which are explained below.

\subsubsection{Combat Rule}
Combat rule is a simple battle policy, and there is only one action in the combat rule model: attacking a sequence of fixed positions. Although the map is unobservable and the enemy's position is unknown, the enemy always resides around the minerals. We only need to make our army attack the fixed positions around the minerals. The attack action uses the built-in AI to do automatic moves and attacks. Thus the result of the attack action depends on the construction of the buildings and the production of the units. Only when the agent learns to carry out building construction better (e.g., do not build redundant buildings, and build more pylons in time when supply is insufficient) can the agent win.

\subsubsection{Combat Network} \label{subsubsection: Combat network}
Though the simple combat rule model is effective, the combat process is slightly naive or rigid and may fail when moving to larger and more complex maps. Below we introduce a smarter attack approach which is called combat network. The output of the combat network consists of three actions and a position vector. These actions are as follows: all attack a certain position, all retreat to a certain position, do nothing. Attack and move positions are specified by the position vector.

The coordinate of location is specified by a position vector. The position vector is a one-hot vector that represents the eight coordinate points on the screen. We can imagine that there is a square in the center of the screen. The side length of the square is half the length of the side of the screen. These 8 points are evenly distributed on all sides of the rectangle, representing the position of the attack and movement. With this setup, the agent can be smarter to choose the target and location of the offense, thus increasing the flexibility of the battle. At the same time, because the combat network does not specify the location of the enemy, it can automatically learn and discover the possible existence of the enemy, so that it can maintain performance when moving to a larger and more complex map.

\begin{figure}[h]
	\begin{minipage}[t]{\linewidth}
		\centering
		\includegraphics[width=0.90\textwidth]{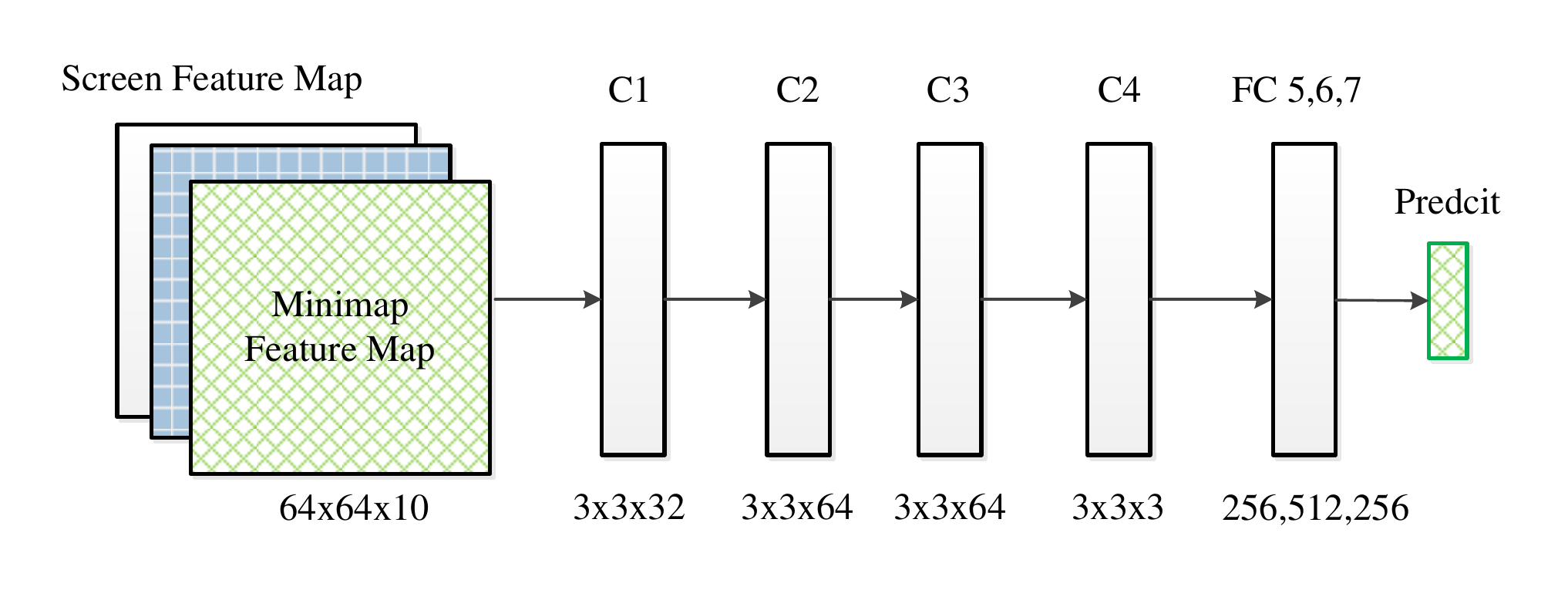}
		\caption{Architecture of the combat network. The explanation is in Section~\ref{subsubsection: Combat network}.}
		\label{fig:combatnet}
	\end{minipage}
\end{figure}

The combat network is constructed as a Convolution Neural Network (CNN) which is shown in Fig.~\ref{fig:combatnet}. This CNN accepts feature maps of minimaps and screens which enables it to know the information of the full map and the unit and building positions on the screen. Moreover, we use a simple strategy to decide the position of the camera. When the controller chooses the base sub-policy, the camera is moved to the location of the agent's base. When the controller chooses the battle sub-policy (alias for combat model), the camera is moved to the location of the army. The location of the army can be chosen in two ways. The first is the center point of all combat units. The second is the center of the most injured unit. Since the injured unit indicated the occurrence of a recent battle, we found that the second way was better in practice.

\subsubsection{Mixture Model}
Although the combat network can be trained well on the high level of difficulty, it will occasionally miss some hidden enemy buildings. We can combine combat network with combat rule into a mixture model. When a certain value is predicted in the position vector of the combat network (e.g., 0), the attack position will use the coordinate predicted by the combat rule.

\subsection{Comparison of Training Method} \label{subsection: Comparison of Training Method}
In the following, we will discuss our training processes from three aspects. Firstly, curriculum learning is the major way we use to train our agents against the enemy from low difficulty levels to high difficulty levels. Secondly, our hierarchical architecture can be used for module training, which is convenient for replacing sub-policies or refining some sub-policies while making others fixed. Finally, our simultaneous training algorithm is sometimes unstable on high difficulty levels, so we instead choose alternate training. Note the comparisons here are not exactly ablation studies in other common deep learning literature. Due to a full run of SC2 training needs 2 to 3 days. Thus we test in a small domain to test whether some tricks are effective. Meanwhile, we make sure that the settings in one test are the same for all algorithms. However, they may be different across tests.

Before training, we show the result of a simple (random) baseline which is randomly sampling using the learned macro-actions. The random sampling win rate for the agent on difficulty level-7 is 0.03 using macro actions (the win rate is 0.00 without the macro actions). From this, we can see that by constructing the macro actions, the agent can have some win rate against the hardest non-cheating levels, verifying the macro actions' effectiveness. From another way, we can see that the macro action can only give the agent a very small win rate (can not get win rate above 0.10). It is the usage of our training method to make the agent grow the win rate from a very low one of 0.03 to a very high one of 0.93, verifying the importance of the RL training.

\begin{figure*}[t]
	\begin{minipage}{0.235\linewidth}
		\centering
		\includegraphics[width=\textwidth]{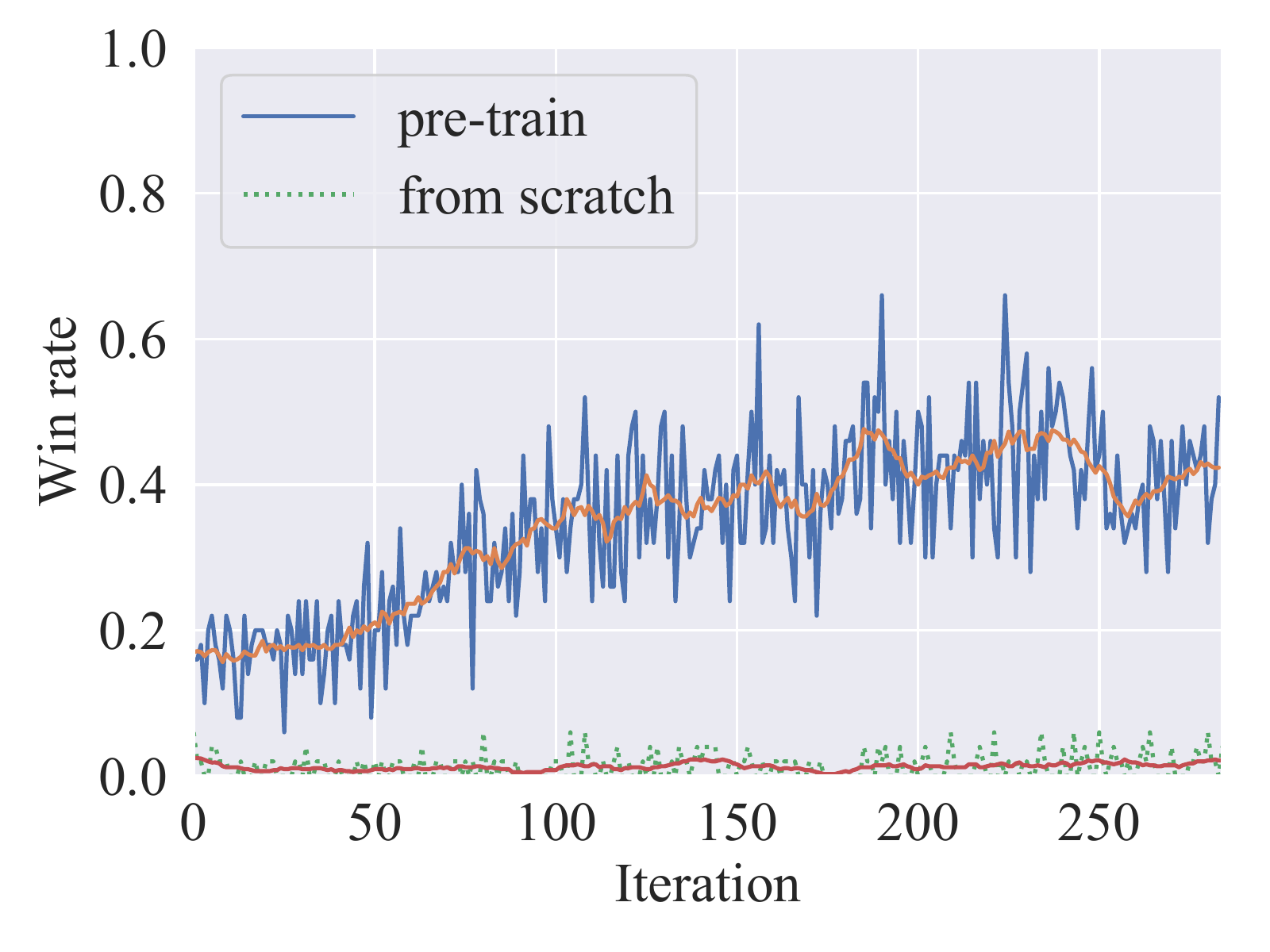}\\
		(a) Curriculum
		\label{fig:s.2.1.1}
	\end{minipage}%
	\begin{minipage}{0.235 \linewidth}
		\centering
		\includegraphics[width=\textwidth]{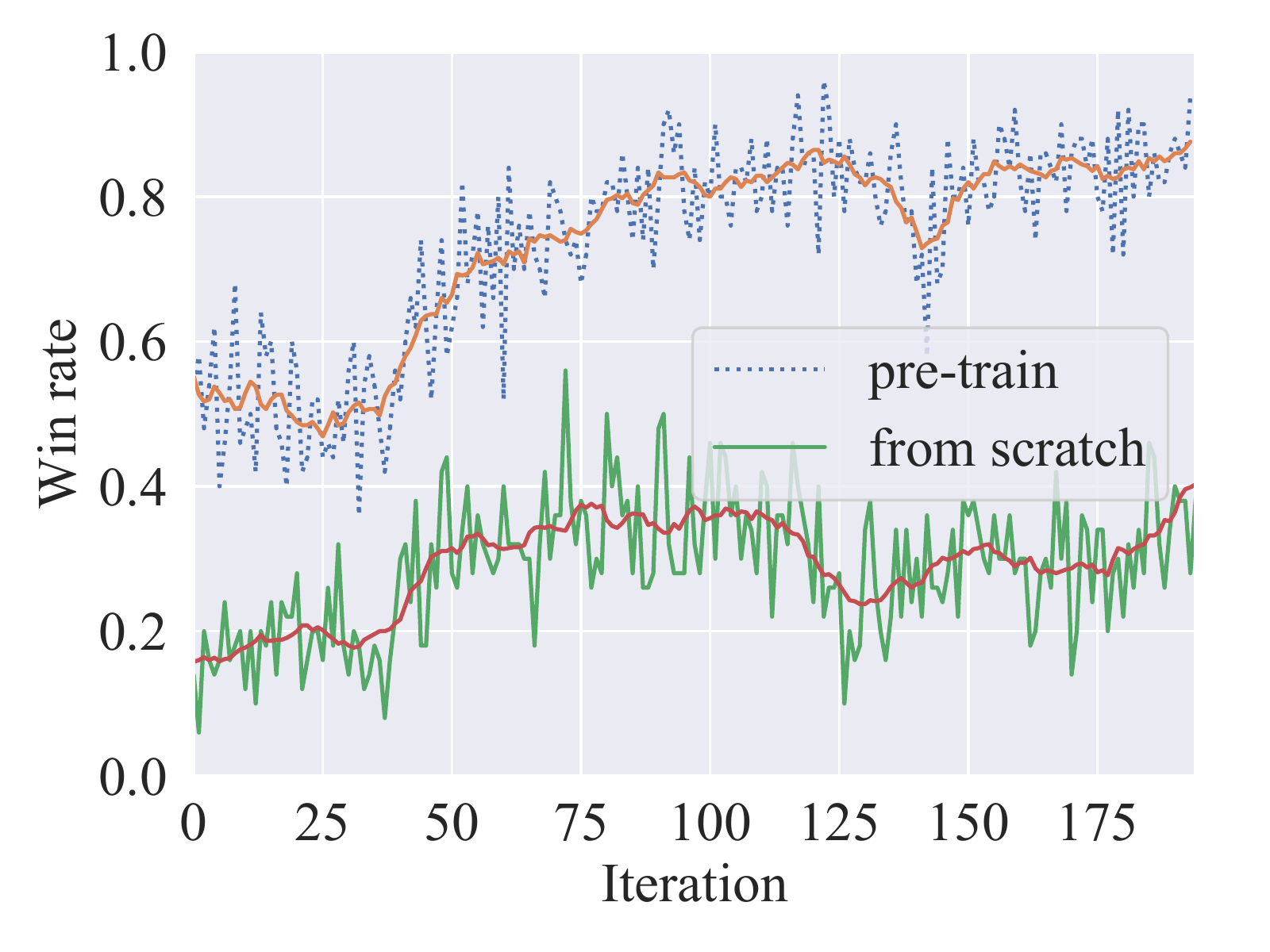}\\
		(b) Modular
		\label{fig:s.2.1.2}
	\end{minipage}
	\begin{minipage}{0.235 \linewidth}
		\centering
		\includegraphics[width=\textwidth]{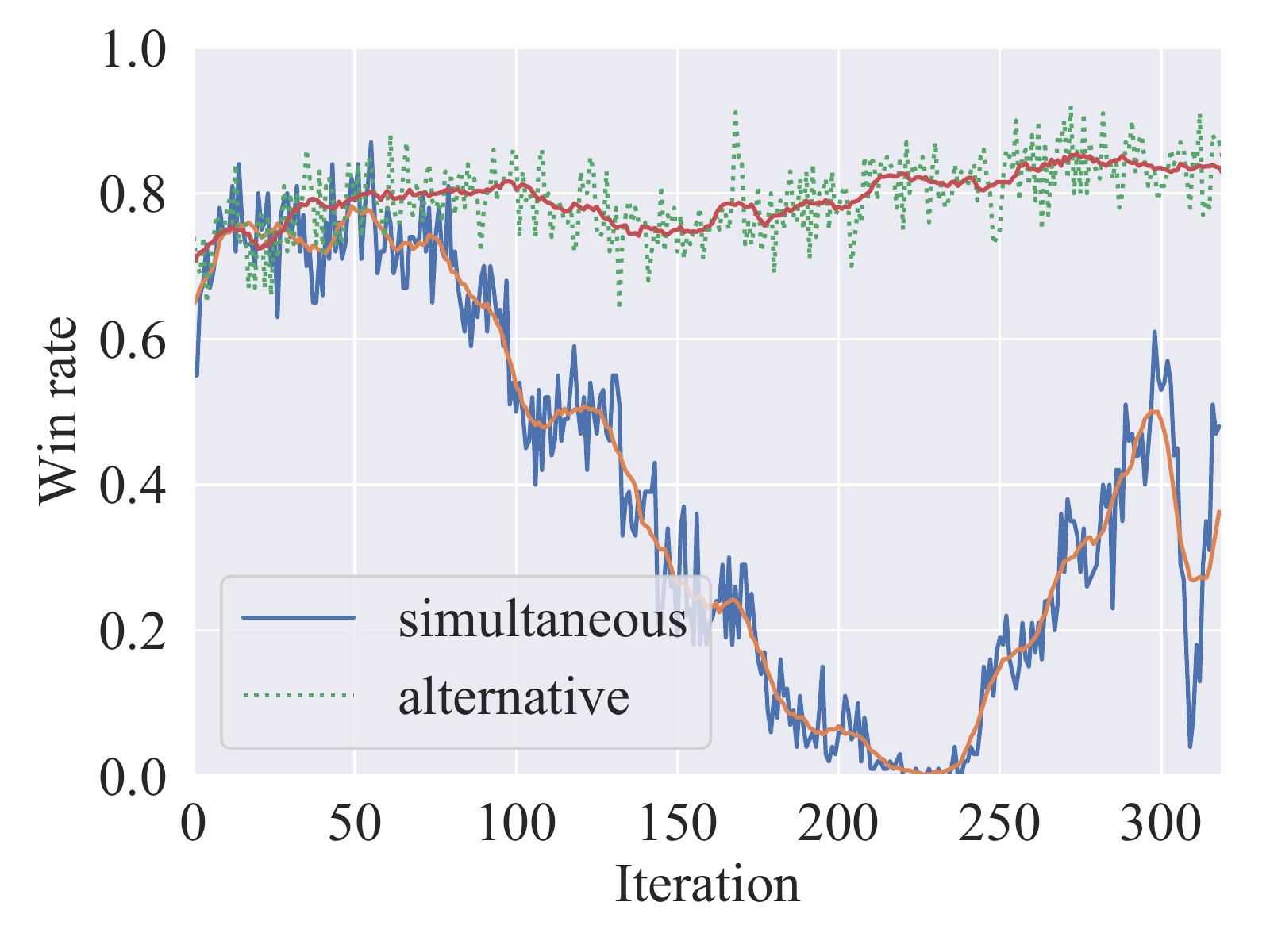}\\
		(c) Simultaneous
		\label{fig:s.2.2.1}
	\end{minipage}
	\begin{minipage}{0.235 \linewidth}
		\centering
		\includegraphics[width=\textwidth]{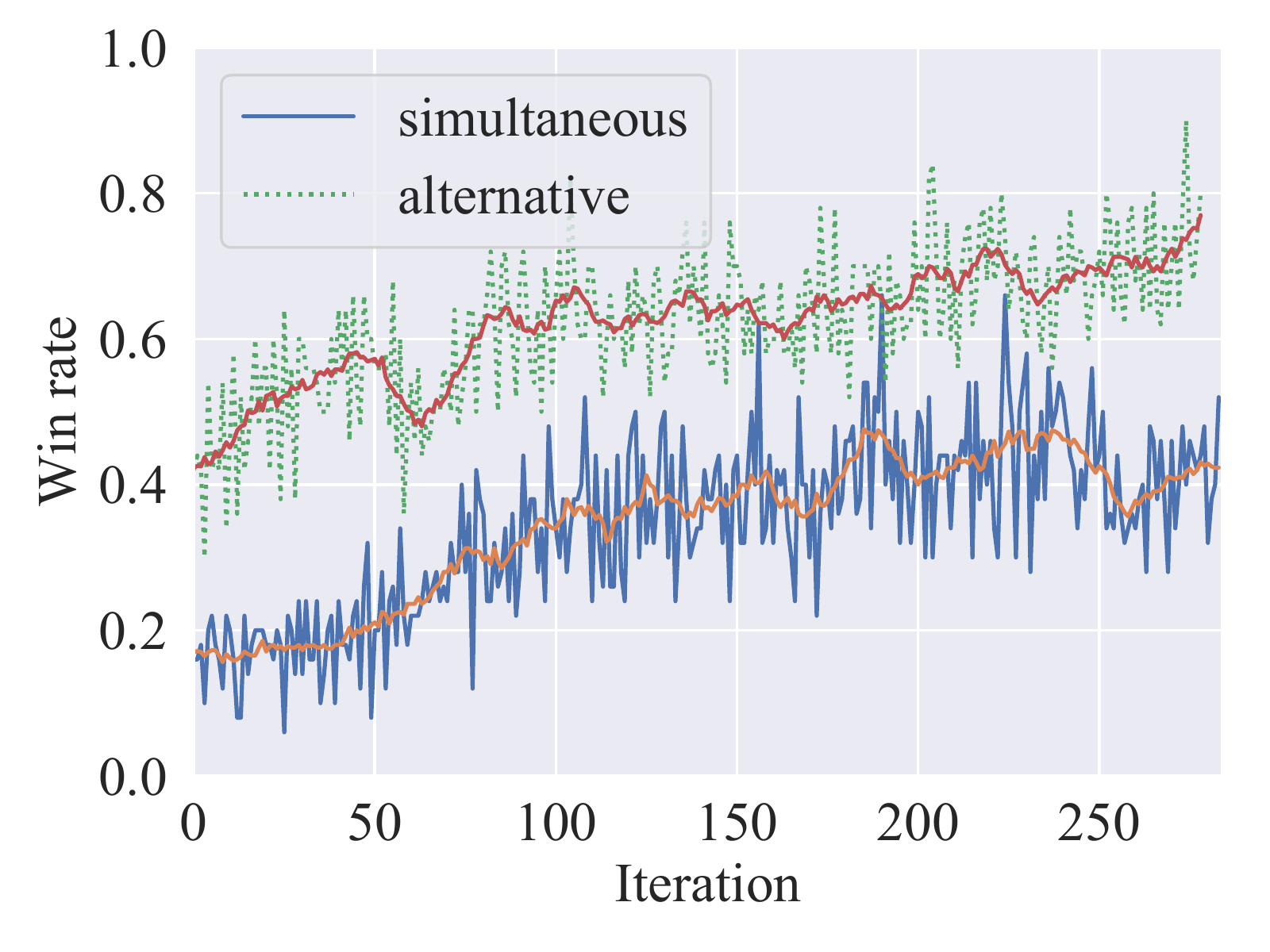}\\
		(d) Alternative
		\label{fig:s.2.2.2}
	\end{minipage}
	\caption{Comparison of Training Method. See Section~\ref{subsection: Comparison of Training Method} for details.}
	\label{fig:s.a.2}
\end{figure*}

\subsubsection{Effect of Curriculum Learning} \label{subsubsection: training settings}
In all of our combat models, we firstly train them on low difficulty levels and then transfer them to high difficulty levels. This process follows the way of curriculum learning. We find that when training directly on high difficulty levels, the performance of our agent is difficult to improve. We could often get better results if we start training from a low difficulty level and then use the low difficulty model as the initial model to train.

Fig.~\ref{fig:s.a.2} (a) shows the comparison between training the pre-trained agent and training from scratch on difficulty level-7. These two agents are using the combat network model, and the first agent has been trained on level-2 and level-5 (we run the experiment in Fig.~\ref{fig:s.a.2} two times to validate it is repeatable. But due to time constraints, we didn't run it for more times and do a statistical analysis).

\subsubsection{Effect of Module Training}
Since the hierarchical architecture we are using is modular, we can easily replace one of the sub-policies. While replacing a sub-policy, the parameters of the controller network and the other sub-policies are retained. Only the newly replaced sub-policy parameters are updated, which can accelerate learning. In the experiment, we have three types of combat models. If we have trained the simple combat rule model on a high difficulty level, we can use the parameters of its controller and other sub-policies for the other two combat models.

Fig.~\ref{fig:s.a.2} (b) shows the comparison between training using pre-trained networks and training from scratch on difficulty level-2. These two agents are using the combat network model. The first agent uses the pre-trained controller and base network parameters, which are retained from the combat rule model trained on difficulty level-7. Due to the first agent needing only to train the combat network while the other needing to train all the networks, the first agent has better training effects.

\subsubsection{Effect of Simultaneous Updating}
For sample efficiency, we let all the networks of the hierarchical architecture collect their own data and update, which may cause unstable problems. Because every network would consider the other networks as part of the environment and ignore the changes of others while updating. The unstable training rarely happens on lower difficulty levels but can be quite often on higher difficulty levels. We propose a strategy to alternately update each network in turn, resulting in a steady improvement. Note the alternate updating needs more samples than simultaneous updating. We can also make the learning rate smaller to alleviate the unstable problem, but the cost is that the training process would be much slower. Therefore we choose alternate updating.

Fig.~\ref{fig:s.a.2} (c) shows the comparison between simultaneous updating and alternate updating on difficulty level-7. The two agents both use the combat rule model and have been trained on level-2 and level-5. Fig.~\ref{fig:s.a.2} (d) also shows the comparison between simultaneous updating and alternate updating on difficulty level-7 but has some differences. The two agents both use the combat network model and have been trained on level-2 and level-5. The learning rate is half smaller than Fig.~\ref{fig:s.a.2} (c). The blue line uses simultaneous updating, and the green line uses alternate updating. We find that the blue line starting with a win rate of 0.2 is very hard to grow beyond the win rate of 0.4 in the later training process (after 250 iterations, not shown in this Figure). We then start the green line training by loading the model of the blue line. The green line grows quickly from the 0.4 win rate to a 0.8 win rate, showing that the alternate updating strategy is more effective than the simultaneous updating.

\subsection{Comparison of Combat Models}
After training the three combat models on difficulty level-7, we perform an evaluation for each of them. Each evaluation tests from level-1 to level-10. We run 100 games in each difficulty level and report the average winning rate. The results are shown in Table~\reflectbox{tab: evaluation results}. 

From level-1 to level-7, we find that the agents perform well. The built-in AIs from level-8 to level-10 use several cheating techniques and randomly select different strategies. So the performances are unstable and with a bit of randomness. However, it can be seen that our trained agents still have a comparative performance in fighting against them \footnote{The videos are at \url{https://www.youtube.com/channel/UCwsJy5CNjC38p3vN91IbZrw/videos}}.

\subsubsection{Combat Rule}
The combat rule model agent achieves good results in difficulty level-1 to difficulty level-5. We find that since agents can only use the two basic military units to fight, they are more inclined to use a fast attack fashion (called ``Rush''), which can guarantee the winning rate.

The behavior of our learned agent is as follows. In the beginning, the agent tends to produce more workers to develop the economy. When the number of workers is close to saturation, the agent will produce soldiers. When the number of soldiers is sufficient, the agent will attack. This layered progressive strategy is automatically learned through our reinforcement learning algorithm, which illustrates the effectiveness of our approach.

\subsubsection{Combat Network}
We also test the combat network model on all 10 difficulty levels. We find that although the combat network model achieves a good winning rate on difficulty level-7, the winning rates on several other difficulty levels are not high. It is worth noting that many game results are draws. This means that the model of the combat network is more difficult to eliminate the enemy's all buildings (due to the attack position being controlled by the network output while some enemy's buildings may be hidden somewhere). Although the built-in AI in the game has surrendered, there is no way for the agent to accept the surrender by PySC2's API (the current API has no functions to let the agent accept the surrender). Thus the agent must destroy all the enemy's buildings, otherwise, the result is a draw. Assuming we count the games in which our agent has nearly destroyed all the enemy's buildings but didn't find some hiding buildings to cause a draw as wins, the overall winning percentage of the model of the combat network is still high.

\subsubsection{Mixture Model}
It can be found in Table~\ref{tab: evaluation results} that the mixture model has achieved the best results from difficulty level-1 to level-9, which can be explained as follows. The agent not only can choose the attacking area within the camera but also can switch to a fierce attack on certain positions, of which freedom can lead to a performance improvement, that makes the mixture model the best in all three combat models. 

\begin{figure*}[t]
    \centering
    \subfloat[Zealot Rush]{
        \centering
        \includegraphics[width=0.47\columnwidth]{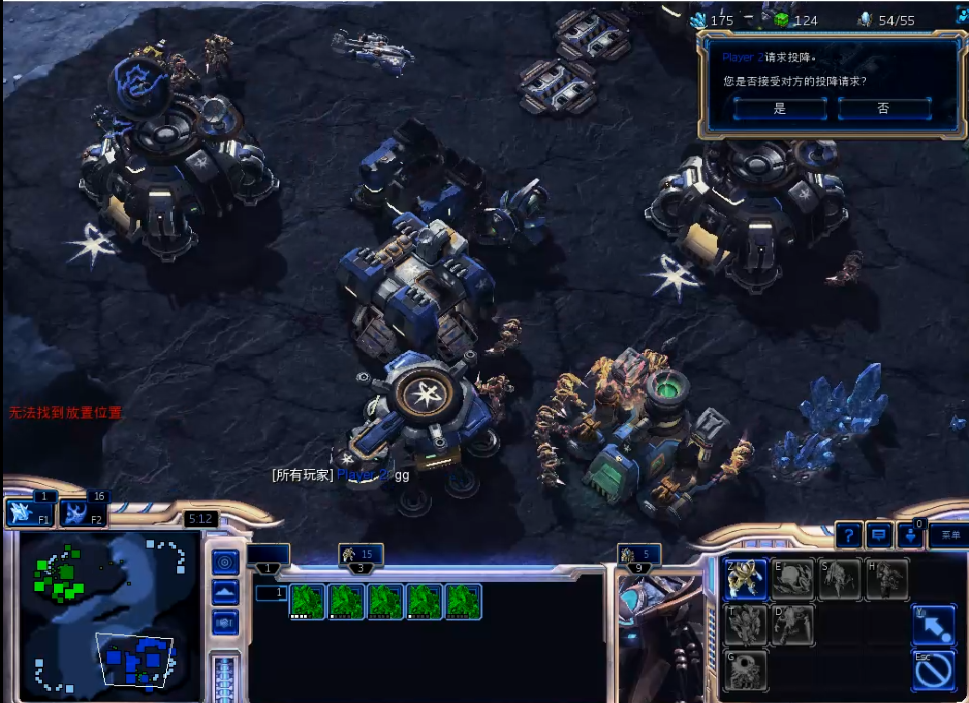}
		\label{fig:s.n.11.1}
   }
    \subfloat[Stalker Timing]{
        \centering
        \includegraphics[width=0.47\columnwidth]{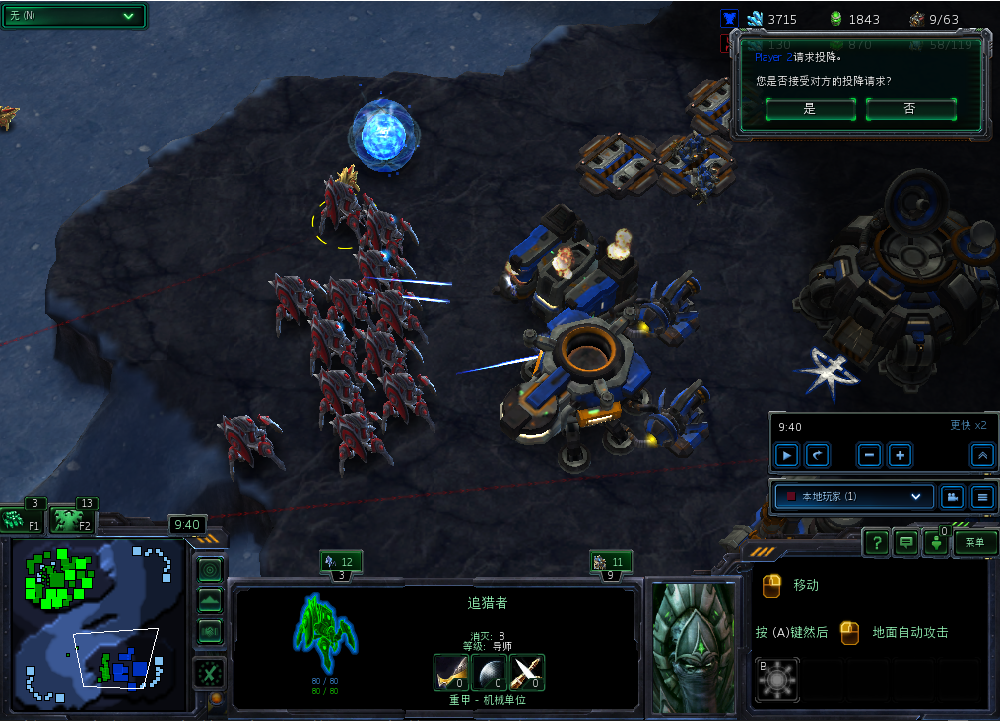}
		\label{fig:s.n.11.2}
    }%
    \caption{The training effects and the learned tactics of our hierarchical approach. (a): ``Zealot Rush''. (b): ``Stalker Timing''. It can be seen that by our approach, the agent can learn different tactics to defeat the opponent.}
    \label{fig:s.n.a.11.1}
\end{figure*}

Fig.~\ref{fig:s.n.a.11.1} (a) is the screenshot of our agent's learned tactic of ``Zealot Rush'' to defeat the level-7 built-in AI. This tactic is optimized through the ``trial and error'' of RL training. The agent's tactic is single yet powerful, which means that if it fights against other types of bots or humans, it can still have a chance to win. However, the agent is relatively hard to win if the opponent knows its tactics and uses effective anti-tactics.

Fig.~\ref{fig:s.n.a.11.1} (b) shows that if we add rewards for the agent to building Stalkers (one Stalker gives a +5 reward), the agent can learn a tactic of assembling many Stalkers and then make a Timing attack to win the game (against the level-2 built-in AI). From these two results, we can see that our agent can learn different SC2 tactics and master more than one aspect of StarCraft. The final agent focuses on using the Zealot rush because this tactic is the more effective strategy to beat the high-level built-in AI (including the cheating level ones) than the Stalker timing attack.


\begin{table*}[h]
	\centering
    \scalebox{0.95}{
    \begin{tabular}{ l | c c c c c c c | c c c }
        \hline
        Opponent's Type   & \multicolumn{7}{|c|}{Non-cheating (Training)} & \multicolumn{3}{|c}{Cheating (No-training)}  \\
        \hline
        Difficulty Level     & 1 & 2 & 3 & 4 & 5  & 6 & 7 & 8 & 9 & 10  \\
        \hline
        Combat Rule         & 1 & 0.99 & 0.94 & 0.99 & 0.95 & 0.88 & 0.78 & 0.70 & 0.73 &0.60 \\
        Combat Network   & 0.98& 0.99& 0.45& 0.47& 0.39& 0.73& 0.66& 0.56& 0.52  &0.41 \\
        Mixture Model       & 1& 1& 0.99& 0.97& 1& 0.90& 0.93& 0.74& 0.71  &0.43 \\
        \hline
    \end{tabular}}
	\caption{Evaluation Results. Results are tested by the agent trained on level-7.}
	\label{tab: evaluation results}
\end{table*}

\subsection{Comparison of Settings}
This section has three experiments to show the importance of hierarchy, design of rewards, and impact of hyper-parameters.
\begin{figure*}[h]
	\begin{minipage}{0.24\linewidth}
		\centering
		\includegraphics[width=\textwidth]{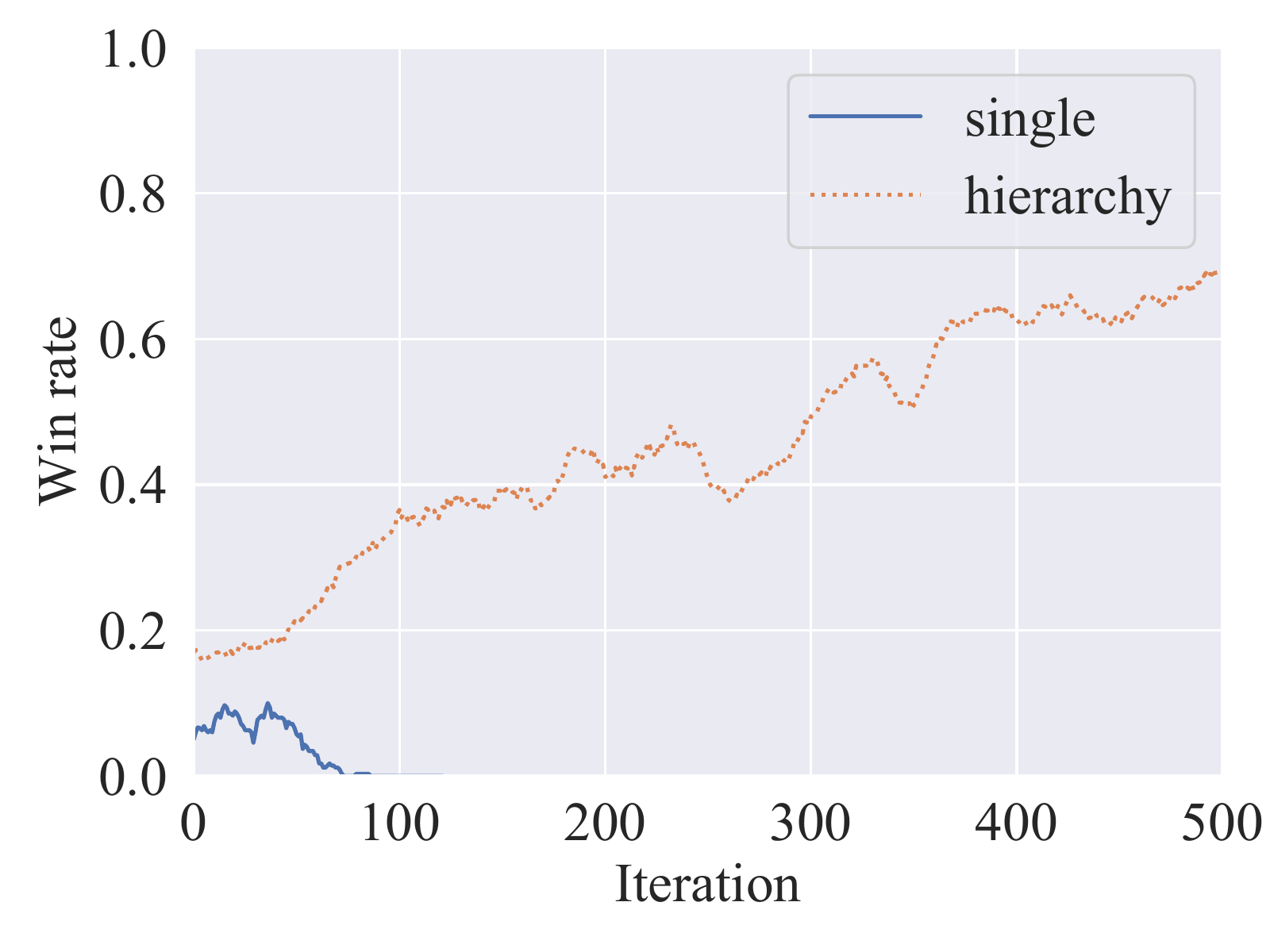}\\
		(a) Hierarchy vs. single
		\label{fig:s.3.1.1}
	\end{minipage}%
	\begin{minipage}{0.24 \linewidth}
		\centering
		\includegraphics[width=\textwidth]{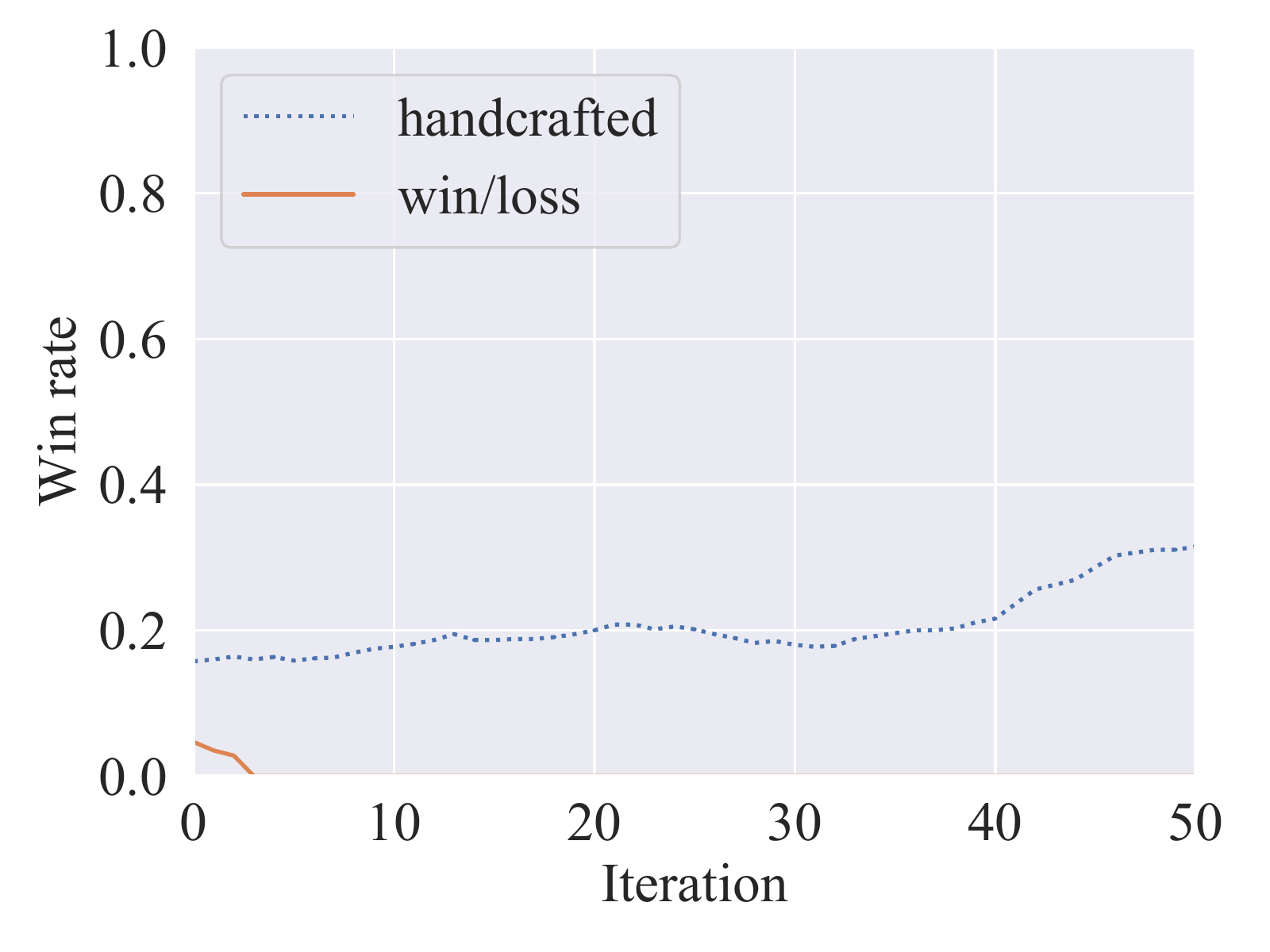}\\
		(b) win/loss vs. handcrafted
		\label{fig:s.3.1.2}
	\end{minipage}
	\begin{minipage}{0.24 \linewidth}
		\centering
		\includegraphics[width=\textwidth]{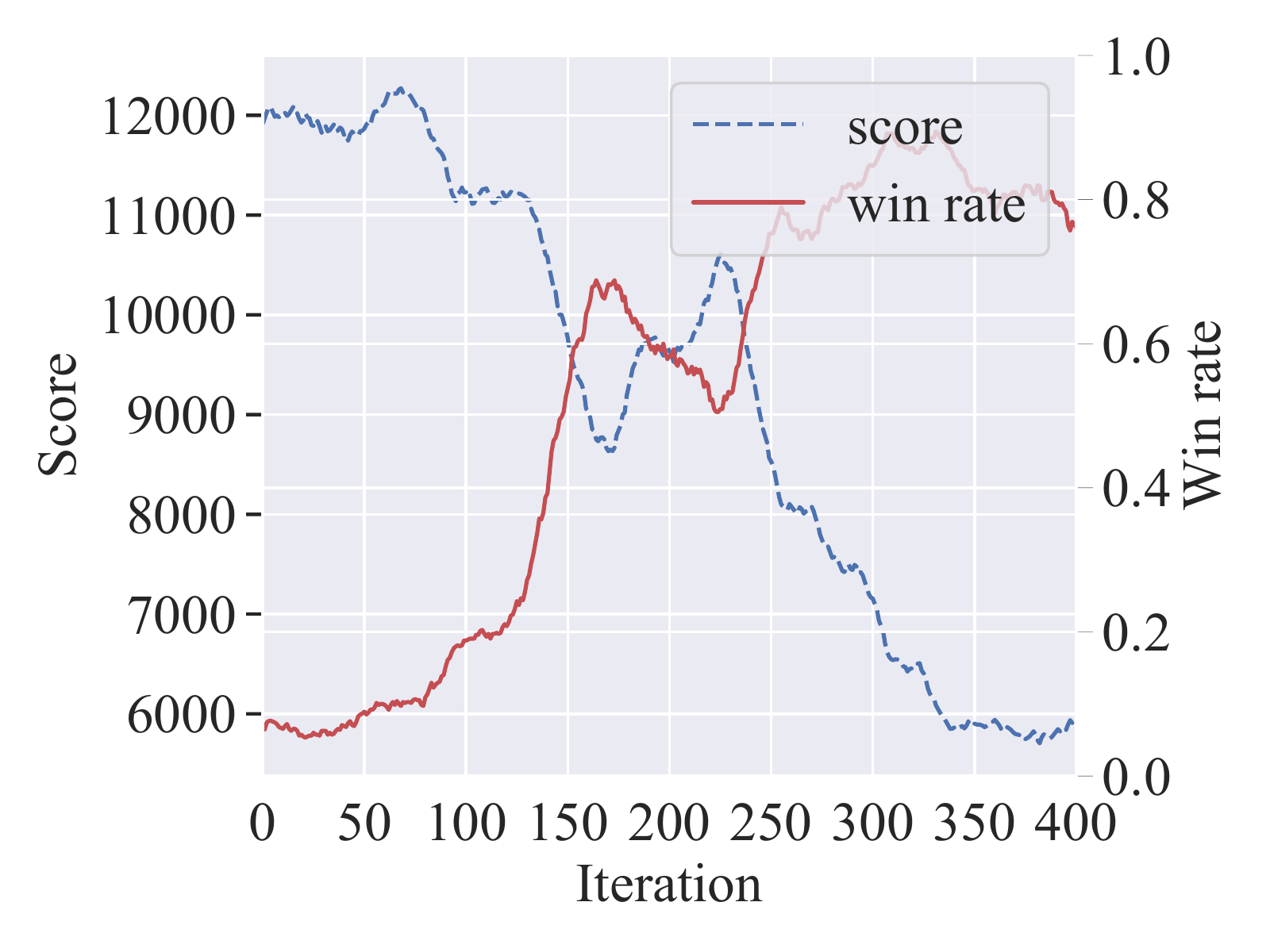}\\
		(c) Score and win rate
		\label{fig:s.3.2.1}
	\end{minipage}
	\begin{minipage}{0.24 \linewidth}
		\centering
		\includegraphics[width=\textwidth]{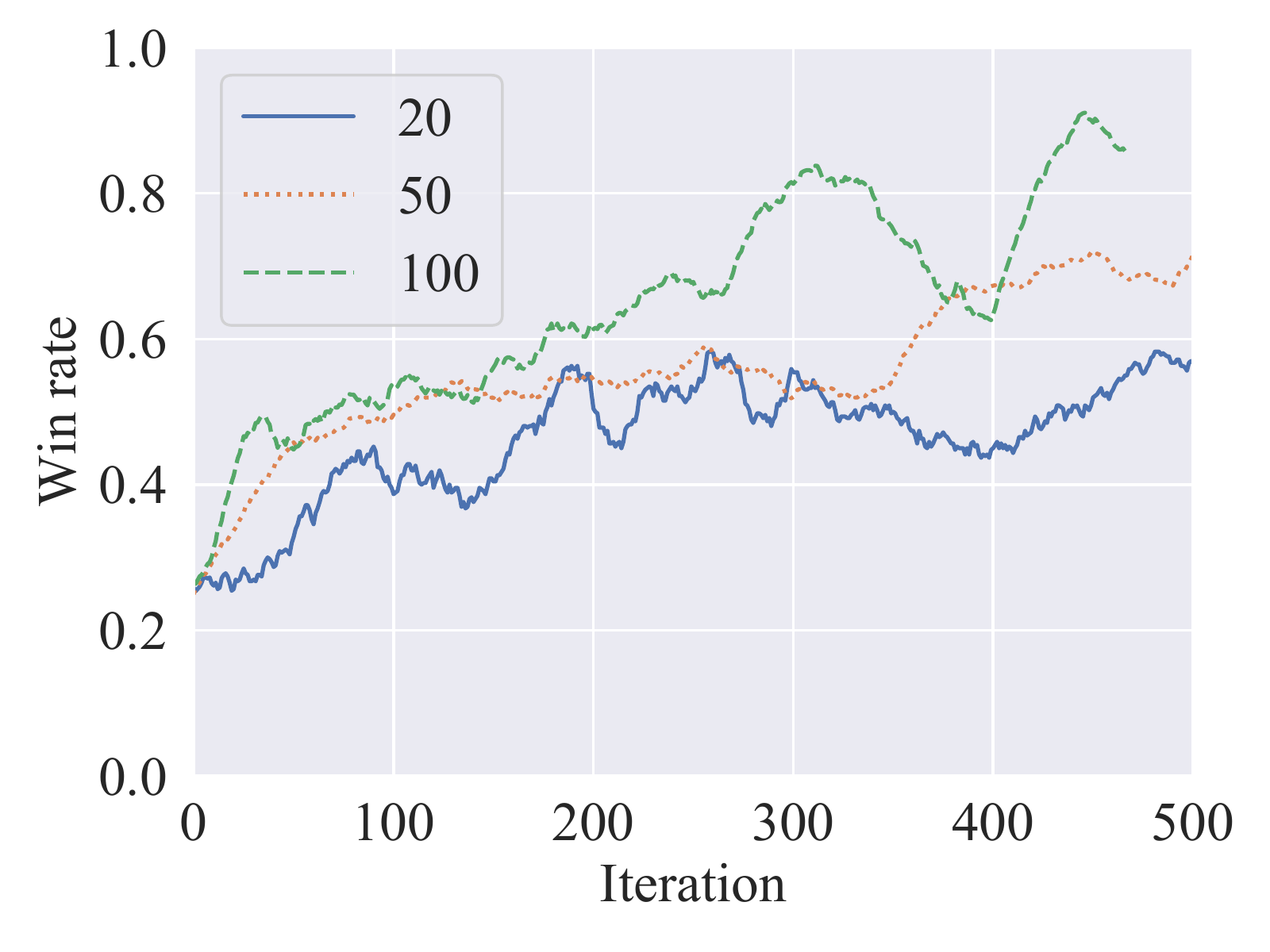}\\
		(d) Impact of episode num
		\label{fig:s.3.2.2}
	\end{minipage}
	\caption{Comparison of Settings. (a) and (b) are run against the level-7 built-in AI. (c) and (d) are run against the level-2 built-in AI. }
	\label{fig:s.a.3}
\end{figure*}

\subsubsection{Hierarchy vs Non-hierarchy}
Since we have used a hierarchical architecture, there is a question of whether a non-hierarchical architecture can handle the SC2 problem is a question. In the paper of SC2LE~\cite{vinyals2017starcraft}, the learning effect of the original state and action space and non-hierarchical RL algorithm is not satisfying. Our architecture has two levels of abstraction: one is the abstraction of the action space, which is done by macro-action, and the other is a multi-layer architecture that uses the controller to select sub-policies. We tested the effect of using macro-actions while not using the controller and sub-policies, which is called the single-policy architecture.

We were surprised to find that on difficulty level-2, the single-policy architecture can be almost as good as the hierarchical architecture, which means that macro actions are effective for training a good policy. However, when moving to the high level of difficulty (level-7), the final performance of the hierarchical architecture is significantly better than the single-policy architecture, as shown in Fig.~\ref{fig:s.a.3} (a) (we run the experiments in Fig.~\ref{fig:s.a.3} two times to validate it is repeatable. But due to time constraints, we didn't run it for more times). It can be explained that when the difficulty level is low, the difference between the hierarchical and non-hierarchical architecture is less obvious. When the difficulty of the problem continues to increase, the performance of the hierarchical model will be better. 

Another reason for using hierarchy is modularity. Modularity facilitates the replacement and refinement of each sub-policy. For example, if we need to replace the battle sub-policy, we can still retain the parameters of other networks, only training the battle sub-policy, which speeds up the learning process. Hence modular training can facilitate curriculum learning. Therefore, scalability is also a reason for us to choose this hierarchical framework.

\subsubsection{Comparison of Reward Function}
The win/loss rewards can achieve similar results on low difficulty levels as the handcrafted reward. However, when training is on high difficulty levels, we find that the performance of win/loss reward on the hierarchical model is relatively poor shows in Fig.~\ref{fig:s.a.3} (b). This figure shows a combat network model agent trained on difficulty level-7. 

In SC2LE, it is mentioned that one can use Blizzard score as a reward~\cite{vinyals2017starcraft}. However, we found that this score and winning rate are sometimes not in a proportional positive relationship, especially on low difficulty levels. When the agent tries to improve the score, it may ignore the attack chance and lose the best opportunity to win. This phenomenon is shown in Fig.~\ref{fig:s.a.3} (c) in which the agent is using the mixture model and being trained on difficulty level-2. The final performance of the agent trained in SC2LE by Blizzard score~\cite{vinyals2017starcraft} also shows this problem, which is the reason why Blizzard score is not an ideal reward function (but we can use part of them, which is shown in Appendix~\ref{append:Reward Functions}).

\subsubsection{Influence of Hyper-parameters}
We have experimented with a variety of different hyper-parameters and find that the number of episodes in each iteration has an influence on the learning speed of the agent. When this number is smaller, the learning speed will be slower. Improving the number of episodes can improve the training effect, which is shown in Fig.~\ref{fig:s.a.3} (d) in which the agent is using the combat rule model and trained on difficulty level-2. Other hyper-parameters have a smaller effect on training. The setting of all the hyper-parameters is shown in the Appendix.

\subsection{Comparison with TStarBots} \label{subsection:Comparison with TStarBots}
In this section, we discuss the differences between our work and Sun's work (TStarBots). Our work uses less human knowledge compared to theirs in two ways. First, they use humans to build macro-actions, which is costly. On the contrary, we mine macro-actions from experts' replays. For a different game, our approach can also benefit from learning macro-actions from replays instead of being in the burden of designing them. Second, Sun's approach uses a trick. Suppose the agent finds the action needed to be executed can't be taken (maybe the precondition is not satisfied). In that case, the agent will then instead perform the precondition action. This trick gives the agent more opportunity to perform an executable action. However, this trick needs human knowledge. On the contrary, our agent does not rely on this human knowledge and can execute any action at any time. This setting gives our agent the challenge. Our agent needs to be trained by our algorithm to know which action should be taken at the right time. Though in the reward function design and mining replays, we also use some human knowledge. But in total, we use less human knowledge than theirs. Moreover, Sun's work uses much more computing resources than ours. E.g., Sun's work uses 3840 CPU cores, while ours uses 48 ones. The complete comparison can be seen in Table~\ref{tab: compare TStarBots}.
\begin{table}[h]
    \centering
    \scalebox{1.0}{
    \begin{tabular}{l | c c c c }
    \toprule
      & Setting & Map & Level-7 & CPU cores \\
    \midrule
    Sun's  & Z vs. Z & AbyssalReef  &   0.99   &  3840   \\
    Ours  & P vs. T &  Simple64  &   0.93   &  48  \\
    \bottomrule
    \end{tabular}
    }
    \caption{Comparison with TStarBots.}
    \label{tab: compare TStarBots}
\end{table}

\subsection{Comparison with AlphaStar} \label{subsection:Comparison with AlphaStar}
This subsection presents the discussion and comparison of our work with AlphaStar. First, we introduce the interface of SC2LE, of which the training difficulties actually differ very much for different versions. Next, we present the two different action spaces that our work and AlphaStar use, concluding that our work and AlphaStar could not be compared based on the same action space. Note that these differences are little discussed in the previous literature. After that, thanks to our created and published work mini-AlphaStar (a limited variant of AlphaStar which can be trained with limited resources), we can give a comparison of our hierarchical approach with mAS using the same computing resources and time.

\subsubsection{Inferface of SC2LE}
The Python interface for the SC2LE is PySC2. All previous RL works on SC2 are based on PySC2, but different versions affect results and training difficulties. PySC2 defines the state and action space of the agent, which are essential factors that affect the RL training. The 1st version of PySC2 (PySC2$_1$) only contains one type of state and action space. The state space is the game image (including the screen image and minimap image, as the humans can see) and some statistics (like the agent's overall minerals). The game image is divided into several image layers (like heightmap, unit type, and so on) to facilitate computer processing (e.g., using DNN to process it), called feature maps in PySC2. There are two main types of actions: one is the so-called ``Select", the other is the ``Command''. Select action is used to select the entity (unit, building, or group), just like human players use the cursor to select an entity on the game screen image. Only after selecting the command action has a meaning. The command (also called ``Order'') action gives the selected entity a command, such as a move, attack, build, etc. Like humans must use the cursor to select an entity and give orders, the agent must first do select action and then do command action. However, the select action and the command action are all included in the action space and have no clear difference. It is the responsibility of the agent to learn the relationship between the two types of actions, which makes the training difficult. Note our work and Sun's work are all based on these actions, which are the only actions that can be used in PySC2$_1$. The space for these actions is called ``feature action space'' in PySC2$_1$. For easy to understand and be distinguished from the below action space, we call it ``human action space'' (HAS) here in this paper.

When we present the conference version of the paper, the 2nd version of PySC2 (PySC2$_2$) is released. It adds more information of units (like hitpoint, damage, range, and so on) in the state space and supports two agents fighting against each other. When AlphaStar was published on Nature, the 3rd version of PySC2 (PySC2$_3$) was released. PySC2$_3$ adds a new action space type, called ``raw action space'' (RAS). AlphaStar has long been using the RAS, a key factor distinguishing it from other works. Note that PySC2$_3$ has two types of action space: HAS and RAS, which have no intersection. Though it is not clearly stated in the Nature paper, through inferring from the paper and pseudocodes, we found that AlphaStar only uses the RAS of PySC2$_3$, which greatly reduces the action space, thus decreasing the training difficulties by a large margin. The reason will be explained later.

\subsubsection{Different Action Spaces}
\begin{figure}[h]
	\begin{minipage}[t]{\linewidth}
		\centering
		\includegraphics[width=0.85\textwidth]{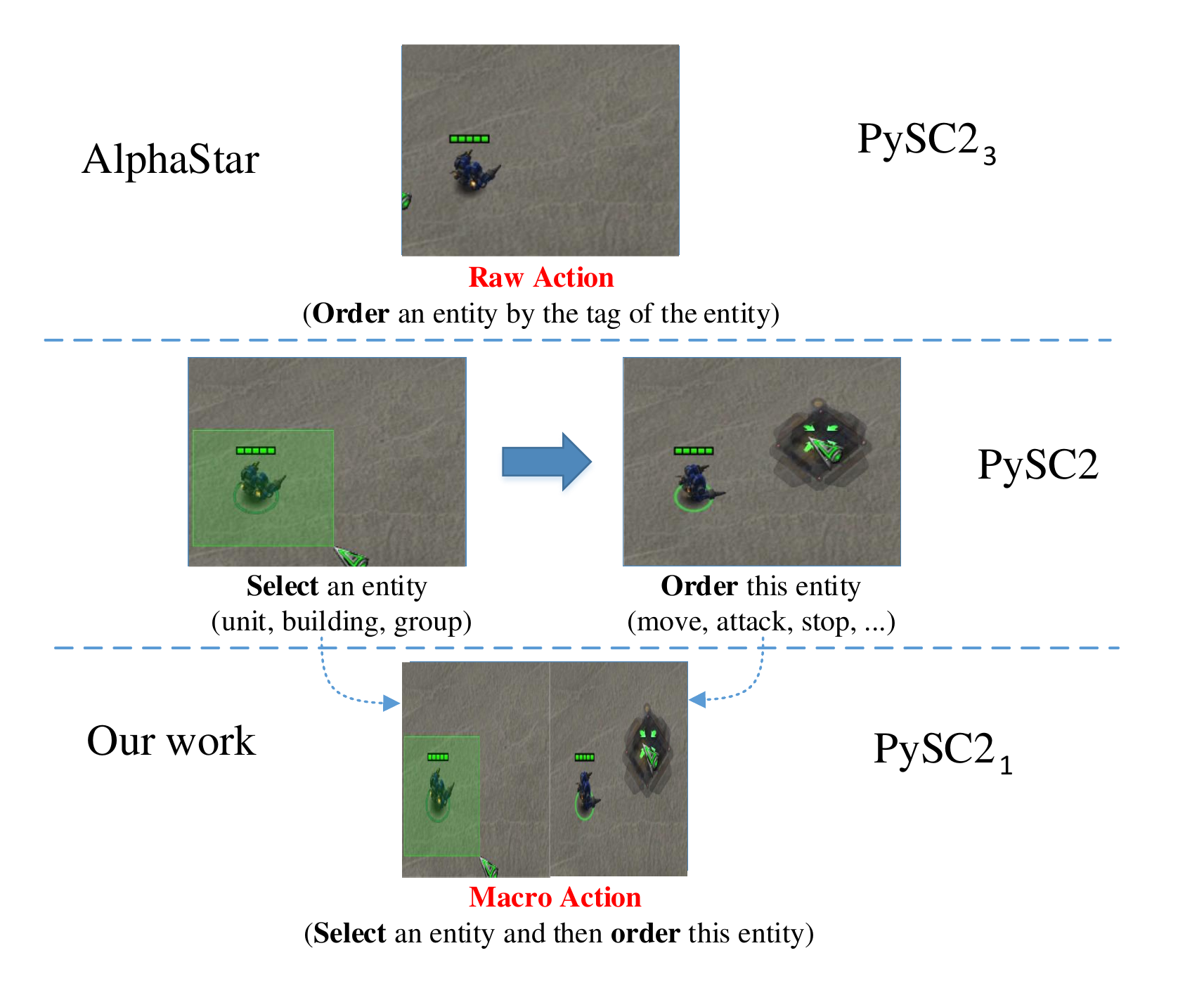}
		\caption{Raw action vs. Macro action based on Human action.}
		\label{fig:RawVsMacro}
		\end{minipage}
\end{figure}

The raw actions in the RAS are the actions only the game engine can execute. It doesn't select any entity but directly gives a command. Which entity is selected is given by the tag of the entity. The tag can be seen as a unique identifier of the entity through the game. The game engine maintains the tag, while human players do not know it. Thus, the raw action space AlphaStar executes is like the game engine can perform. It is similar to the bots in StarCraft I, not the mode like the purpose of creating PySC2 (simulating the human's actions). In contrast, if we use the human action space (HAS), we must select the entity from the game screen image (just like humans do). One possible way is to use computer vision methods to do object detection or segmentation on the screen image. However, there has been little research in this way until now. Our approach is to mine the relationships between these HAS actions and combine them into macro actions. The new space for macro actions is called macro actions space (MAS). Comparison of RAS and MAS are shown in Fig.~\ref{fig:RawVsMacro}.

Using the action that can only be accessed by the engine greatly reduces the difficulties for RL training. That's one of the reasons why A3C fails in the first experiment of SC2LE~\cite{vinyals2017starcraft}, but a similar algorithm succeeds in the AlphaStar~\cite{AlphaStarNature} (except for changes in the action space, the improvements of the neural network structure, the use of human knowledge in rewards, and more computing resource and training time are also the reasons for AlphaStar's success). The training algorithm in AlphaStar is similar to the A3C algorithm in the manner of an asynchronous, actor-critic way of training. However, when using the HAS, the agent not only needs to learn the position of the unit it locates but also how to give the commands for the unit. This setting increases the learning difficulties by a lot, making the A3C algorithm unable to achieve a win against the easiest built-in AI. In contrast, when using the RAS, the agent only needs to learn how to issue command actions, which makes the algorithm in AlphaStar able to learn an effective agent in complex scenes. 

\subsubsection{Discussion of Differences}
As discussed before, AlphaStar uses the RAS instead of the HAS as its action space which is only available in the PySC2$_3$. Due to timeline reason, our work and Sun's work only uses the PySC2$_1$, thus our work can only base on HAS. Due to the difficulties of training directly on the HAS, our work and Sun's work instead try to build macro-actions, removing the need for select action. Considering the PySC2$_3$ is now public, could our work switch to using the RAS? We argue it should not be, due to the following reasons.

First, because both the HAS and RAS reside as one type of action space in the PySC2$_3$, if we use PySC2$_3$, we can continue using the HAS. Second, the research on each action space should not be overlooked. RAS has the advantage of ease for agents to learn, but its results are only helpful for the game of SC2. On the contrary, the research on HAS may extend to other domains, as long as that domain's action space is similar (like select some entity and then do some action). Thus, though we can use PySC2$_3$ now, we still focus on learning based on human action space (HAS).

Though our work has a clear difference from AlphaStar, we can still compare with them with the help of our created mini-AlphaStar (mAS). By making the mAS' hyper-parameters adjustable, we can set its hyper-parameters small enough to fit into training on a single common commercial machine. Then we can compare our hierarchical approach with mAS under the same computing resource. Next, we will introduce the mini-AlphaStar, which may provide a baseline on future research of large-scale RL problems on SC2.

\section{Introduction of Mini-AlphaStar} \label{section: Introduction of mini-AlphaStar}
\begin{figure}[h]
	\begin{minipage}[t]{\linewidth}
		\centering
		\includegraphics[width=0.95\textwidth]{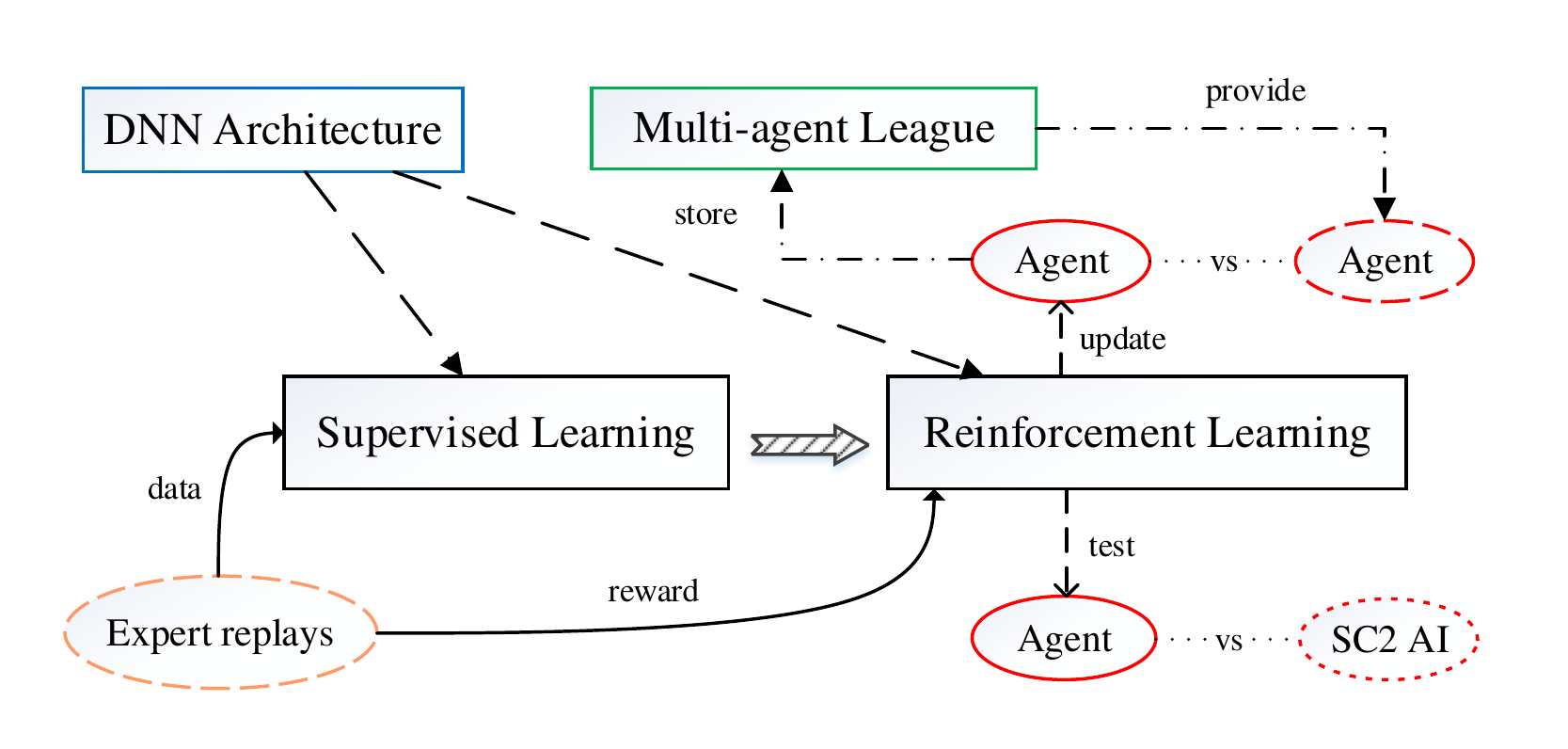}
		\caption{The mini-AlphaStar architecture. The explanation is in Section~\ref{section: Introduction of mini-AlphaStar}.}
		\label{fig: MiniAlphaStar}
	\end{minipage}
\end{figure}

Our mini-AlphaStar project has implemented most of the components of the AlphaStar program~\footnote{A technical report of mAS is at Arxiv with the name ``An Introduction of mini-AlphaStar" (\url{https://arxiv.org/abs/2104.06890}).}. It consists of four parts: neural network architecture, supervised learning method, reinforcement learning method, and multi-agent league training mechanism. Fig.~\ref{fig: MiniAlphaStar} shows the architecture of the mAS.

The DNN architecture is for both supervised learning and reinforcement learning. The data for supervised learning are expert replays. Meanwhile, mAS can also use expert replays as rewards in reinforcement learning. Reinforcement learning is applied in two ways in mAS: the agent playing against the built-in AI and the agent playing against the agent itself. If playing against the built-in AI, the training process is somewhat like the hierarchical approach in this paper. If in self-play, the training procedure will periodically select a suitable snapshot of the agent based on the multi-agent league mechanism to be the opponent of the agent.

Next, we will present the details of training mAS in one single common commercial machine. This machine is the same one as we train our hierarchical approach: 48 cores Intel(R) Xeon(R) Gold 6248 CPU 2.50GHz, a memory of 400G, disk space of 1T, and 8 NVIDIA Tesla V100 32G GPUs (in the hierarchical approach, we only use 4 of them). We first do the supervised learning and then do reinforcement learning.

\subsection{Supervised Learning in Mini-AlphaStar}
We first use human replays to do supervised learning. AlphaStar used 971K replays from version 4.8.2 to 4.8.6 to train its SL model. Differently, we use a replay pack of the 3.16.1 version provided by Blizzard. We use the 3.16.1 version of SC2 to transform these replays (because the game can only process the same version of replays) and then use newer versions to do RL training (because the 3.16.1 version doesn't support self-play with PySC2).

In mAS, we concentrate on using the Protoss race. Due to the agent and the opponent being all controlled by ourselves in self-play, we only collect the Protoss vs. Protoss (P vs. P) replays in the replay pack. Moreover, we only consider the replays on the map of AbyssalReef. We extract the 3.16.1 replay pack and filter replays that don't fit these conditions. There is a total of 608 replays after filtering. Later, to compare with our hierarchical approach, we will state how to collect (P vs. T) replays on Simple64.

We present how we process the replays for training. We sweep the replay step (or called frame) by step and only record the steps when the player gives actions. E.g., in step 576, the player orders a Probe to build a Pylon at a specific position. Thus the action type is ``Build Pylon'', and the two action arguments are: selected units being the Probe, and target location being the position of the building Pylon. The other action arguments are none. We stored the action type, action arguments, along with this step's state in the disk space. To be consistent with the convention in the SL training, we call the state the feature and call the action the label.

There are two ways to save the features and labels. One way is to save them as the tensor of the neural network's inputs, e.g., tensors of the deep learning frameworks. The other way is to save them as objects, like Pickles in Python. The former provides faster training but at the cost of more disk space. E.g., a replay file stored as a tensor costs about 20x more disk space than the Pickle file. The latter way needs that in the training phase we extract the Pickle file and process them to tensors, hence costing more time. The mAS supports both ways. We first tried to use Pickles for training but found it too slow. We then use the tensor way, but with an improved method (reducing the capacity by 70\%). So, next, we mainly use tensors to save replays.

We select some (e.g., 100) replays from all replays and divide them into one training set and one test set. We train dozons of epochs. In each epoch, we run over the training set by hundreds of iterations and do one evaluation over the test set. We collect one [BATCH, SEQ, FEATURE] feature and one [BATCH, SEQ, LABEL] label in each training iteration. We do a forward pass of the feature through the entire deep networks, getting one output of which shape is [BATCH, SEQ, PREDICT]. We then calculate one loss by the label and the output. Then we use the Adam optimization algorithm to update the parameters of the network. We set the value of BATCH and SEQ according to the available GPU memory and FEATURE and LABEL according to the state and action spaces (they are all adjustable). The value of PREDICT is the same as LABEL.

\subsection{Supervised Learning Details}
For comparison with our HRL work, we now give the details of mAS's SL training. SL is critical for AlphaStar's performance: it is hard for RL to explore a good policy without human data~\cite{AlphaStarNature}. We find the mAS agent with randomly initialed parameters can hardly get positive signals in training. E.g., the agent randomly chooses actions (most of these can't be executed due to their preconditions being not satisfied) at the early age of the training and can't do anything for a long time. To handle this problem, a trained SL model is critical for the performance of mAS's RL training, which is consistent with DI-Star~\cite{opendilab2021distar} and SC2IL~\cite{metataro2021SC2IL}.

The paper of AlphaStar~\cite{AlphaStarNature} didn't give details of SL, while the pseudocode in it is only an outline. However, SL is not easy and not trivial because AlphaStar's network architecture is quite complex. Hence, in this manuscript, we provide the training details as a compliment. For comparison with our HRL work, we train mAS on Simple64. Due to there being no official training replays on Simple64, we generated some replays by ourselves on this map, of which the number is 60. Combined with the replays for generating macro-actions, we have a total of 90 replays on Simple64. Then we do the following steps to do the SL training:
\begin{enumerate}
	\item Before training, we transform these replays to tensor data to eliminate the time-consuming processing process in the training phase.
	\item In the training phase, we train based on the processed replay tensors. We use auto-regressive training way to train the model.
	\item After one batch of training, we use multi metrics to evaluate the model based on this batch of data. After one epoch of training, we evaluate the model based on the test set.
	\item After a complete run of SL training, we evaluate the model in the RL environment to assess the trained SL model's performance. This step is essential.
	\item Based on the RL results, we judge whether we should go back to step 1 and do the following: modifying the transforming ways, improving the codes, or adding more replays. If the result is satisfying, the training is over.
\end{enumerate}

We find that preprocessing in the first step is critical for training speed. If we delay the preprocessing to the training phase, the speed may be 10x slower. We store one replay data as a sequence like [REPLAY-LENGTH, FEALAB] (FEALAB is the concatenation of FEATURE with LABEL). We then concatenate all replays' data (we didn't handle the problem of stepping over from one replay to the next) to [ALL-LENGTH, FEALAB]. We randomly sample one [BATCH, SEQ, FEALAB] data and input it into the neural network.

\begin{figure}[h]
	\begin{minipage}[t]{\linewidth}
		\centering
		\includegraphics[width=0.95\textwidth]{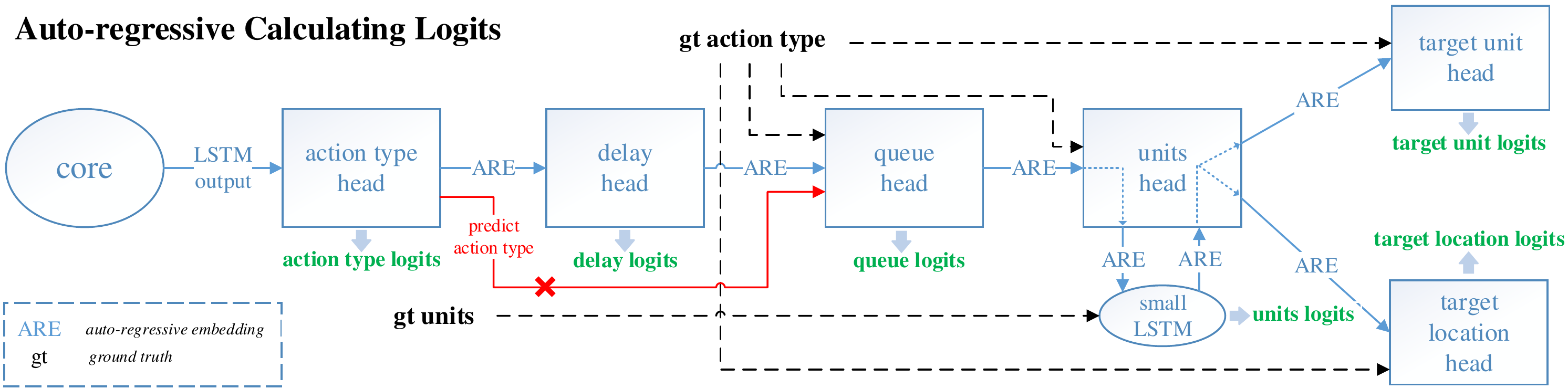}
		\caption{Supervised learning in mAS (auto-regressive calculating). We use the ground truth (label) instead of the predicted value (output) of the ``action type'' head to input into the ``queue'' head. This way is repeatedly used for the other action argument heads. The ``units'' (alias for ``selected units'') head is special and contains a small LSTM chain in its calculating pathway. Hence, except for inputting the ground truth of action types, we also input the ground truth of the selected units to calculate the units logits in the ``units'' head's pathway. The green logits are the outputs of these heads.}
		\label{fig: ARElogits}
	\end{minipage}
\end{figure}

We train based on the tensor in the second step. Unlike computing all heads' logits at once and then calculating the loss, we use the auto-regressive training ways in NLP (inspired by DI-Star~\cite{opendilab2021distar}). That is, we use the label (ground truth) instead of the output as the previous head's output when computing each head's logits (see Fig.~\ref{fig: ARElogits}). This processing can increase the accuracy of each head from an average of $0.2$ to about $0.7$. We will also use this auto-regressive training way in the RL loss calculation.

\begin{figure*}[t]
    \centering
    \subfloat[SU Loss]{
        \centering
        \includegraphics[width=0.230\columnwidth]{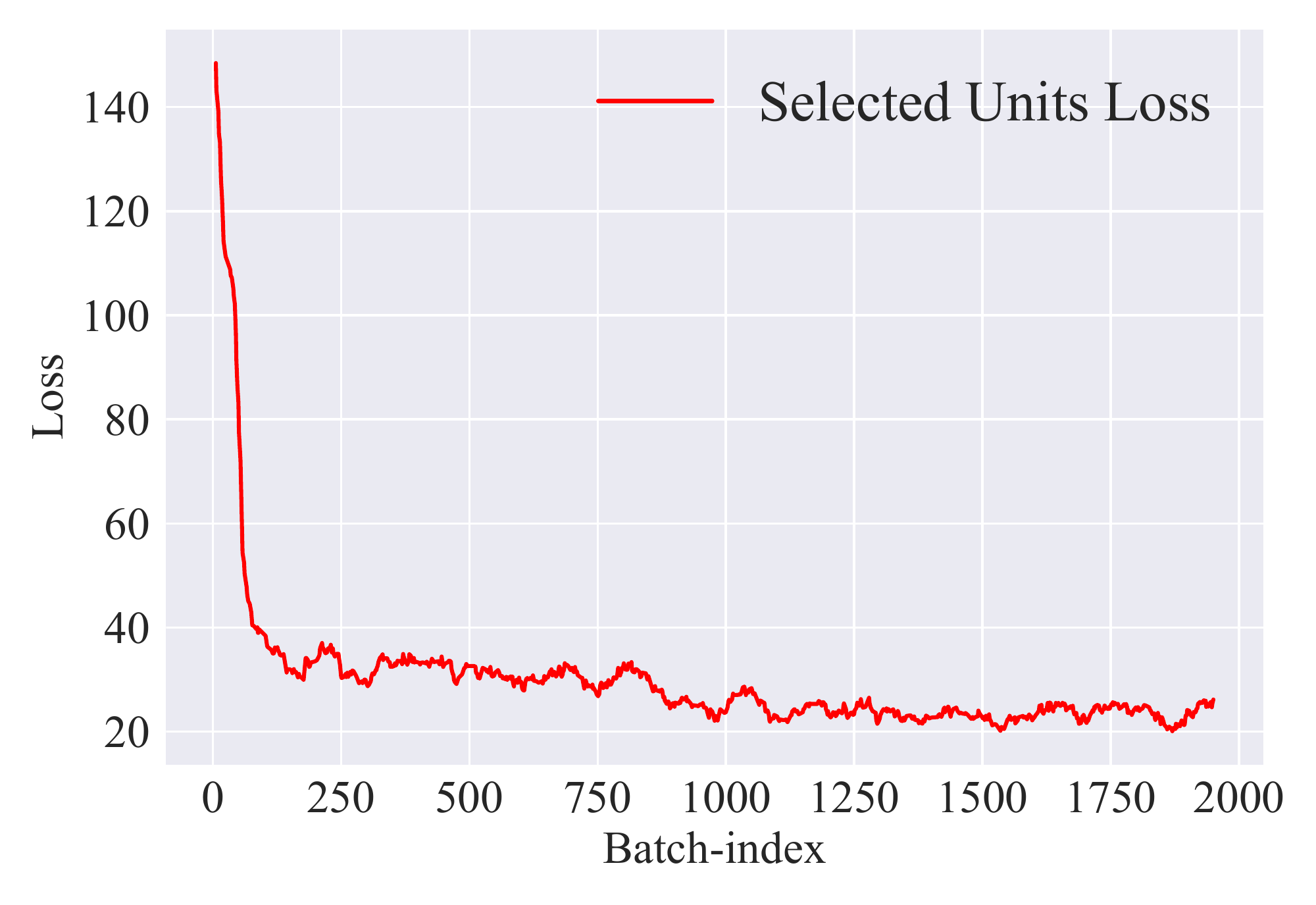}
		\label{fig:s.n.9.1}
   }
    \subfloat[SU Accuracy]{
        \centering
        \includegraphics[width=0.230\columnwidth]{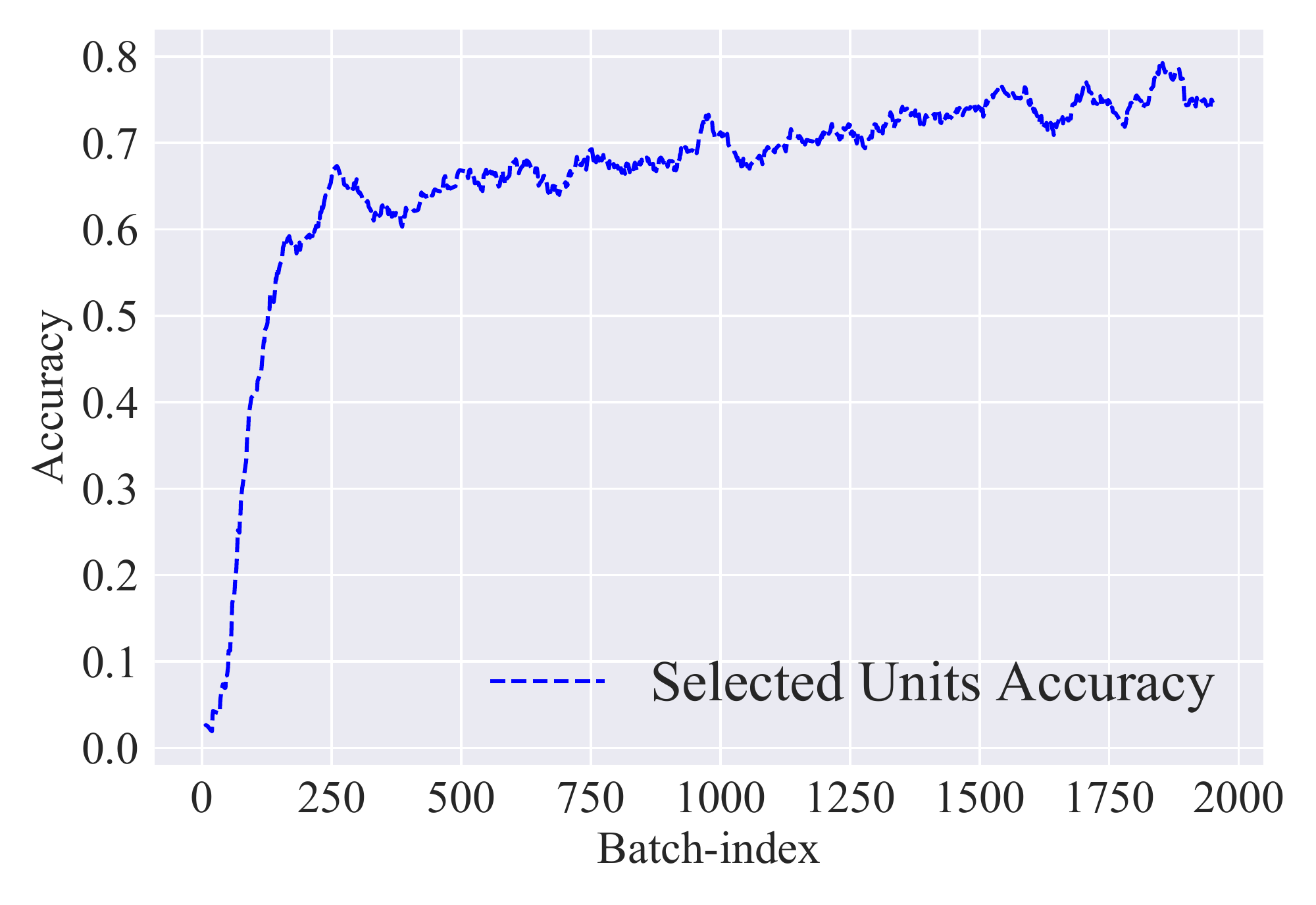}
		\label{fig:s.n.9.2}
    }
    \subfloat[SU Num Acc]{
        \centering
        \includegraphics[width=0.230\columnwidth]{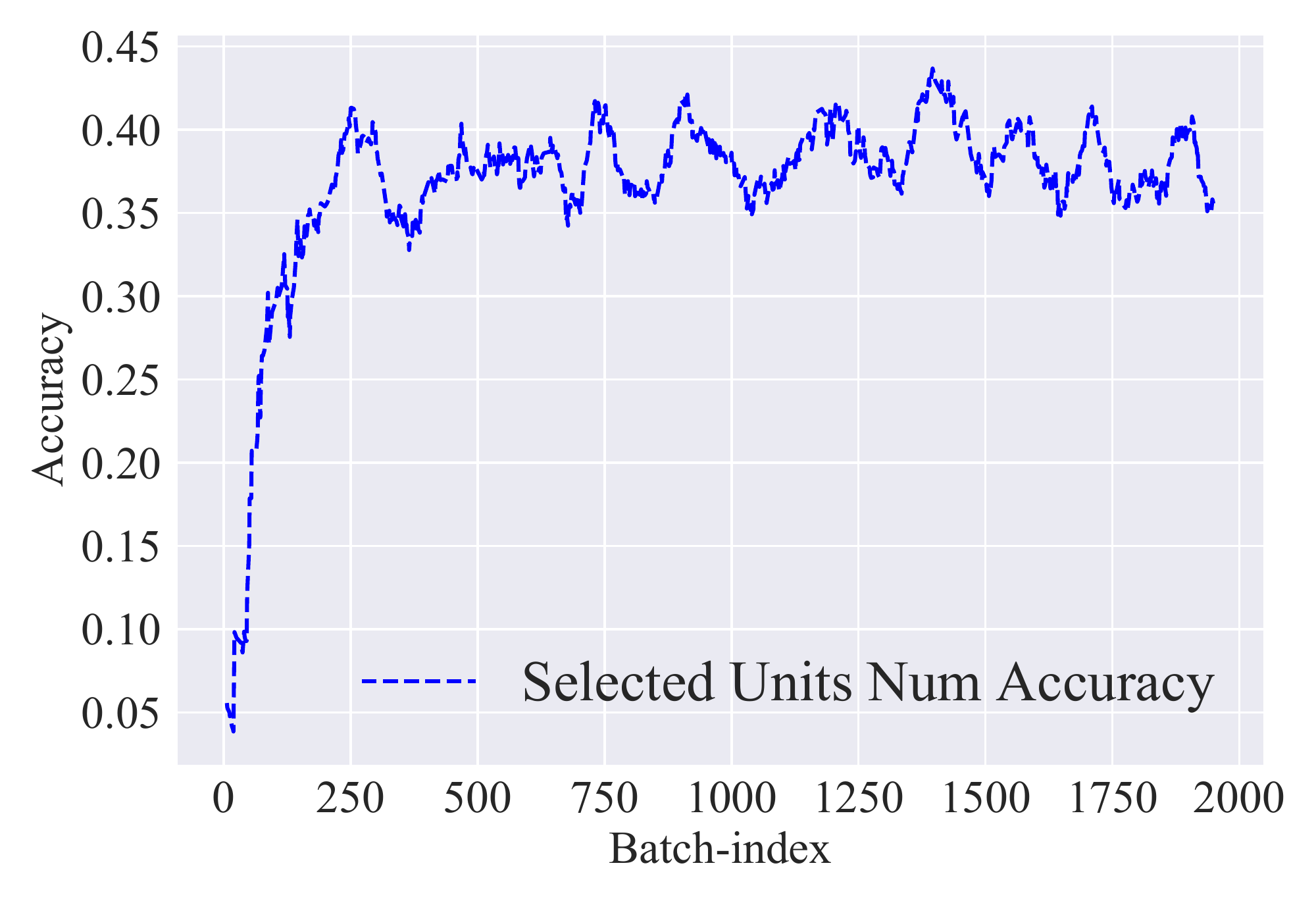}
		\label{fig:s.n.9.3}
   }	
   \subfloat[SU Type Acc]{
		\centering
		\includegraphics[width=0.230\columnwidth]{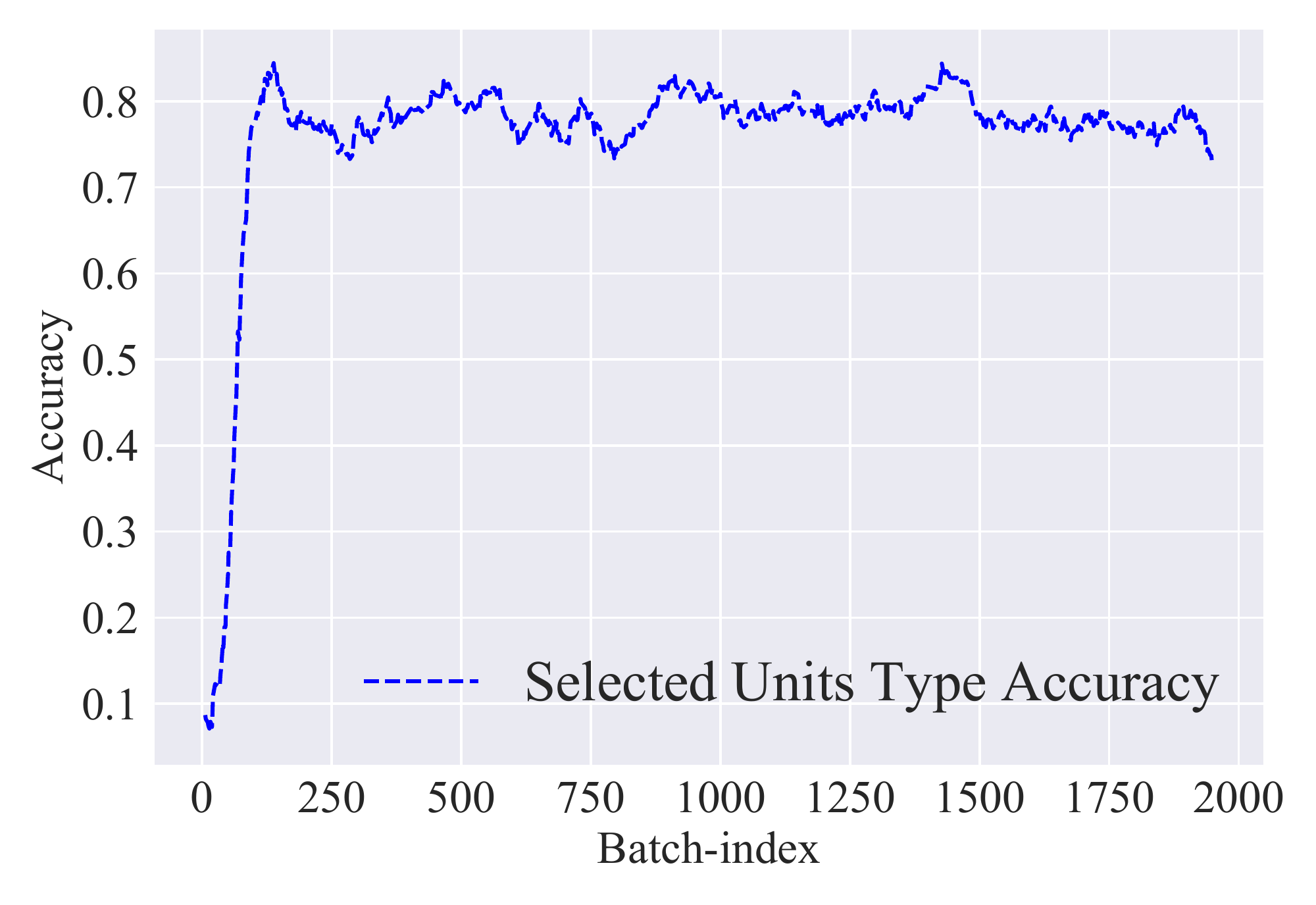}
		\label{fig:s.n.9.4}
	}	
    \caption{Training curves of the \textbf{selected units} (SU) head. (a): the loss curve of the SU head. (b): the accuracy curve of the SU head (only if the predicted unit tag of the selected units is the same as ground truth, it is considered correct). (c): the number accuracy curve of the SU head (compare the number of units). (d): the unit type accuracy curve of the SU head (compare the unit type of units).}
    \label{fig:s.n.a.9.1}
\end{figure*}

In the third step, we use accuracy and some other metrics to assess the model's performance because loss didn't intuitively evaluate the performance. However, mAS's action is composed of different action arguments. We should use different metrics for each action argument. E.g., for the ``selected units'' head, we use identifier accuracy, number accuracy, and type accuracy (the meaning of these metrics are shown in Fig.~\ref{fig:s.n.a.9.1}) to evaluate its performance fully.

We evaluate the model in the RL environment in the fourth step. This is due to two reasons: 1. The SL trained model has a natural ``domain shift" to the RL environment. The model of high SL accuracy may not behave well in the RL domain; 2. SL training aims to provide a good RL initial model. Hence we should evaluate the model in the final RL environment.

\begin{table}[h]
    \centering
    \scalebox{0.95}{
    \begin{tabular}{l | c | c c c c c}
    \toprule
	LR, Epochs & action acc. & win rate & used food & economy & development & military \\
    \midrule
	1e-4, 8  & 0.19 & \textbf{0.05} & \textbf{68.0} & \textbf{9282.6} & \textbf{5247.5} & \textbf{1033.8} \\
    1e-4, 100  & \textbf{0.93} & 0.00  & 13.7 & 4686.5 & 733.8 & 30.0 \\
    \bottomrule
    \end{tabular}
    }
    \caption{Lower SL action accuracy instead gets higher performance in the RL environment.}
    \label{tab: SL model in RL task}
\end{table}

In the fifth step, we use several metrics to evaluate the SL model's performance in the RL environment. These metrics include win rate, used food, and Blizzard scores. The Blizzard scores assess the agent from the economy, development, and military (also called killed points). Table~\ref{tab: SL model in RL task} shows a case in which is that a lower action accuracy gets a better RL performance than a higher action accuracy. This case verifies that the evaluation in the RL environment is necessary. If the agent does not behave well in the RL environment, the reason may be the unbalanced weight in the actions, insufficient data, or overfitting of training. We will finetune these hyper-parameters to start a new training (back to the first step).


\begin{figure*}[t]
    \centering
    \subfloat[Sum of Loss]{
        \centering
        \includegraphics[width=0.230\columnwidth]{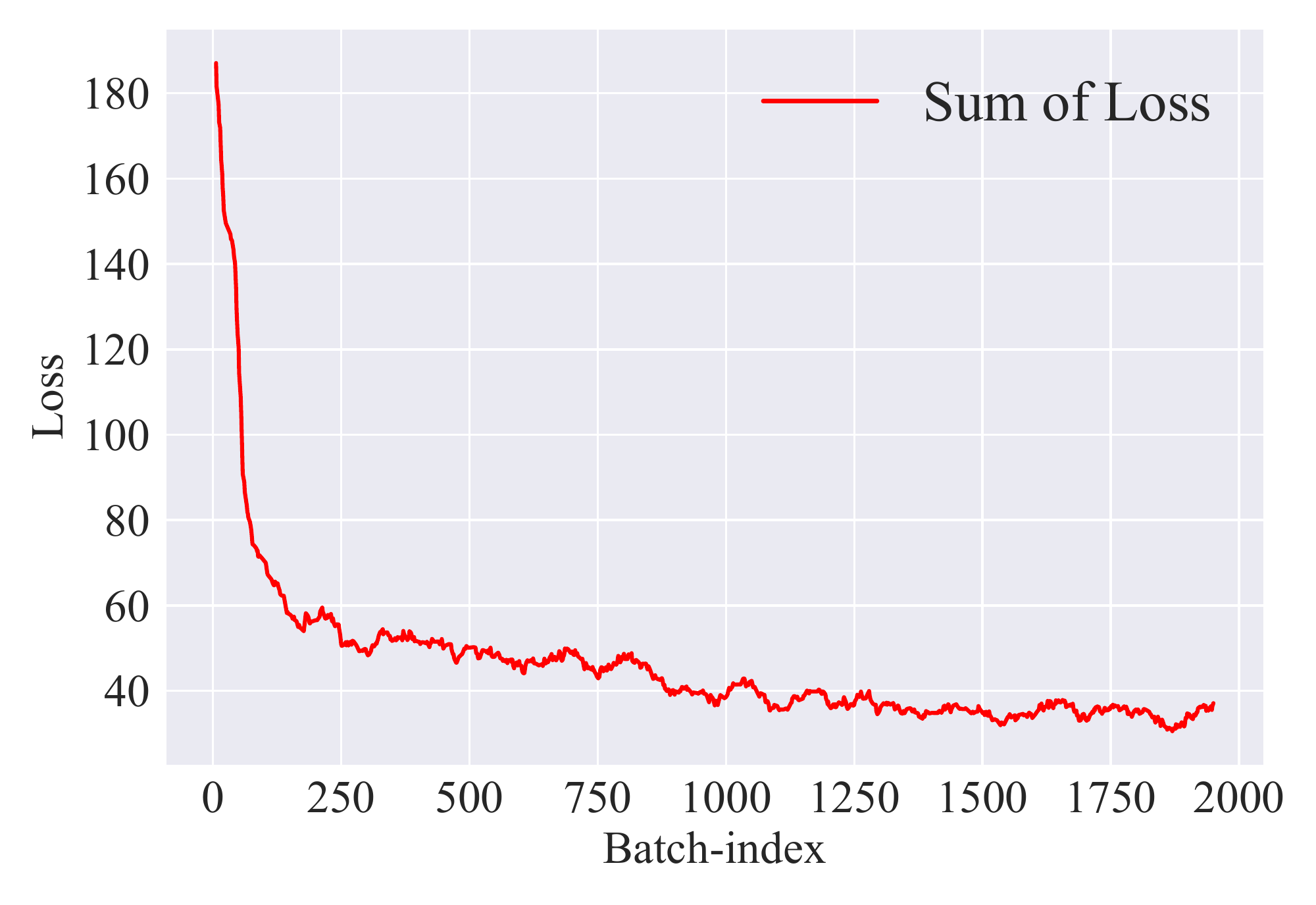}
		\label{fig:s.n.9.9}
   }
    \subfloat[AT Loss]{
        \centering
        \includegraphics[width=0.230\columnwidth]{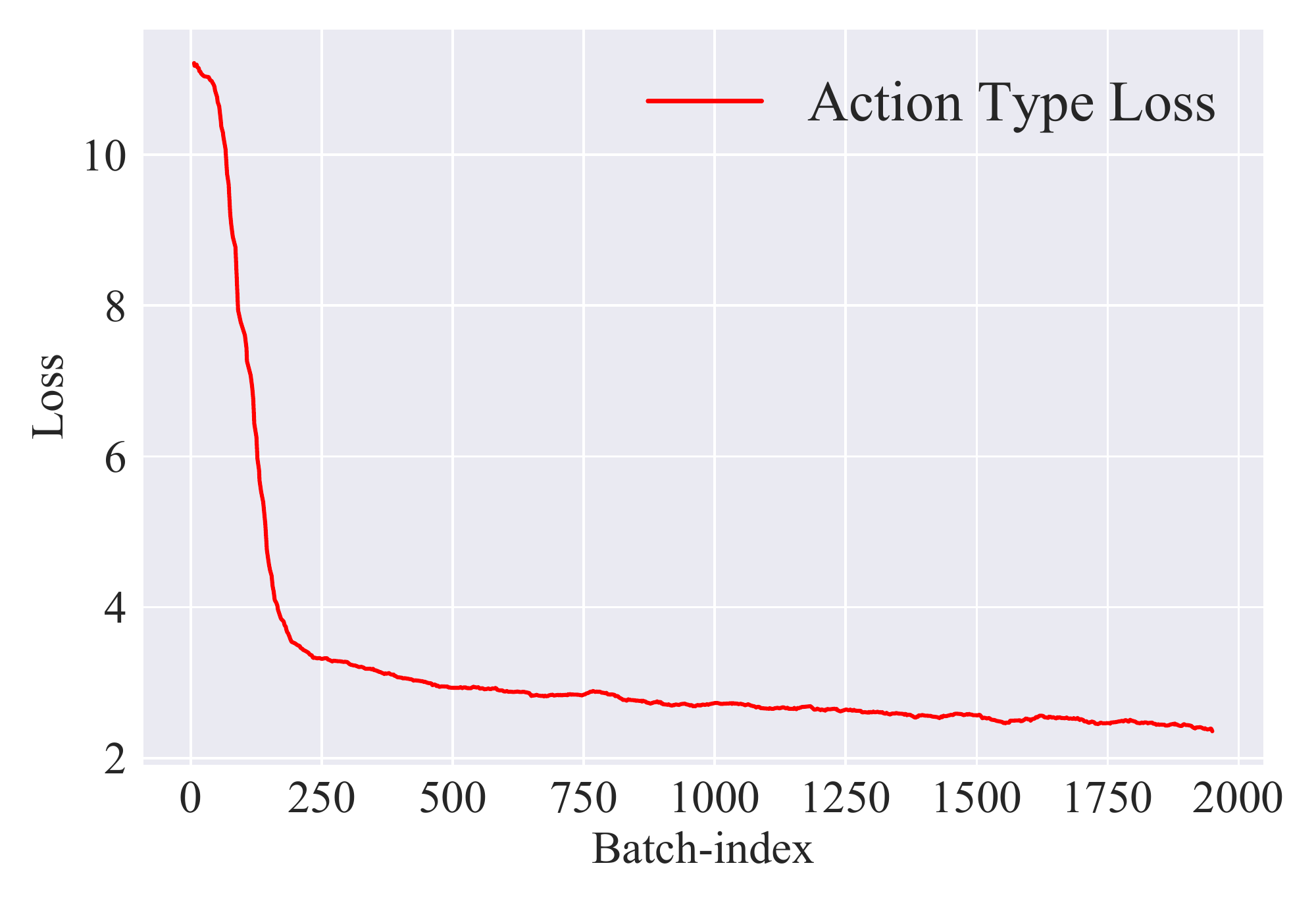}
		\label{fig:s.n.9.10}
    }
    \subfloat[AT Accuracy]{
        \centering
        \includegraphics[width=0.230\columnwidth]{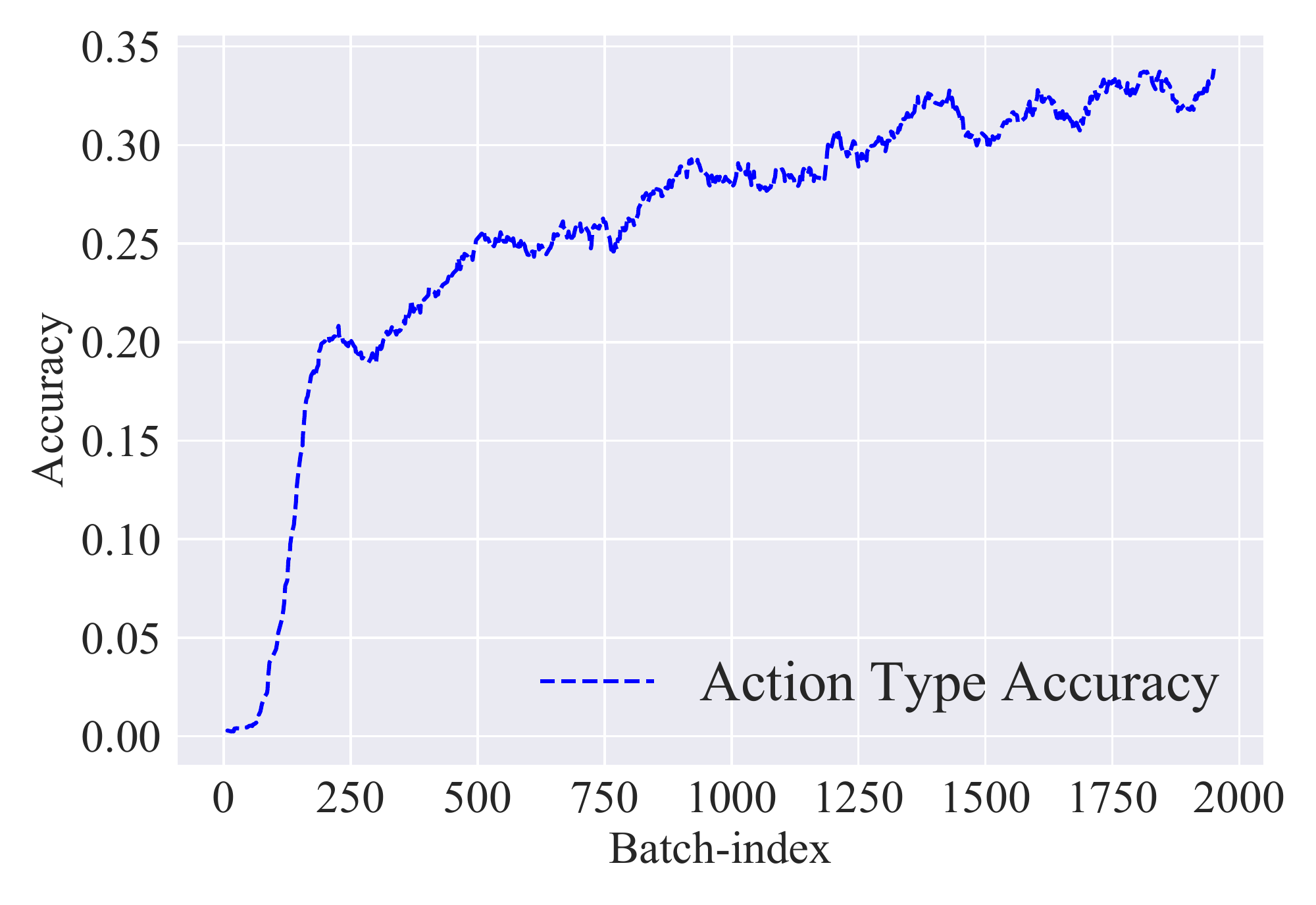}
		\label{fig:s.n.9.11}
   }	
   \subfloat[Imp. AT Acc]{
		\centering
		\includegraphics[width=0.230\columnwidth]{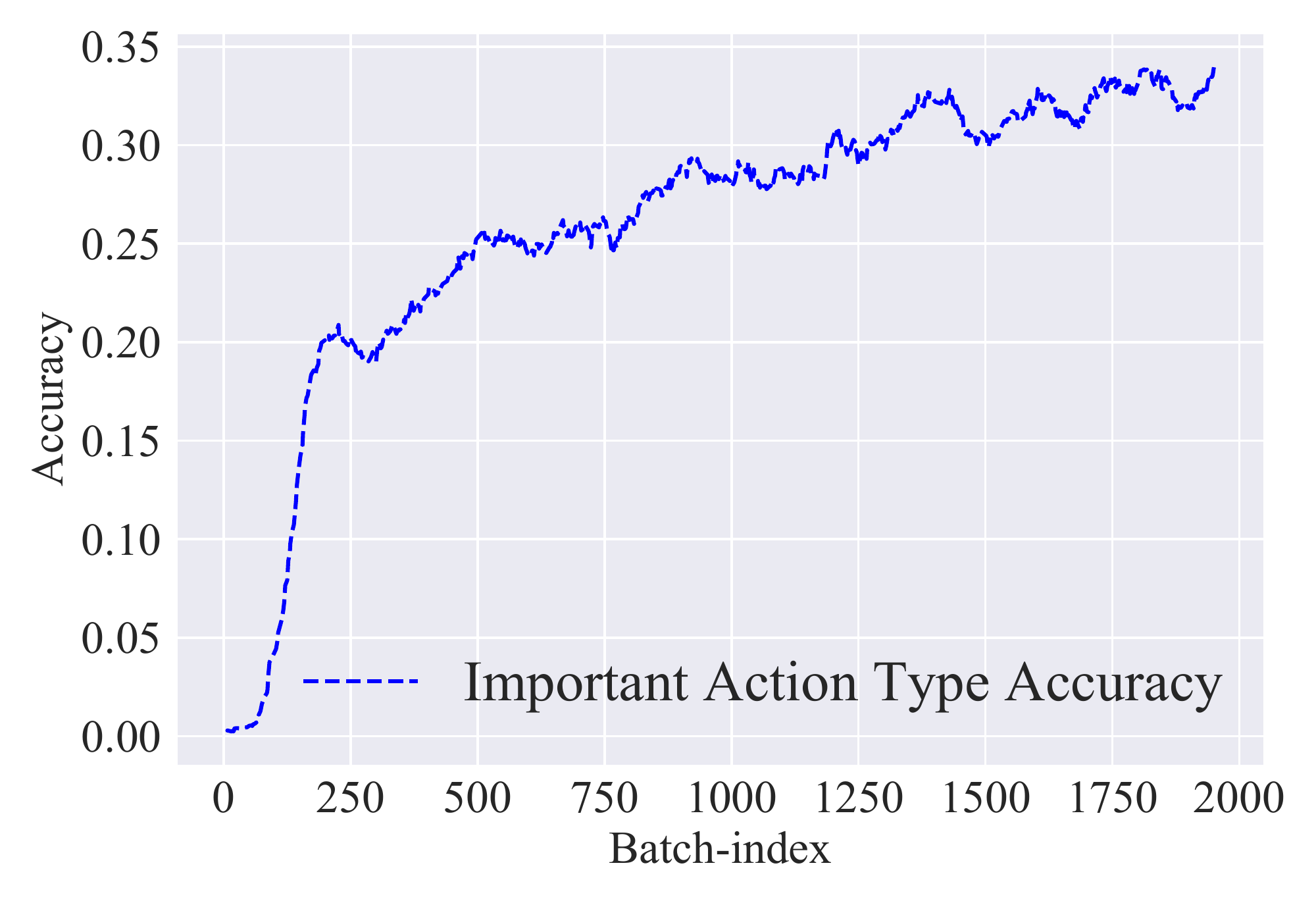}
		\label{fig:s.n.9.12}
	}	
    \caption{Training curves of all losses and \textbf{action type} (AT) head. (a): the curve of the sum of all losses (AT and five action arguments). (b): the loss curve of the AT head. (c): the accuracy curve of the AT head. (d): the accuracy curve of some important actions (such as building Pylon or training Zealot) in the AT head.}
    \label{fig:s.n.a.9.3}
\end{figure*}


\begin{figure*}[t]
    \centering
    \subfloat[TL Loss]{
        \centering
        \includegraphics[width=0.230\columnwidth]{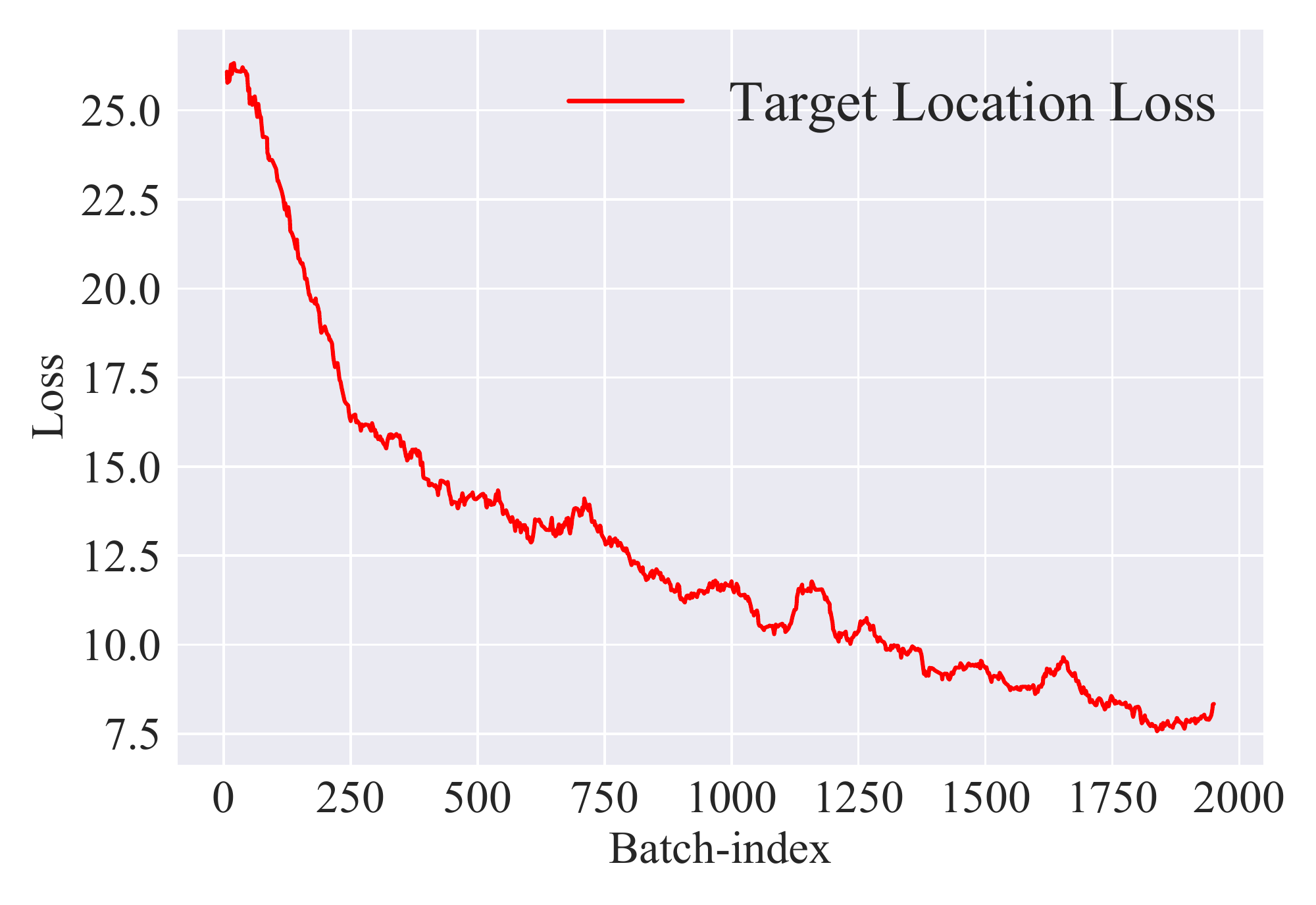}
		\label{fig:s.n.9.5}
   }
    \subfloat[TL Accuracy]{
        \centering
        \includegraphics[width=0.230\columnwidth]{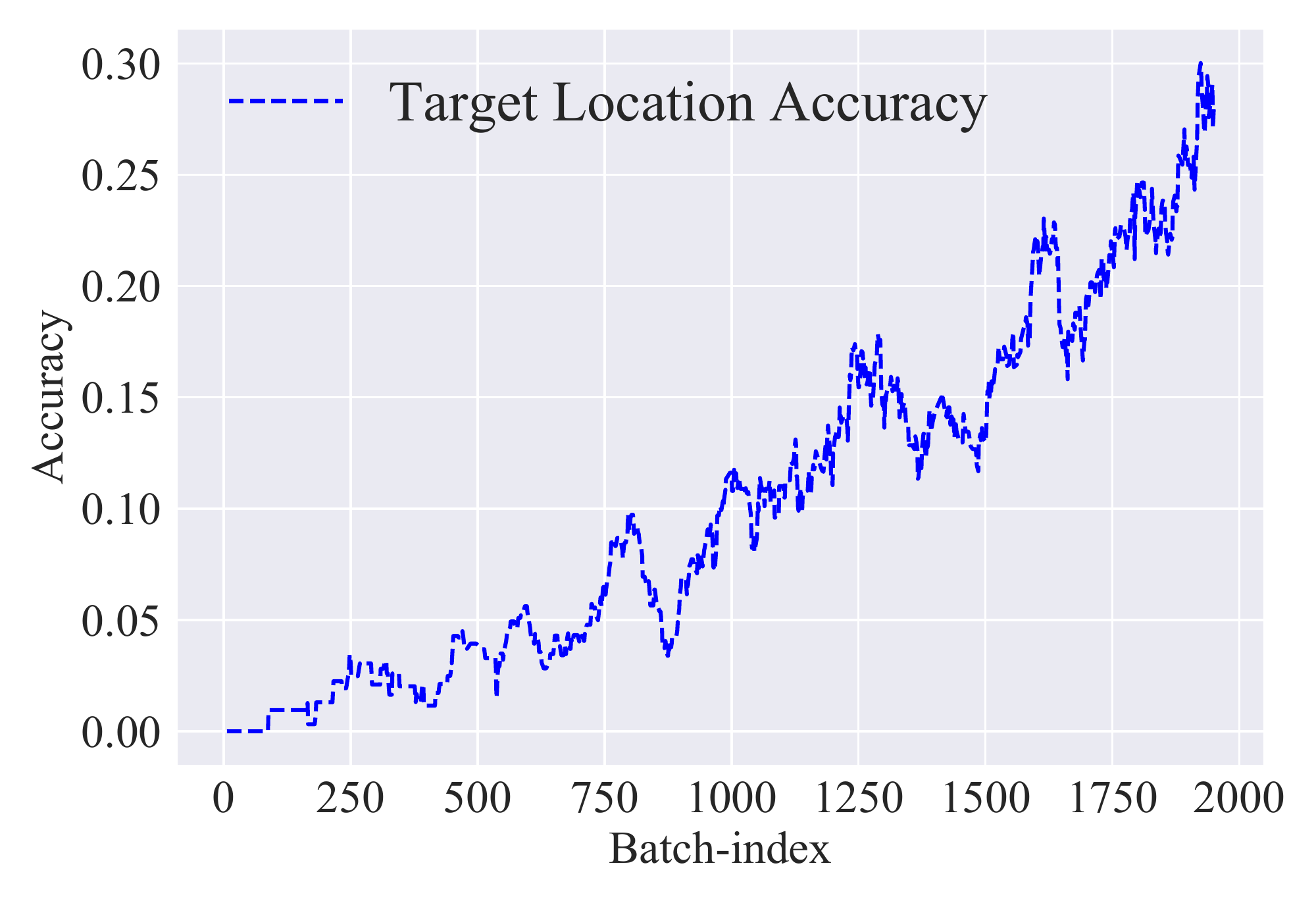}
		\label{fig:s.n.9.6}
    }
    \subfloat[TL Distance]{
        \centering
        \includegraphics[width=0.230\columnwidth]{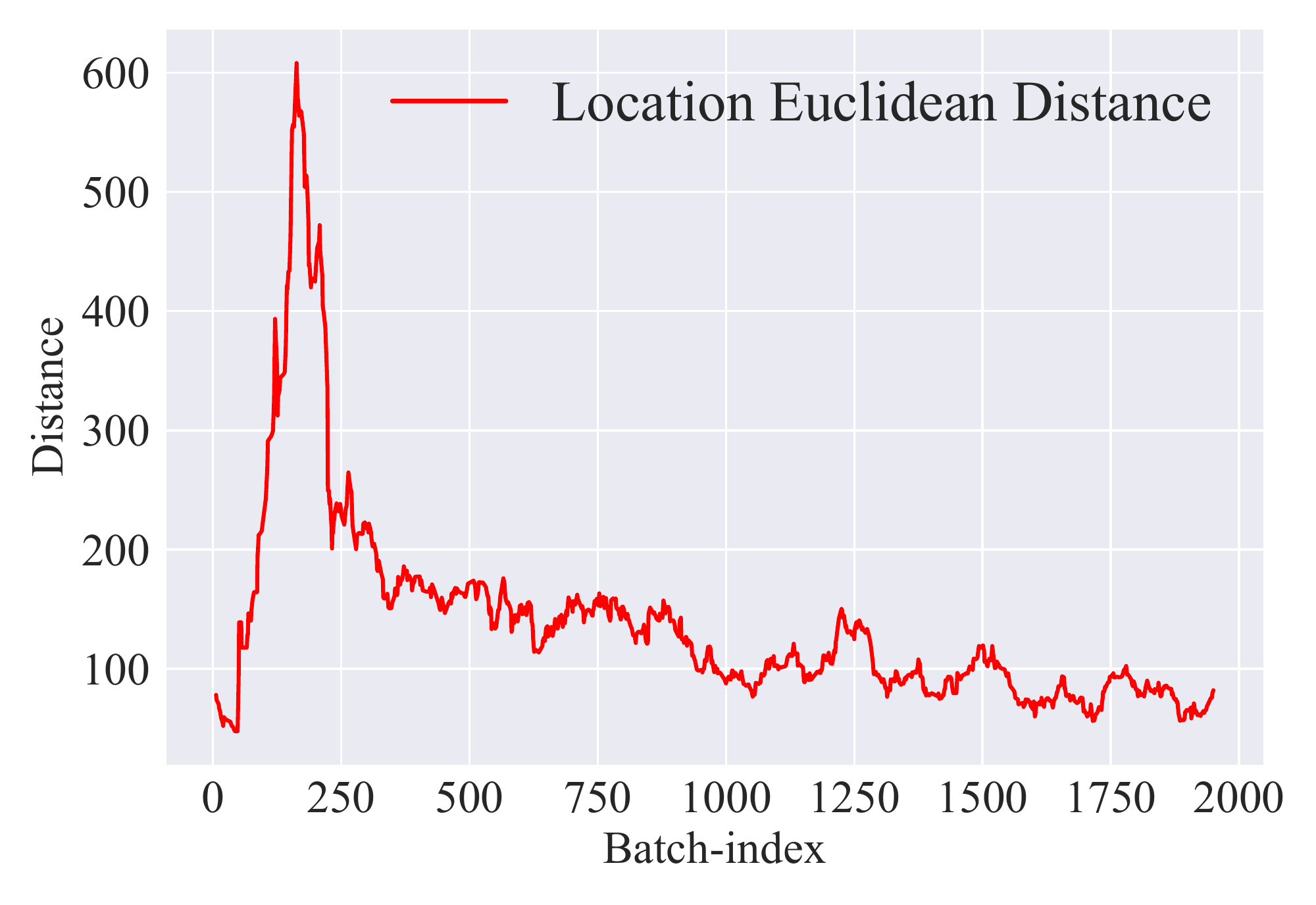}
		\label{fig:s.n.9.7}
   }	
   \subfloat[Queue Loss]{
		\centering
		\includegraphics[width=0.230\columnwidth]{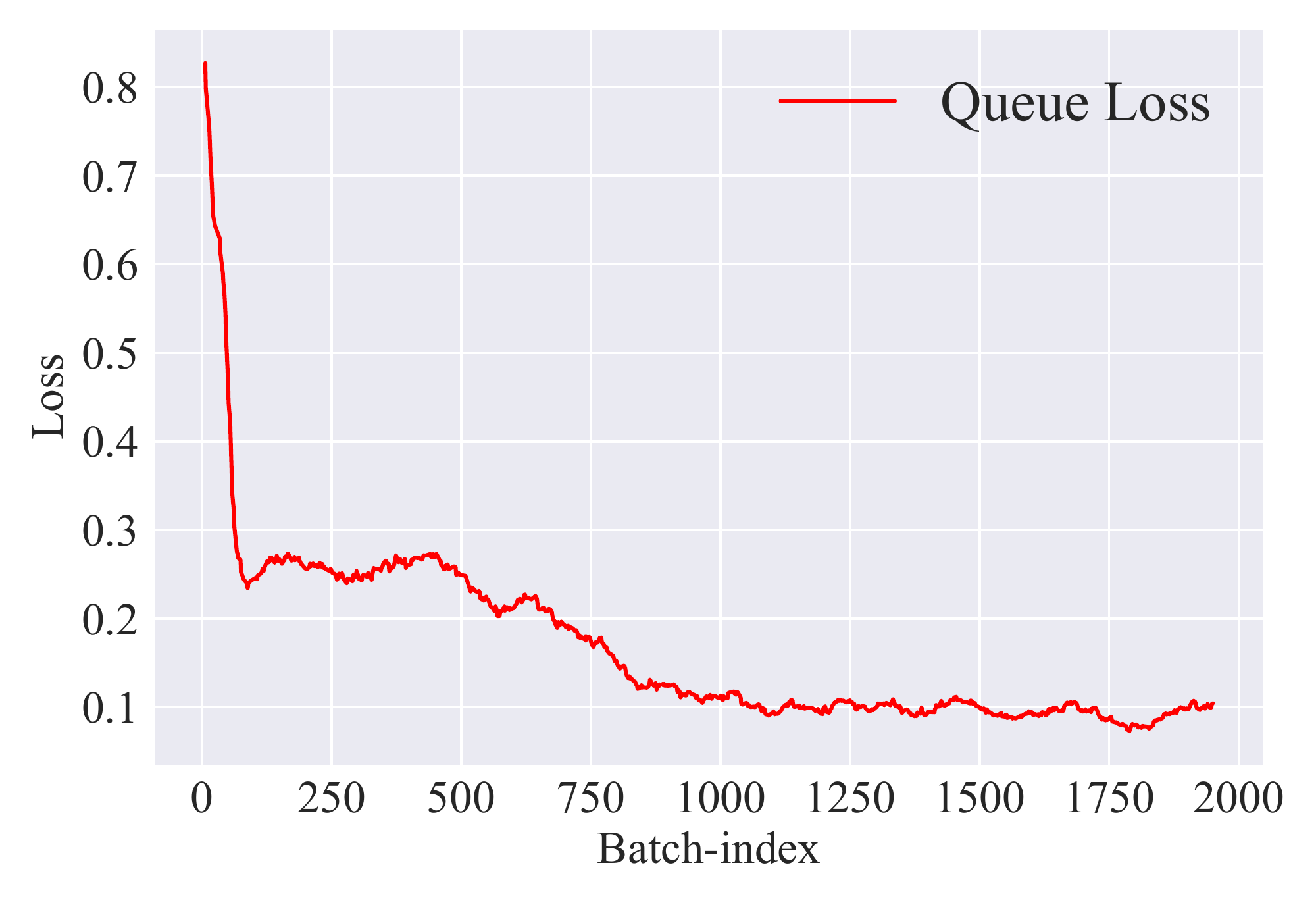}
		\label{fig:s.n.9.8}
	}	
    \caption{Training curves of the \textbf{target location} (TL) head and the queue head. (a): the loss curve of the TL head. (b): the accuracy curve of the TL head. (c): the training curve of the Euclidean distance between the predicted location and the ground truth. (d): the loss curve of the queue head.}
    \label{fig:s.n.a.9.2}
\end{figure*}

\begin{figure*}[h]
    \centering
    \subfloat[At 7:30]{
        \centering
        \includegraphics[width=0.45\columnwidth]{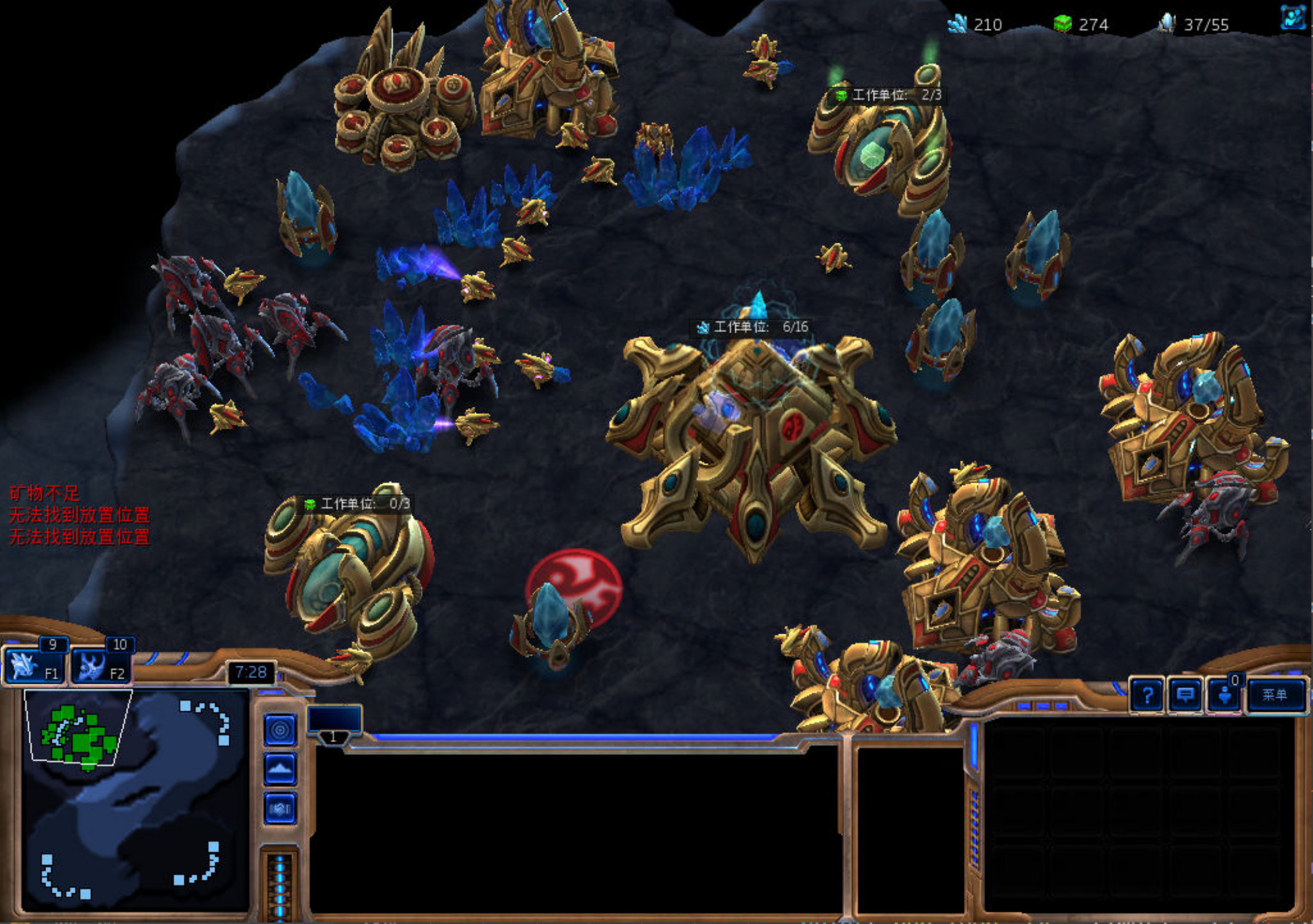}
		\label{fig:s.n.11.3}
   }
    \subfloat[At 14:00]{
        \centering
        \includegraphics[width=0.45\columnwidth]{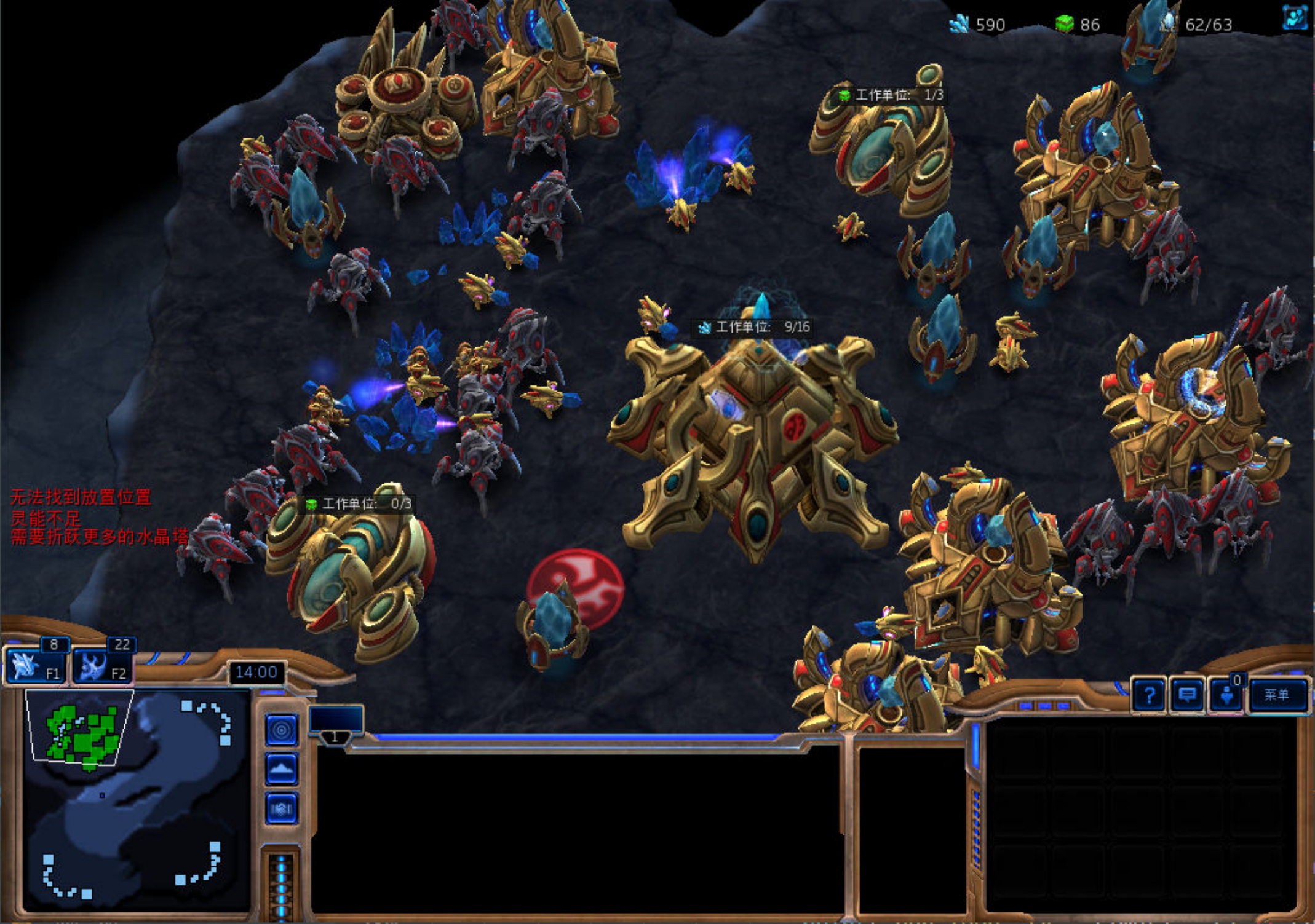}
		\label{fig:s.n.11.4}
    }%
    \caption{\textbf{The training results of the mAS's supervised learning}. (a): at the middle (7:30) of the game. (b): at the end (14:00) of the game. It can be seen that the trained agent can select units to collect resources, build buildings in the proper position, and select Gateways to train soldiers (Zealots and Stalkers) continually.}
    \label{fig:s.n.a.11.2}
\end{figure*}

In SL training, we found a learning rate of 1e-4 and 10 training epochs achieve the best result. The best model achieves a $0.15$ win rate against the level-1 built-in AI. Note that though this result is not as good as that we acquire in the HRL method, the training here faces 564 actions, thus is much difficult. The 1e-4 learning rate is also selected by experiments and is different from the default 1e-3 in the AlphaStar pseudocodes. We find that training more than 10 epochs will easily fall in overfitting, making the agent can't do any meaningful things. Table.~\ref{tab: mAS compare SL in all} compares training results with different learning rates and epochs (average over 20 games). Fig.~\ref{fig:s.n.a.9.3}, \ref{fig:s.n.a.9.1}, and \ref{fig:s.n.a.9.2} show the training curves (including loss, accuracy and other metrics) of the SL process. Note we use red curves to represent loss (the lower, the better), and blue curves to represent accuracy (the higher, the better). Fig.~\ref{fig:s.n.a.11.2} shows the SL's training effect \footnote{The videos of supervised learning results of the mini-AlphaStar agents can be found on Youtube at \url{https://youtu.be/mTtA0vdAULw}}.


\subsection{Simplifications}
This subsection introduces how we make AlphaStar's structure can be trained in one single server while maintaining its potential of power. AlphaStar is composed of components that are essential for its learning ability. We sustain all of its components but reduce their size by a fixed scale. E.g., if one linear layer has 256 hidden units, we make the same layer in mAS have 64 hidden units, such the rescale ratio is 1/4. We make the same scaling for other layers as possible. The rescaling can also apply to other hyper-parameters. E.g., the feature map input into AlphaStar is 128x128, while we only use 64x64 (which made the length and width of these maps divided by 2, in order to reduce the needed memory). However, some hyper-parameters can not be reduced (doing so will cause a degradation of performance). We explain these hyper-parameters here:
\begin{itemize}
    \item  \textbf{Max\_entities}. This value controls how many entities AlphaStar can perceive in one game step. The entities here are the same meaning as units, including buildings and moveable units. The entities consist of ours, enemies (with only the ones in our visible areas), and neutrals. The default value in AlphaStar is 512. Considering the larger number of units we can build in one game, it is reasonable. However, in extreme cases, this value may also be not enough, such as both of enemies and us make hundreds of Zerglings, and they all fight in the same area. Decreasing max\_entities can significantly affect the performance. Because the ``selected units head'' and ``target unit head'' all need these entities' information. If the value is set to 64, the 65th entities (and the ones after) can not be perceived by the AlphaStar. It can not even control them because the outputs of ``selected units head'' and ``target unit head'' are values between 0 and max\_entities (meaning the index of these entities). So max\_entities is an important value that can not be simplified. In mAS, we set it to the same value (512) as in the AlphaStar.
    \item \textbf{Max\_selected}. This hyper-parameter is used only in ``selected units head'', which judge the number of the most units we can select at one time in AlphaStar. This hyper-parameter is set to 64 in AlphaStar, which is reasonable because, in one SC2 game, we don't select units more than 64 at one step most time. Decreasing this value can also affect performance. In extreme cases, such as the value is 1, we can only select one unit at a time. In that case, we can't control many combat units simultaneously to attack. However, larger values for this hyper-parameter also cost more memory. Hence, in mAS, we choose an intermediate value: 12, the same as the number of the most units we can set for one group in SC1.
    \item \textbf{BATCH} and \textbf{SEQ}. Due to the design of the network structure, setting the two hyper-parameters to any values will not affect the running of the programs. E.g., in the inference time, we set both BATCH and SEQ to 1, getting one action at a time. If we want to get faster training in the training time, we should set these hyper-parameters to more than dozens. However, larger BATCH and SEQ also cost much more GPU memory. These two values should be decided according to the specific available resources. In mAS, the default settings of these two values are 12 and 12.
\end{itemize}
For the hyper-parameters of others, such as the reduced scale ratios of the specific layers in the neural network, we recommend the reader refer to the technical report~\cite{liu2021mASarxiv}.

\begin{table}[t]
    \centering
    \scalebox{1.0}{
    \begin{tabular}{l | c c c c}
    \toprule
	LR \textbackslash Epochs & 5 & 10 & 15 & 20 \\
    \midrule
	1e-3  & 0 & 2981.2 & 325.0 & 30.0 \\
    1e-4  & - & \textbf{3567.5} & 1208.8 & 1577.5 \\
	\midrule
	\midrule
	LR \textbackslash Epochs & 5 & 10 & 15 & 20 \\
    \midrule
    1e-3  & 16.8 & 25.4 & 7.5 & 5.8 \\
	1e-4  & - & \textbf{89.0} & 30.4 & 48.2 \\
    \bottomrule
    \end{tabular}
    }
    \caption{The two metrics (\textit{killed points} (on top) and \textit{used food} (at bottom)) trained by different hyper-parameters. LR=learning rate.}
    \label{tab: mAS compare SL in all}
\end{table}


\subsection{Reinforcement Learning in Mini-AlphaStar}
In this subsection, we present the settings and training process of the RL of mini-AlphaStar. For mAS, the DNN architecture used in RL is the same as in SL. We train the RL agent based on the model trained in the SL process, fine-tuning the model's parameters by a smaller learning rate. The state features of RL contain three parts, the features of all units, the feature maps of the game maps, and the features of statistic scalars. RL's action space is the RAS of PySC2$_3$. Note that this differs from the HAS we used in the hierarchical approach and differs from the constructed macro actions. E.g., the number of our constructed macro actions is dozen. On the contrary, the RAS has a total of $564$ actions. Note the state and action are called features and labels in SL. 

Our hierarchical approach uses a fixed decision frequency that takes one action every 1 second. In contrast, AlphaStar uses a dynamic decision frequency which means that the interval times between two actions may not be the same. AlphaStar uses the Delay head to output a delay value $d$, telling the agent to do the next action after $d$ game steps. The Delay head gives freedom for the time for the action decision. Our mini-AlphaStar also implements this feature. However, in mAS's experiment, we found the training using dynamic decision frequency (based on the output from the Delay head) is not well. On the contrary, using a fixed delay (e.g., set step\_mul = 8) achieves better results. Hence in the following training, we set the weight of the Delay loss to 0 in the loss calculation.

The mAS provides two RL training ways: self-play training and training against built-in AIs. In self-play, the opponent is a snapshot of the historical agents. Which snapshot will be selected is decided by the multi-agent league mechanism. The mAS supports the three types of players like in AlphaStar: main player, main exploiter, and league exploiter. Due to the training time constraint, we only use a game league containing one main player in this experiment. For the sake of comparing with our approach, in the following experiments, we mainly consider RL training against built-in AIs. We set the maximum game steps of one episode to be $18000$ steps. After that, if there is no winner, the game result is a draw. Simple64 is the game map. The race of the training agent is Protoss, and the race of the built-in AI is Terran.

\subsection{Reinforcement Learning Details}
We now give the details of the RL training process. First, we introduce the RL loss calculation of AlphaStar and mAS. The AlphaStar's RL loss is complex due to 4 reasons: 1. Its action outputs have 6 heads (one action type and 5 action arguments), each of which loss is calculated separately, multiplying the overall loss numbers by nearly 6 times; 2. Except for policy gradient (V-trace) loss and value (TD-lambda) loss, AlphaStar also introduces a UPGO loss, a KL distillation loss, and an entropy loss, which increases the number of losses; 3. Besides the win-loss reward, AlphaStar introduces a pseudo-reward to estimate how well the agent follows a human player's replay, which has the same race and is on the same map as the current game. AlphaStar introduces 4 other baselines for this pseudo-reward, making the number of the overall baselines to be 5; 4. The TD-lambda loss and V-trace loss are used for all 5 baselines, while the UPGO loss is used for the win-loss baseline. Except for the TD-lambda loss, all other losses are used for all action arguments. Hence AlphaStar's losses contain dozens of ones.

\begin{figure}[h]
	\begin{minipage}[t]{\linewidth}
		\centering
		\includegraphics[width=0.95\textwidth]{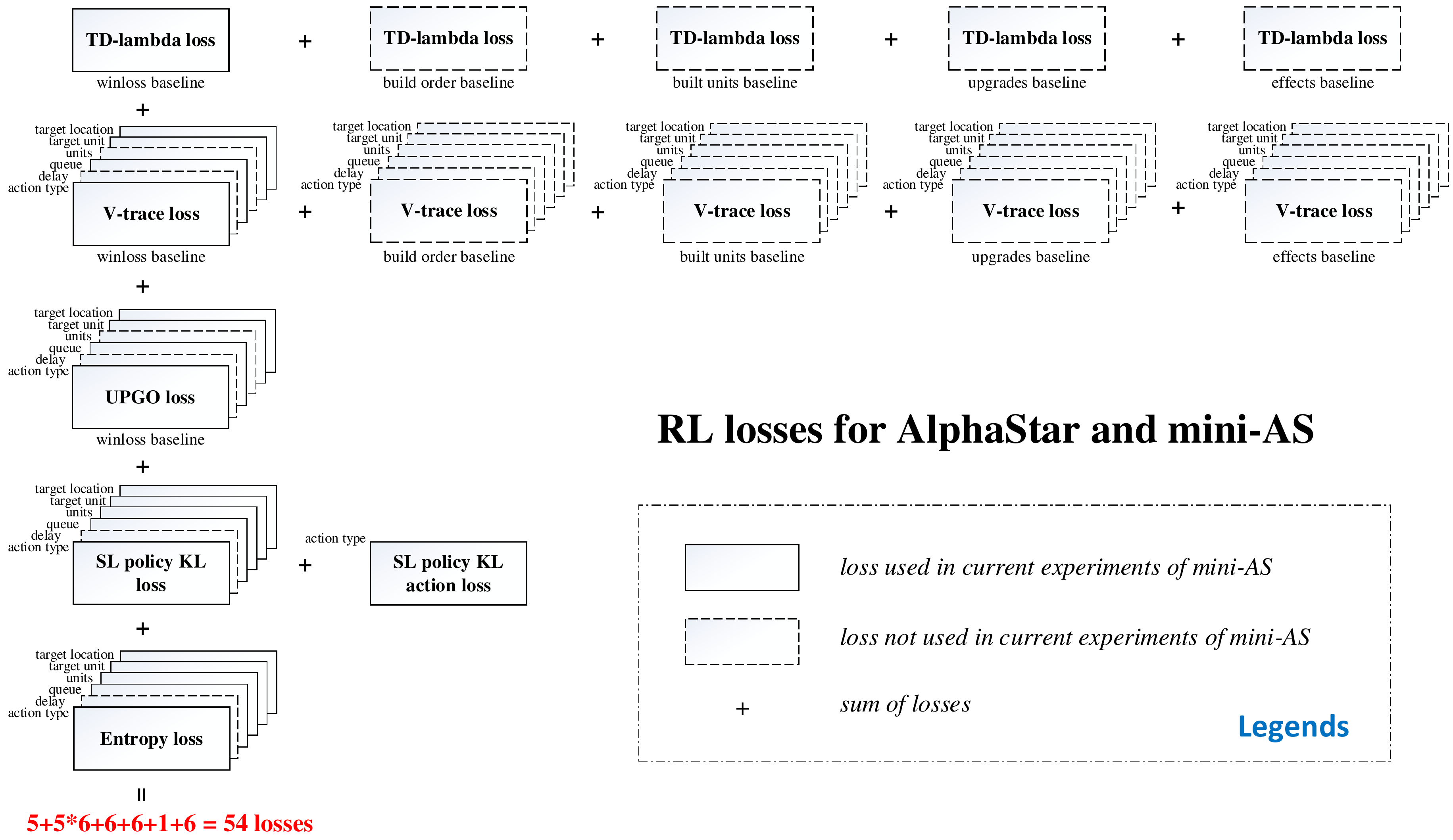}
		\caption{RL losses for AlphaStar and mAS (mini-AS).}
		\label{fig: RLloss}
	\end{minipage}
\end{figure}

The total of AlphaStar's losses are 5 TD-lambda losses, $5 * 6 = 30$ V-trace policy gradient losses, $1* 6 = 6$ UPGO losses, 6 KL distillation losses, 1 KL action type distillation loss (which only applies to actions types in every 4 first minutes of the game), and 6 entropy losses. The total number is: $5 + 30 + 6 + 6 + 1 + 6 = 54$ (illustration shows in Fig.~\ref{fig: RLloss}). To compare, our hierarchical approach contains 1 policy gradient loss, 1 value (baseline) loss, and 1 entropy loss, a total of 3 losses.

For simplicity, we make mAS only use the win-loss reward in the comparison experiments, which is similar to our setting because the hierarchical approach doesn't use replays in RL training. As in supervised learning, we don't use the Delay head's loss (making its weight be 0). We also find the loss for the ``units" head in V-trace and UPGO is hard to optimize in the RL phase. We will give the comparison experiment of training with/without units loss. However, the best training curve in our experiments doesn't use units loss. Therefore, the training losses of mAS consist of 1 TD-lambda loss, 4 V-trace policy gradient loss, 4 UPGO loss, 5 KL distillation loss, 1 KL action type distillation loss, and 5 entropy loss. The total number of losses is $1 + 4 + 4 + 5 + 1 + 5 = 20$. Notice this number of losses is still more than six times our hierarchical approach.


\begin{figure*}[h]
    \centering
    \subfloat[Value net]{
        \centering
        \includegraphics[width=0.230\columnwidth]{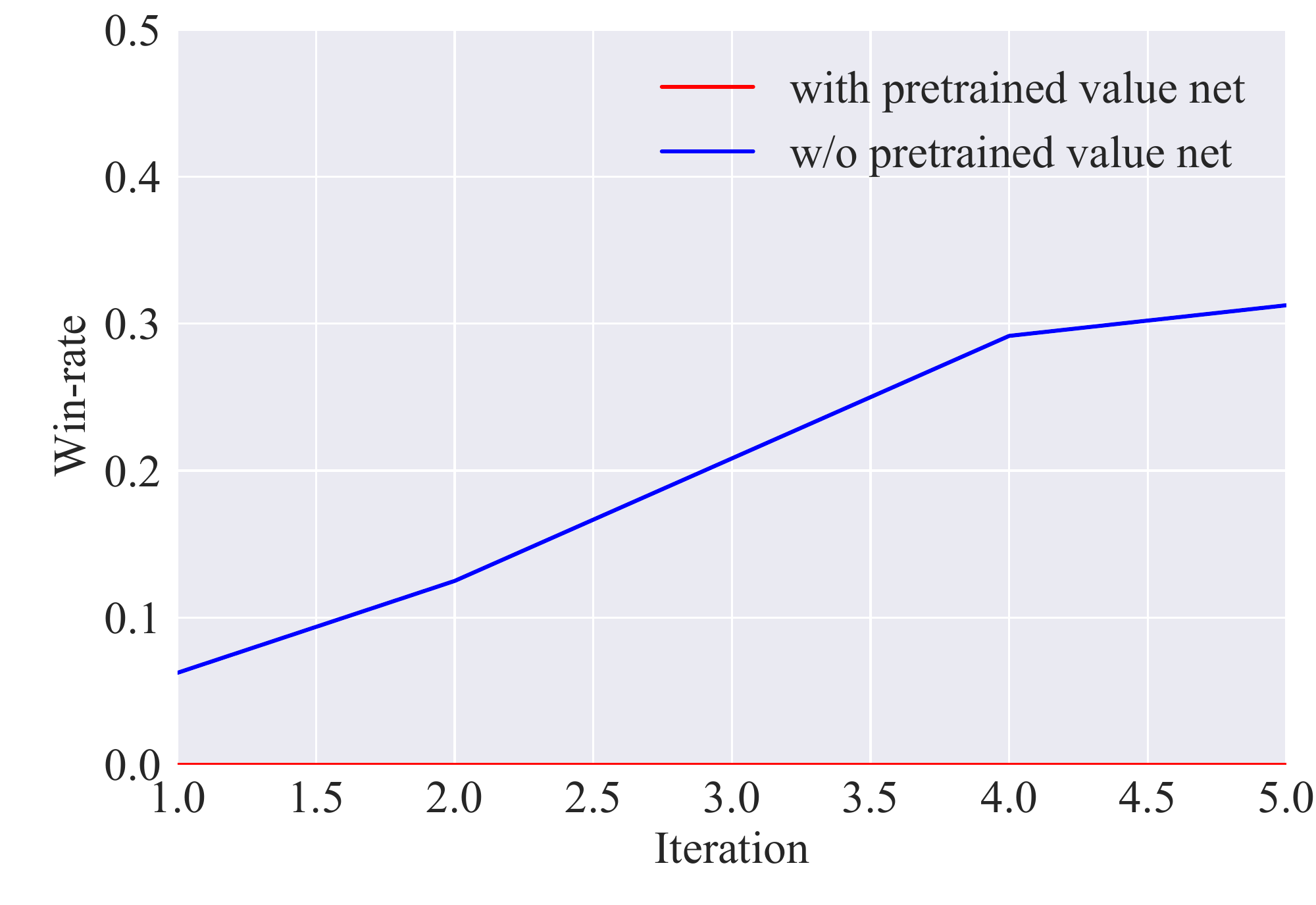}
		\label{fig:s.n.10.5}
   }
    \subfloat[Learning rate]{
        \centering
        \includegraphics[width=0.230\columnwidth]{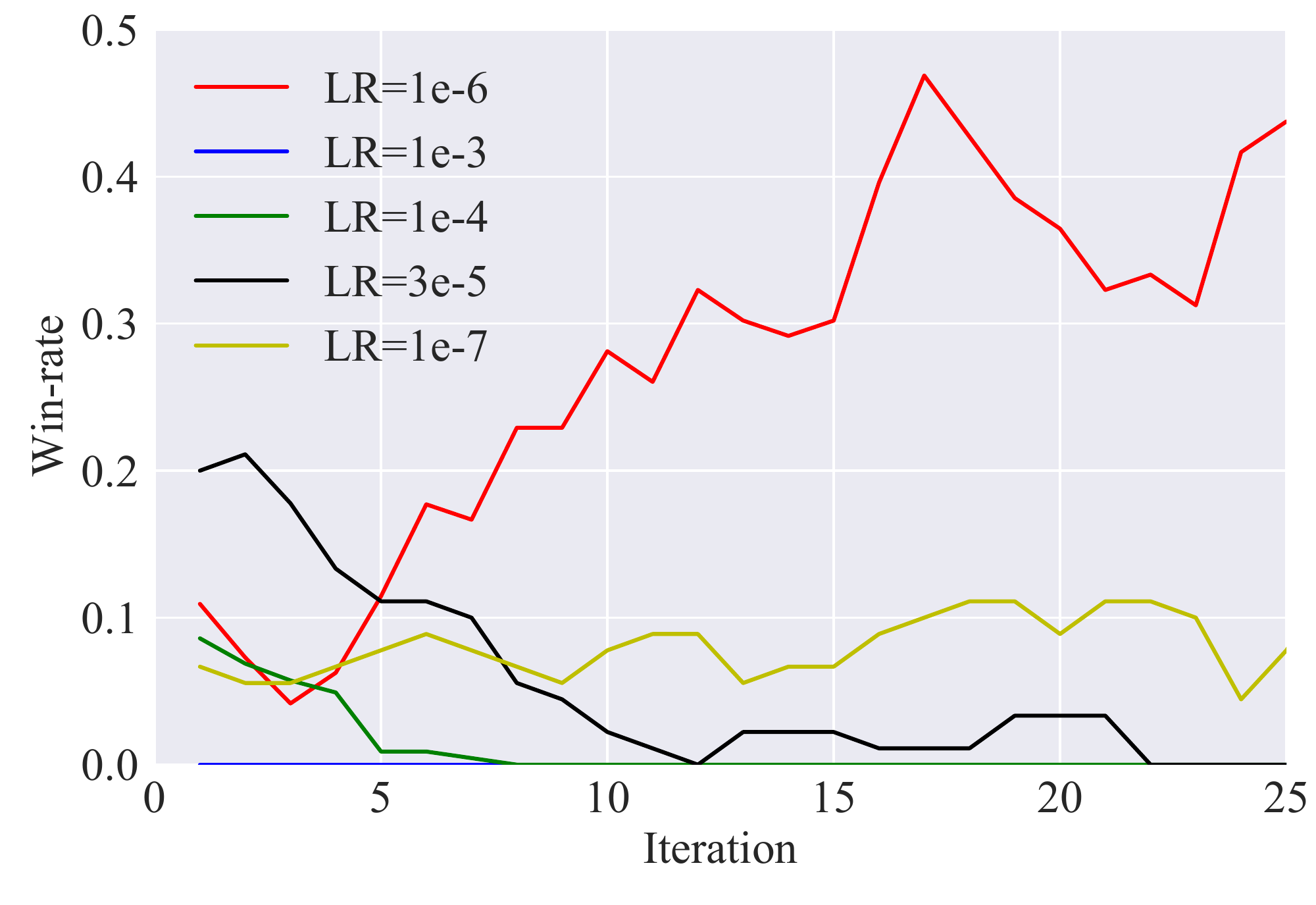}
		\label{fig:s.n.10.6}
    }
    \subfloat[Units loss]{
        \centering
        \includegraphics[width=0.230\columnwidth]{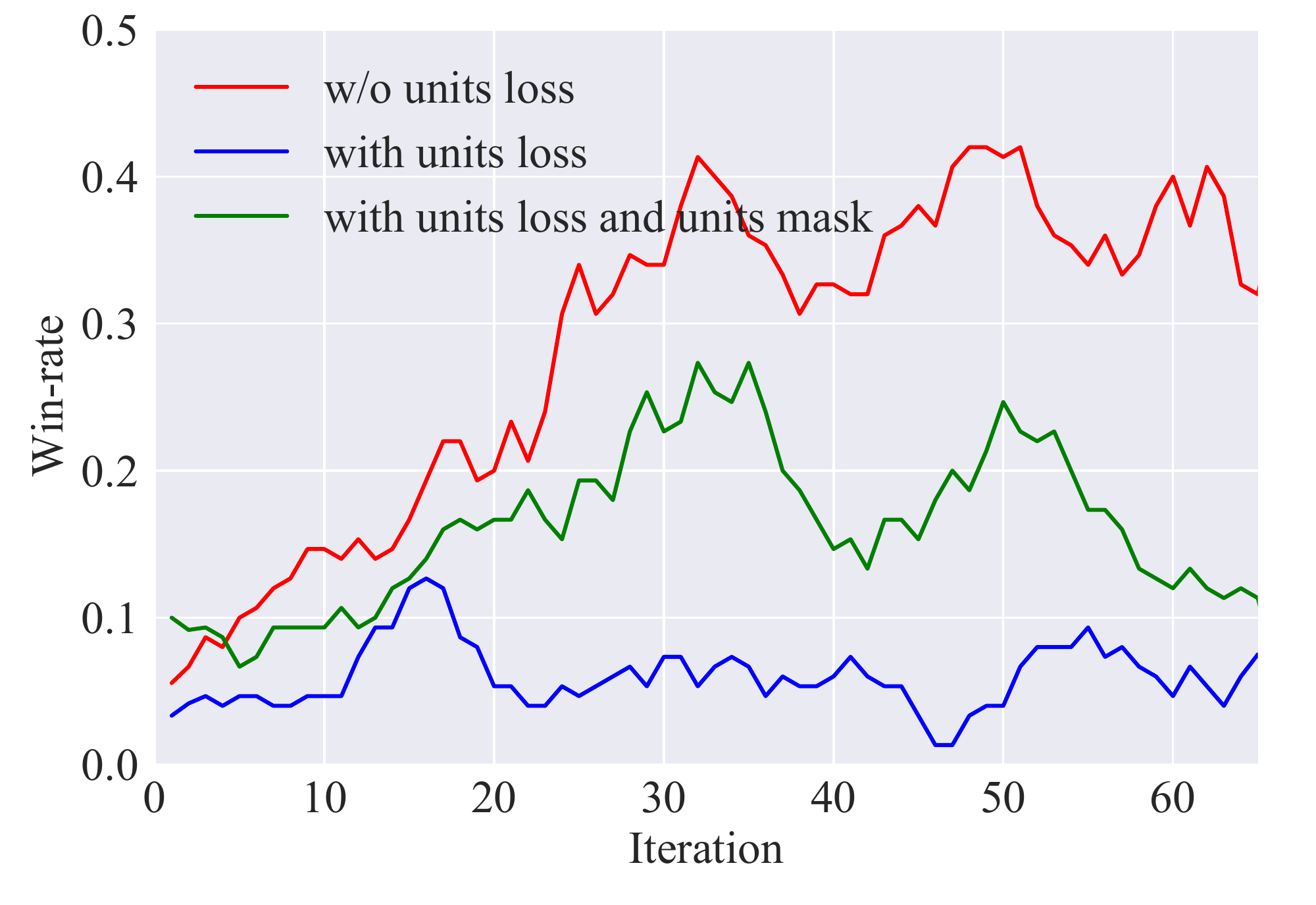}
		\label{fig:s.n.10.7}
   }	
   \subfloat[Shared Adam]{
		\centering
		\includegraphics[width=0.230\columnwidth]{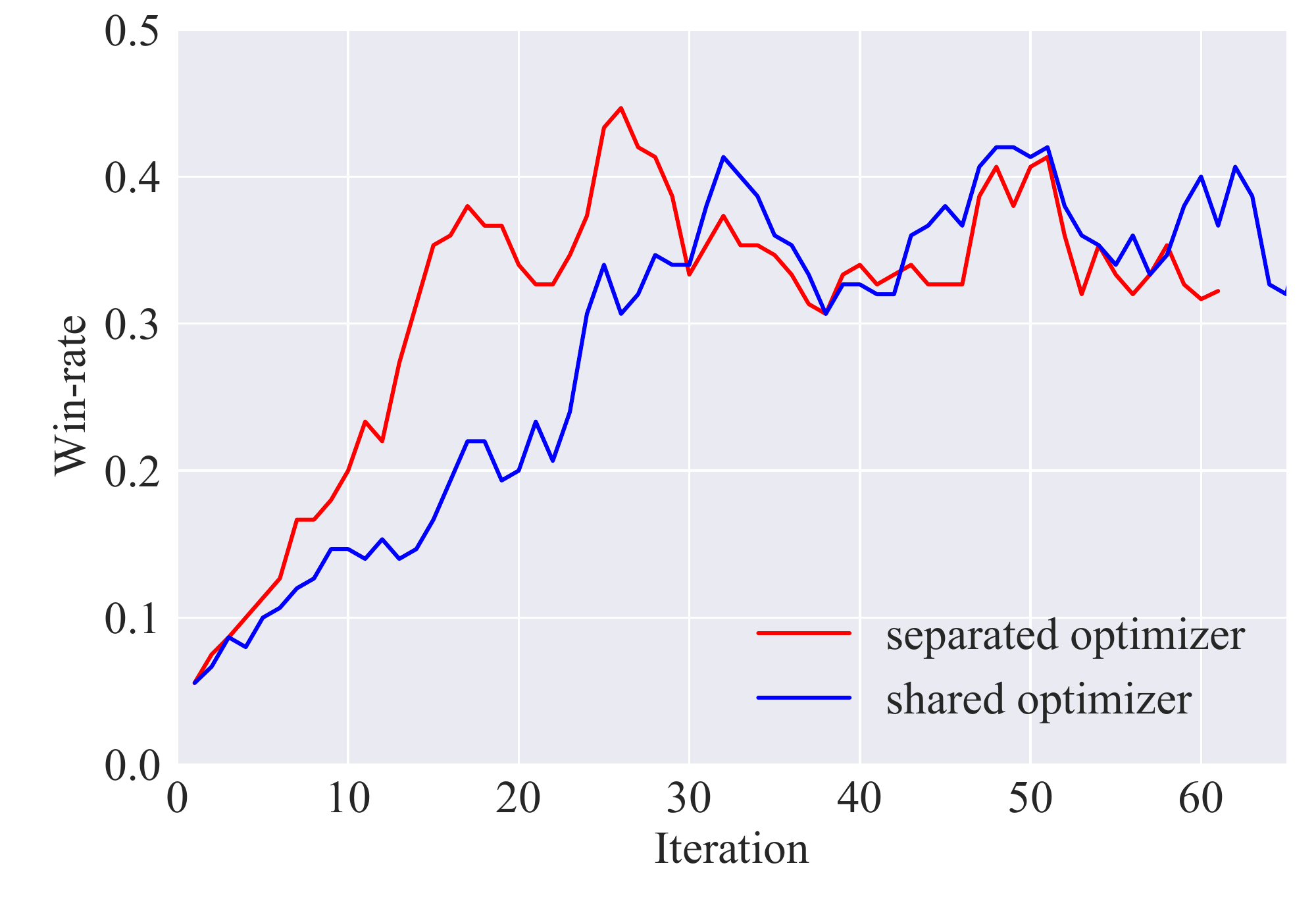}
		\label{fig:s.n.10.8}
	}	
    \caption{\textbf{Comparison of different settings and hyper-parameters in RL training of mAS (best viewed in color).} (a): the impact of whether to use the pre-trained value network. Note that the training curve of using a pre-trained value network lies at 0. (b): the impact of using different learning rate. (c): the impact of using units loss, using no units loss, or using units loss plus an entity mask. (d): the impact of using a shared Adam optimizer or using a separated Adam optimizer.}
    \label{fig:s.n.a.10.2}
\end{figure*}

First, we test whether we should train the value network of mAS several epochs before training its policy network. Due to the value network is not trained in the SL phase, the training may fail if the policy network is trained based on a random initialed value network. However, we find using the random initialed value network can already have enough good training effect. On the contrary, if we pre-train the value network several epochs, the value network may instead be over-fitting, causing the training effect of the policy network to be bad. The comparison is shown in Fig.~\ref{fig:s.n.a.10.2} (a) (note due to the comparison being so obvious, we only run this for 5 iterations to save time).

The AlphaStar pseudocode provides an actor-learner multi-thread training scheme. In each thread, one actor runs an agent in one SC2 environment, generates the trajectory, and sends it to a central learner. The learner processes all the actors' trajectories, updating the parameters of the network when the number of trajectories exceeds the value of BATCH. In updating, the learner passes the batch of trajectory data into the RL loss calculation. The trajectories data contains the agent's states, memories, behavior logits, and actions. The states and memories are used for calculating target logits. Then target logits, behavior logits, and actions are used for the off-policy loss calculation. For calculating the advantages in the RL loss, AlphaStar uses the V-trace~\cite{Espeholt2018vtace} and UPGO~\cite{AlphaStarNature} algorithms. We use the auto-regressive way to calculate the target logits, i.e., injecting the actors' true action types in calculating as in supervised learning.

\begin{figure}[h]
	\begin{minipage}[t]{\linewidth}
		\centering
		\includegraphics[width=0.95\textwidth]{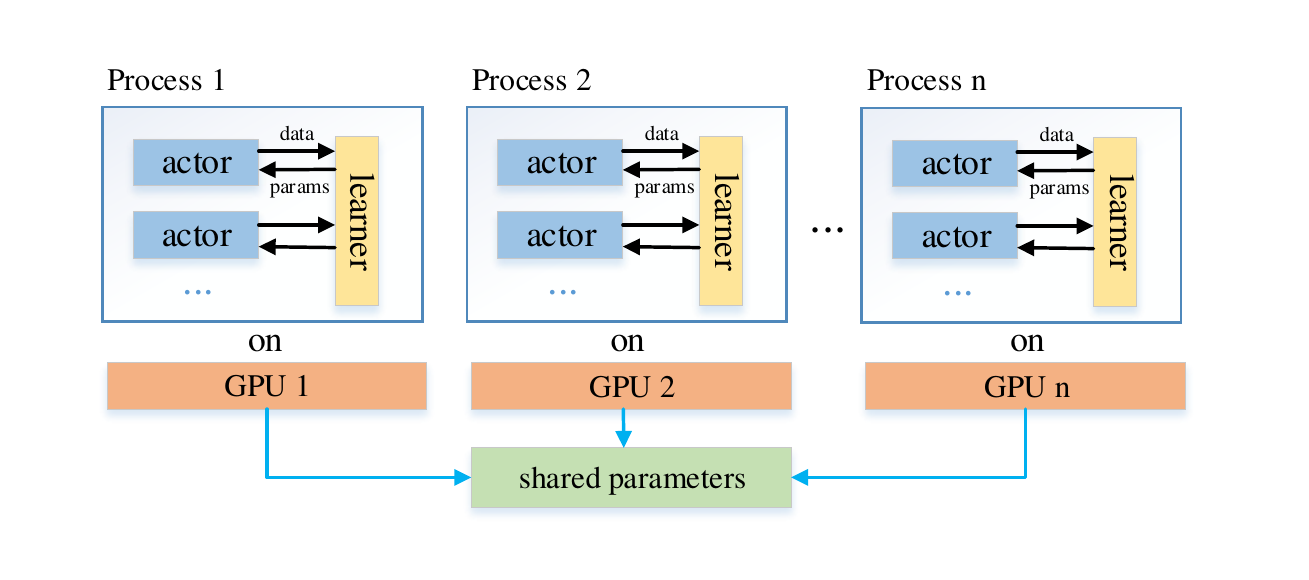}
		\caption{The multi-process RL training scheme in mini-AlphaStar.}
		\label{fig: RLTrainingArch}
	\end{minipage}
\end{figure}

We extend the multi-thread training scheme to a multi-process plus multi-thread training scheme (see Fig.~\ref{fig: RLTrainingArch}). The reason is that a multi-process of Python programming can increase the sampling speed by a margin. We let the multi-thread actor-learner be as before but create several copies of them, each of them in one separate process. Learners among all processes share the same neural network model. In each process, actors collect data and then send them to the learner for updating. Due to the model sharing, the updating will affect all learners. Every 10 seconds, actors will copy parameters from the learner which lie in the same process. This scheme implements a ``data parallelism between threads, parameters parallelism between process" scheme, similar to the multi-process training in our hierarchical approach. However, this new training scheme suits the AlphaStar architecture and off-policy RL training.

We make the RL training use multi-GPUs. We set the BATCH and SEQ in RL training to a proper value such that one process uses less than half of one card of GPU memory. We use 8 GPUs and run 15 processes at the same time. Due to the shared model residing on the first GPU costing more memory, the first GPU sustains 1 process, and each of the other 7 GPUs sustains 2 processes. In each process, we run 2 actor threads and 1 learner thread. We test two different updating ways: 1. use a shared Adam optimizer for all learners; 2. use one separate Adam optimizer for one learner. We find their results are similar (the comparison can be seen in Fig.~\ref{fig:s.n.a.10.2} (d)). By these settings, we run $15 * 2 = 30$ SC2 environments simultaneously. We statistic the win rate in every 30 games (call it one iteration) and draw learning curves. The number 30 is smaller than the 100 in the hierarchical approach. However, we find 30 is a statistical value enough to reflect the training effect. One iteration of this setting costs about 8 minutes, which is 197\% faster than before without using this new training scheme.

\begin{table}[t]
    \centering
    \scalebox{1.0}{
    \begin{tabular}{l | c  c }
    \toprule
      & mAS & Ours  \\
    \midrule  
	Action space               &raw action space                     &human  action space \\
	Learning method			   &SL and RL			     &RL \\
	Preprocessing              &none                    &data mining \\
	RL action space            &raw action space         &macro  action space \\
	Max game steps             &18000                    &18000 \\
	Decision interval          &16 step                  &1 seconds $\approx$ 22 step \\
	Reward                     &winloss (-1, 0, 1)       &winloss (-1, 0, 1) \\
    \bottomrule
    \end{tabular}
    }
    \caption{Comparison of training settings between mAS and our hierarchical approach.}
    \label{tab: mas Compare Setting}
\end{table}

Fig.~\ref{fig:s.n.a.10.2} shows a comparison of different RL training settings and hyper-parameters. Fig.~\ref{fig:s.n.a.10.2} (a) shows the difference of training with/without a pre-trained value network. We find the learning rate keeps a crucial role in the training. Among several learning rates, we find 1e-6 is the best (the comparison is in Fig.~\ref{fig:s.n.a.10.2} (b)). Note that AlphaStar's default learning rate (3e-5) doesn't fit our training, which shows that a reasonable learning rate depends on specific settings. We also compare the training with/without units loss in Fig.~\ref{fig:s.n.a.10.2} (c). Using a ``unit type entity mask", we find that the training curve with units loss can also grow, but its result is not as good as that without units loss. The reason may be that unit loss is the most complicated one in all action arguments, making its training sensitive and unstable. Sufficient SL training (using more data and more training time) may alleviate this problem. Fig.~\ref{fig:s.n.a.10.2} (d) compares the effect of a shared Adam optimizer for all processes with a separated Adam optimizer for each process.


\subsection{Final Results}
Here we compare the final results of the mini-AlphaStar with our proposed hierarchical approach. The comparison of training settings between mAS and the hierarchical approach is shown in Table~\ref{tab: mas Compare Setting}. For mAS, we first use the SL method to train a model and then use the RL method to train a model based on the SL model. We test the agent's performance by playing against the level-1, level-2, level-3, and level-7 built-in AI. The test setting is the same as our hierarchical approach. The opponent is a Terran built-in AI, and the map is Simple64.

\begin{table}[h]
    \centering
    \scalebox{1.0}{
    \begin{tabular}{l | c c c | c c c c }
    \toprule
      & Train time  & CPUs & Train type & Level-1 & Level-2 & Level-3 & Level-7 \\
    \midrule
    mAS  & 0.3 day   &  48   &  SL & 0.15  & 0.07  & 0.02 & 0.00    \\
	mAS  & 1 day  &  48   &  RL & 0.54  & 0.30  & 0.02 & 0.00    \\
	mAS  & 3 days  &  48   &  RL & 0.74  & 0.51  & 0.08 & 0.00    \\
	\midrule
    Ours  &  3 days   & 48   &  RL  & \textbf{1.00}  & \textbf{1.00}  & \textbf{0.99}  & \textbf{0.93}     \\
    \bottomrule
    \end{tabular}
    }
    \caption{Comparison of training results between mAS and our hierarchical approach.}
    \label{tab: mas Compare Results}
\end{table}

We run 100 games (repeated by 3 times) for each level and record the win rate results. We show the mAS's training results of 3 stages which are: 1. a supervised learning model which is trained by 8 hours; 2. a reinforcement learning model which is trained by 1 day based on the previous SL model; 3. a final reinforcement learning model which is trained by 3 days based on the previous RL model. From Table~\ref{tab: mas Compare Results}, we can see that the mAS's performance is continually increased by the advancement of training time, showing that training time is a key factor for good performance of this architecture.

We find that though the agent after training can get some win rate on the low levels, it can not get one victory on the high level of level-7. The reason may due to the following two ones. Firstly, supervised learning of mAS still suffers from insufficient data due to our machine's capacity restriction. Using more disk space or optimizing storage for these replays may alleviate this problem. Secondly, reinforcement learning of mAS also has insufficient training time due to the restriction of the number of machines. In our experiment, we found that RL training with 3 days is significantly better than RL training with 1 day. Using more machines for RL training or giving more training time may alleviate this problem.


We can see that the technology in AlphaStar relies fairly on the usage of computing resources, while our hierarchical approach shows more advantages when using limited resources. Note the comparison here is not a comparison to the original AlphaStar's results but a comparison to the results of a limited variant version (mAS) with limited resources. To make the comparison fair, we make the two ones using the same computing resource and training time.

\section{Additional Experiments} \label{section: Additional Experiments}
In this section, we handle the problem of how to train an agent on the cheating levels. We also give the results of training on two other maps than Simple64.


\begin{figure*}[h]
    \centering
    \subfloat[Harder levels (R)]{
        \centering
        \includegraphics[width=0.230\columnwidth]{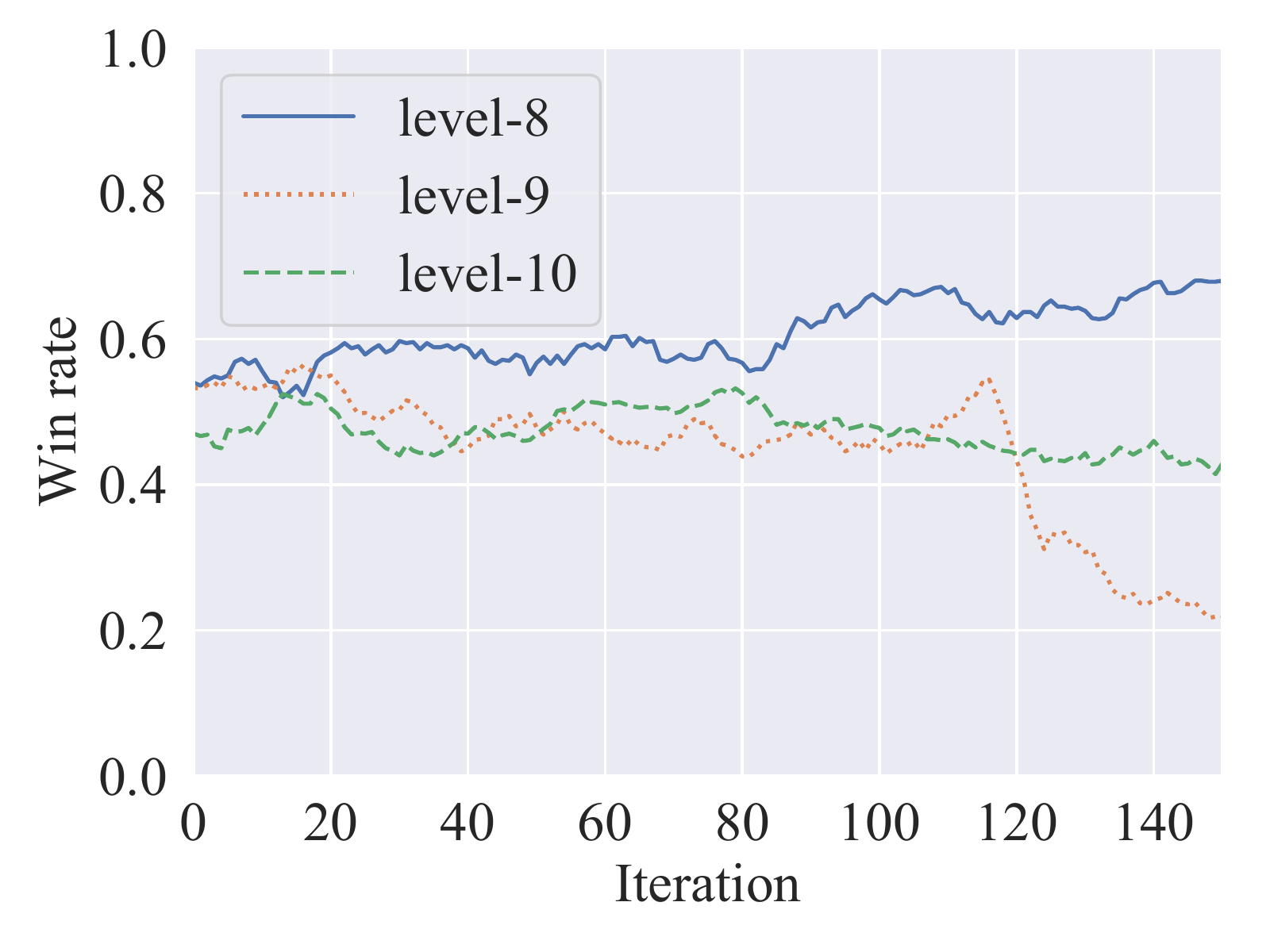}
		\label{fig:s.4.1.1}
   }
    \subfloat[Harder levels (F)]{
        \centering
        \includegraphics[width=0.230\columnwidth]{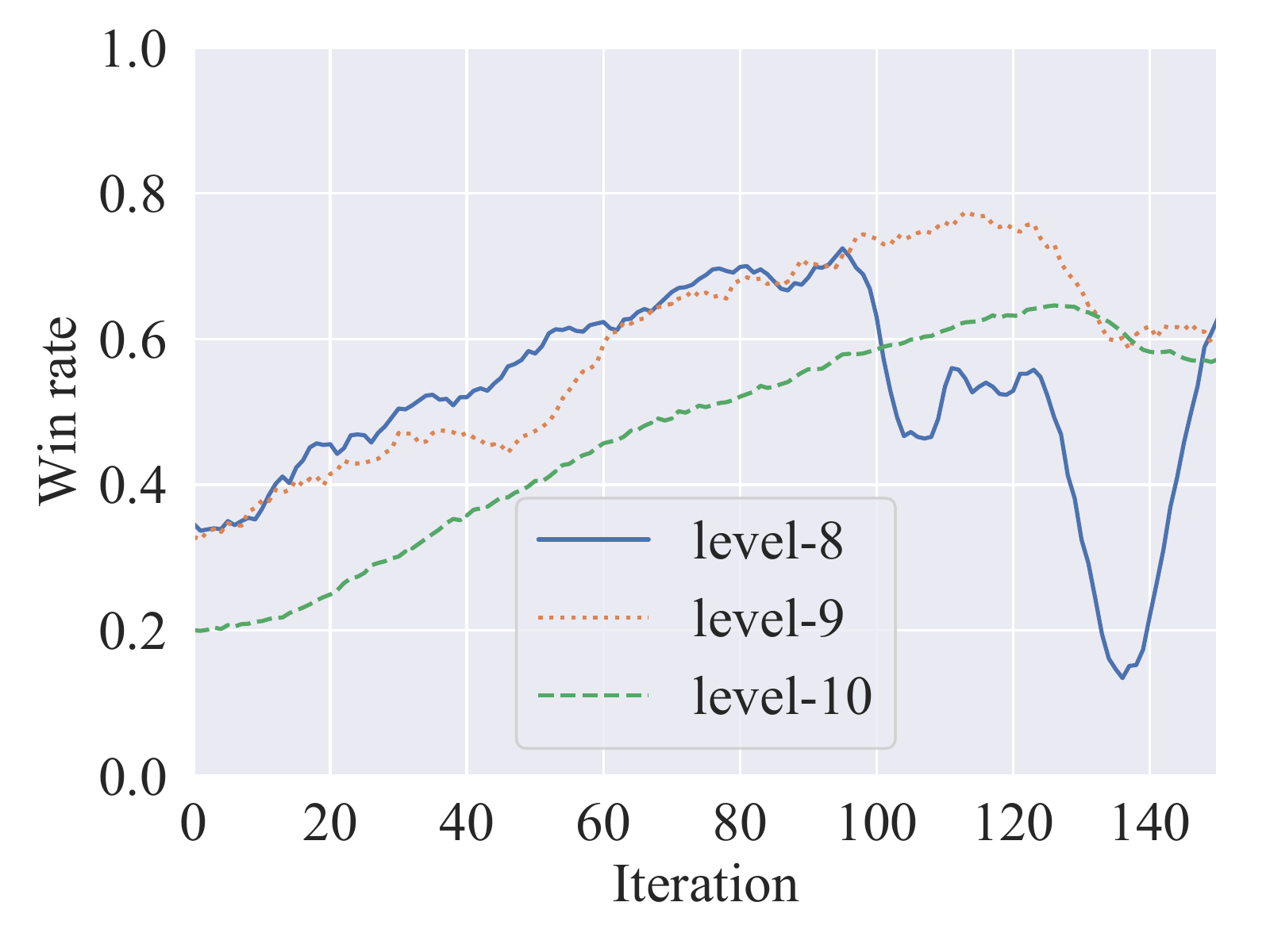}
		\label{fig:s.4.1.2}
    }
    \subfloat[Other maps (R)]{
        \centering
        \includegraphics[width=0.230\columnwidth]{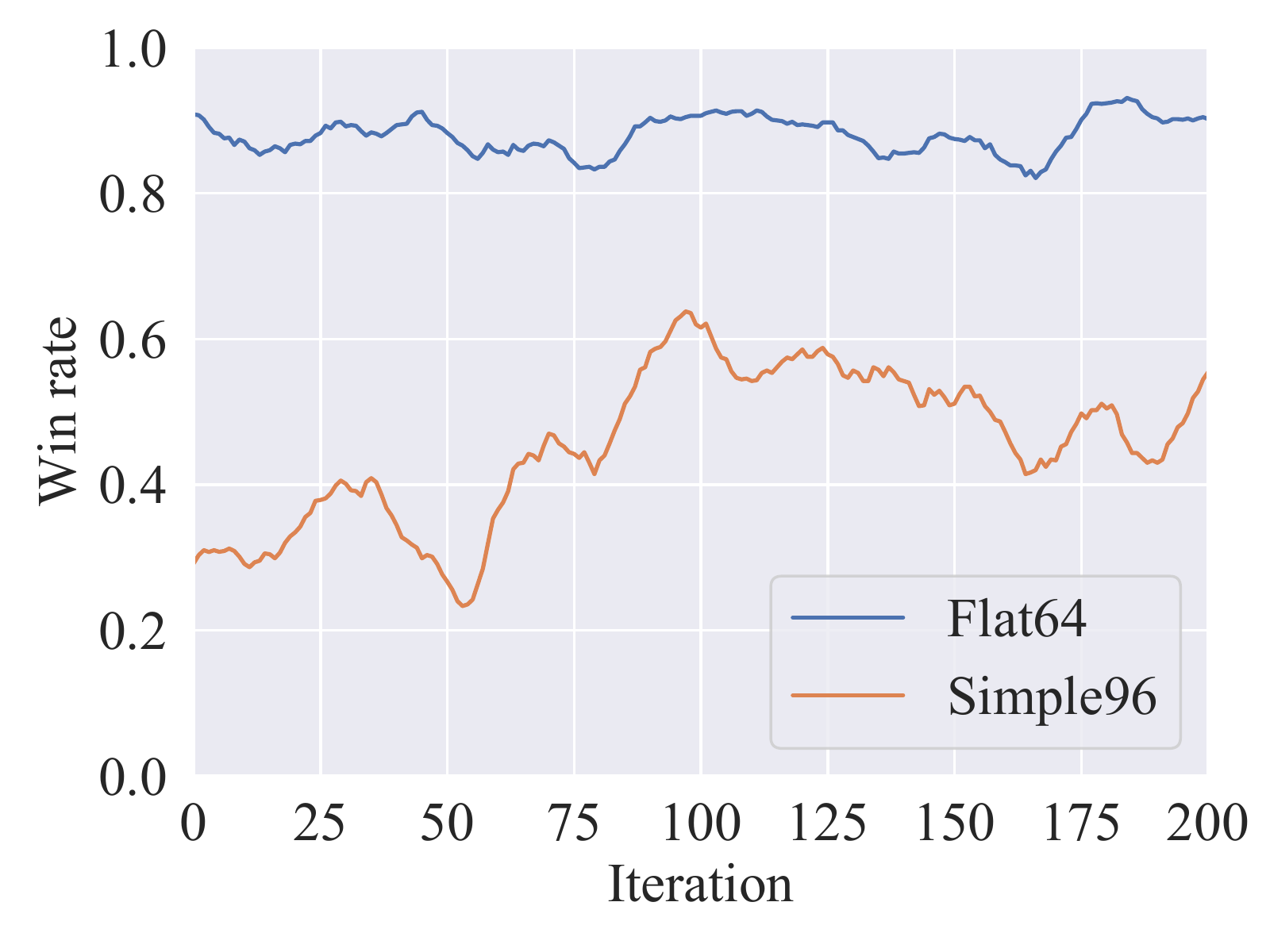}
		\label{fig:s.4.2.1}
   }	
   \subfloat[Other maps (F)]{
		\centering
		\includegraphics[width=0.230\columnwidth]{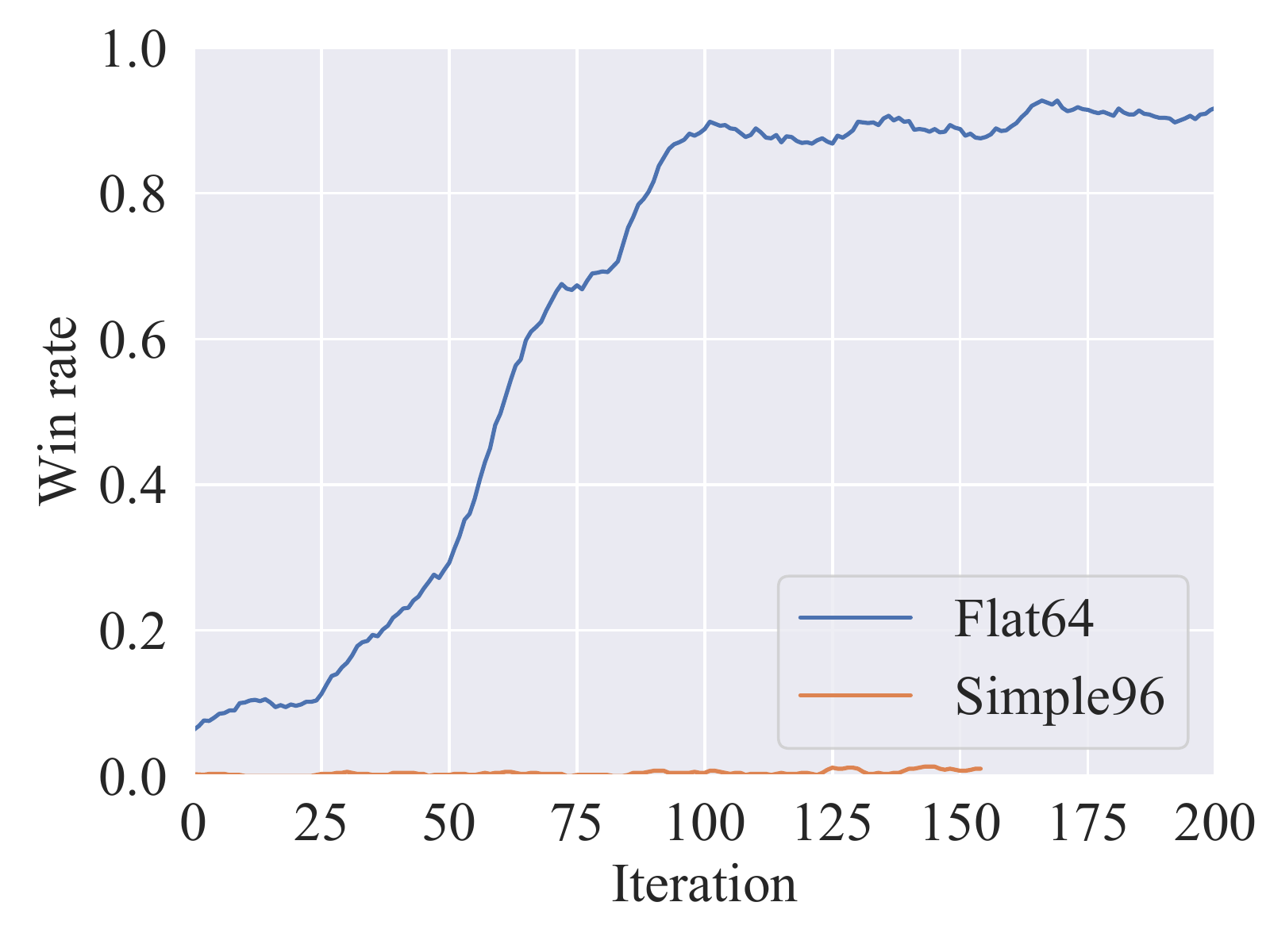}
		\label{fig:s.4.2.2}
	}	
    \caption{Training results in the additional experiments. R=Restore Model, F=From Scratch.}
    \label{fig:s.a.4}
\end{figure*}

\subsection{Training on Cheating Levels}
We did not train on the cheating difficulty levels in the previous experiments. Because the AI of cheating difficulty levels apply some cheating techniques. For example, the built-in AI of difficulty level-8 will see what our agent can see. The difficulty level-9 AI can get more resources than our agent. The built-in AI of difficulty level-10 (also called level-A) can get both the advantage of level-8 and level-9. In the conference version of this paper, we trained our agent against a level-7 built-in AI and then tested our agent on the difficulty level-8 to level-10. Here we try to train an agent against these cheating difficulty level AIs by using a new network architecture and reward settings.

The new training framework is as follows. We extended the two-layer hierarchical architecture to a three-layer (3-layer) hierarchical one. We built a new layer under the original base sub-policy, including two sub-networks: the building sub-network and the population sub-network. The building sub-network is responsible for constructing the buildings, and the population sub-network is accountable for producing soldiers and workers. Due to the flexibility of our hierarchical architecture, our training algorithm remained the same when we added a layer under the base sub-policy. It is still that each sub-network updates its network parameters with its replay buffer. The rewards of the base sub-policy come from the building sub-network and the population sub-network.

In the previous discussion, we point out that when using win/loss rewards, it is hard for agents to learn in the SC2 environment. So we use some hand-designed rewards for training. Hand-designed rewards can make the reward denser, thus making the agent learn faster. However, hand-designed rewards may also make the agent learn something that we don't want or try to get more rewards instead of winning a game. We try to use the outcome as our reward in training on cheating levels. The reward is 0 on all the time steps, except a +1 (win), 0 (tie), or -1 (loss) reward on the last step. As discussed before, this setting will make the rewards sparse, making learning difficult. However, on the cheating difficulty levels, we find that this result reward can get better results, though the learning at first is slower.

Fig.~\ref{fig:s.a.4} (a) and Fig.~\ref{fig:s.a.4} (b) show the training curves of the agent using our 3-layer hierarchical architecture on difficulty level-8 to level-10. We find that when training on cheating difficulties levels, both the 3-layer hierarchy and win/loss reward are important for training an effective agent. Using only one technique will cause less performance than before, which means the combination of them is the key to training an effective one.

Fig.~\ref{fig:s.a.4} (a) shows training curves that are all based on the previous model of difficulty level-7. The training effect based on the previous model is not good due to the randomness of the cheating difficulty level built-in AI. Even in some difficulty levels, the effect is degraded. It suggests that when the two environments are not the same (e.g., from non-cheating levels to cheating levels), transfer learning is not good. Therefore, we tried to train from scratch, which is shown in Fig.~\ref{fig:s.a.4} (b). It can be seen in Fig.~\ref{fig:s.a.4} (b) that although the vibration is more intense, the learning curve as a whole shows an upward trend. 

Since the agent's learning fluctuates wildly, we do not store the model in each iteration. Instead, the model is stored only when the win rate of the current iteration exceeds the previous best one. In this way, we obtained a win rate of  0.79 in iteration 188 on difficulty level-8, a win rate of 0.87 in 213 iterations on difficulty level-9, and a win rate of 0.86 in 636 iterations on difficulty level-10. The comparison results are shown in Table~\ref{tab: against cheating}.


\begin{table}[h]
    \centering
    \scalebox{1.0}{
    \begin{tabular}{l | c c c }
    \toprule
    Difficult  & Level-8 & Level-9 & Level-10 \\
    \midrule
    Previous 2-layer  &   0.74   &  0.71   &  0.43 \\
    3-layer  &    \textbf{0.79}   &  \textbf{0.87}   &  \textbf{0.86} \\
    \bottomrule
    \end{tabular}
    }
    \caption{Training results using 3-layer hierarchy against cheating-level AIs.}
    \label{tab: against cheating}
\end{table}

\subsection{Training on Other Maps}
To test the scalability of our method, we train the hierarchical architecture on other maps. The maps we tested are \textit{Flat64} and \textit{Simple96}. Flat64 is similar to Simple64, but the shape of the map is square. Simple96 expands the map of Simple64 by 50 percent. The settings for training here are the same as the settings in training for cheating difficulty levels. The agent race is Protoss, and the built-in AI race is Terran. The difficulty of the opponent is level-7. The combat model we use here is the combat rules. 

At first, we tried to train from scratch. We found that this setup was slower, especially on Simple96. We instead use the idea of transfer learning. We use the policy trained on Simple64 as the initial model. Fig.~\ref{fig:s.a.4} (c) shows that the initial win rate of Flat64 is high. While the initial win rate on Simple96 is not high but can grow fastly. From these results, it can be seen that transfer learning is effective between different maps.

Simple96 is more challenging than Flat64 because when the map is larger, the execution requirements for the Rush tactic become more strict, making it take more time to learn a good policy. The effect of learning from scratch without transfer learning is shown in Fig.~\ref{fig:s.a.4} (d). It can be seen that although it is more challenging to learn on Simple96, the effect on Flat64 is good.

\subsection{Final 3-layer Hierarchical Architecture}
\begin{figure}[h]
	\begin{minipage}[t]{\linewidth}
		\centering
		\includegraphics[width=0.90\textwidth]{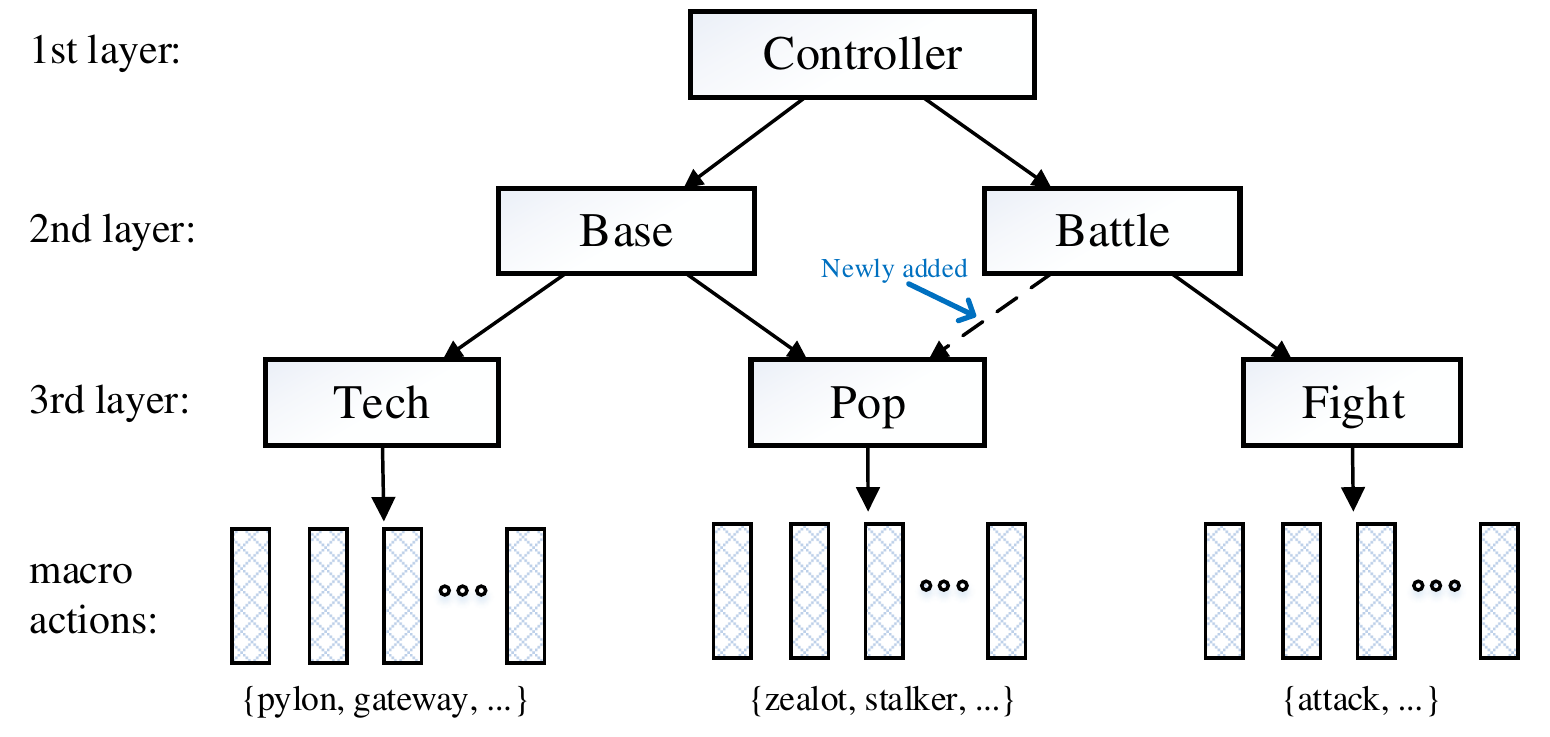}
		\caption{The Final 3-layer Hierarchical Architecture. Note the added control line, which makes the hierarchy more symmetrical.}
		\label{fig:NewThreeLayer}
		\end{minipage}
\end{figure}

We found that if the agent can control the battle and produce soldiers alternatively in a period of time, it will have a higher win rate. This result makes sense: for humans, doing the two things in a small time span simultaneously increases the offensive efficiency. Inspired by this assumption, we refined the 3-layer architecture to a ``final 3-layer'' architecture (shown in Fig.~\ref{fig:NewThreeLayer}). We find that using the final 3-layer architecture, the training becomes faster, and the result becomes better, which can achieve more than a 90\% win rate against the level-10 built-in AI from scratch (see Fig.~\ref{fig:s.a.9.1} (a)).

Note that we have added a control action from the battle sub-network to the population sub-network. Now, the battle sub-network also has a child sub-network called the ``fight'' sub-network. These changes make the overall hierarchy more symmetrical and similar to the classical hierarchical control architecture in the Taxi-driven problem in MAXQ~\cite{dietterich2000maxq}.

\subsection{Best Practices for Training Agents}
\begin{figure*}[t]
    \centering
    \subfloat[Layer archt.]{
        \centering
        \includegraphics[width=0.230\columnwidth]{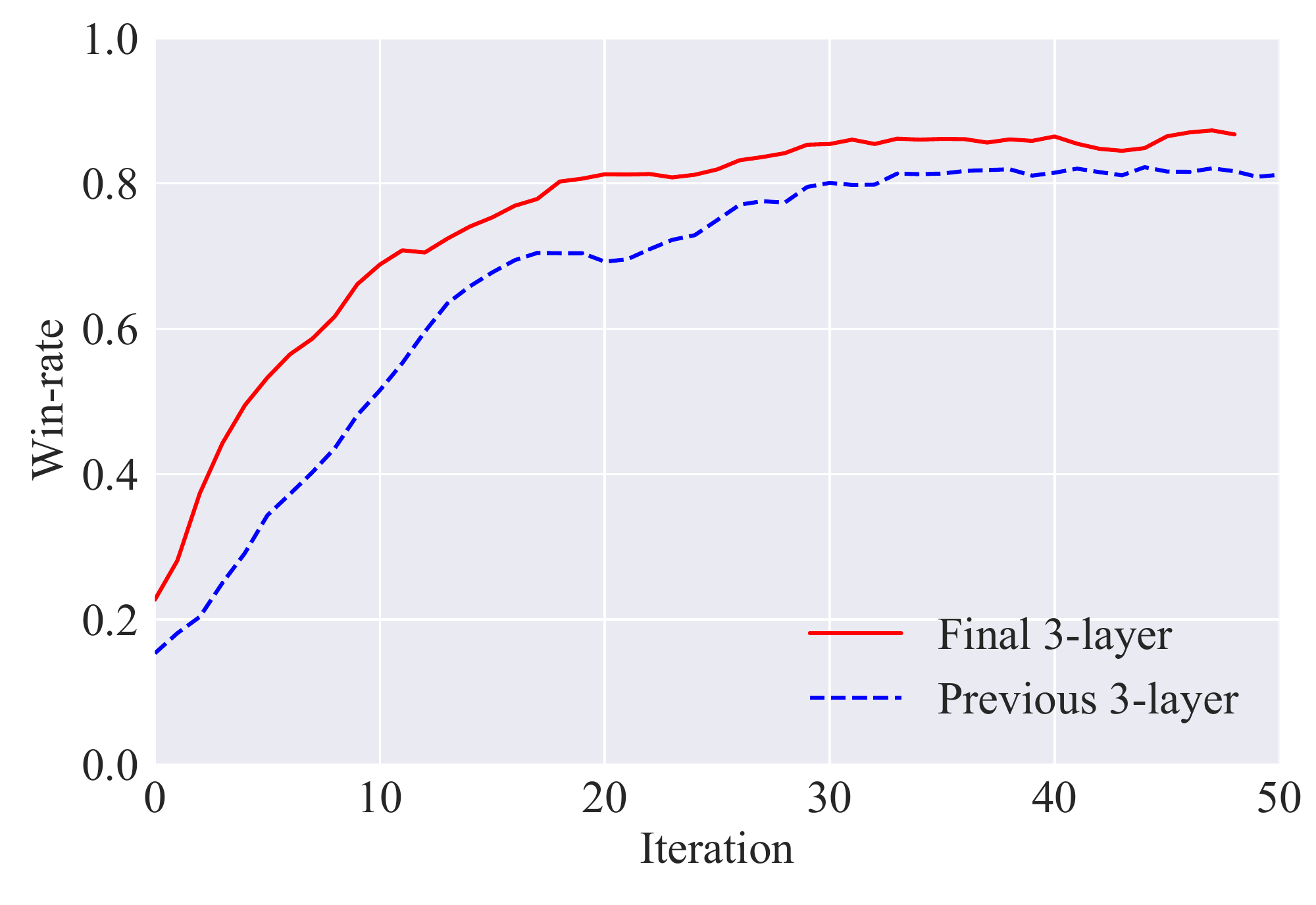}
		\label{fig:s.4.9.1}
   }
    \subfloat[Shared nets]{
        \centering
        \includegraphics[width=0.230\columnwidth]{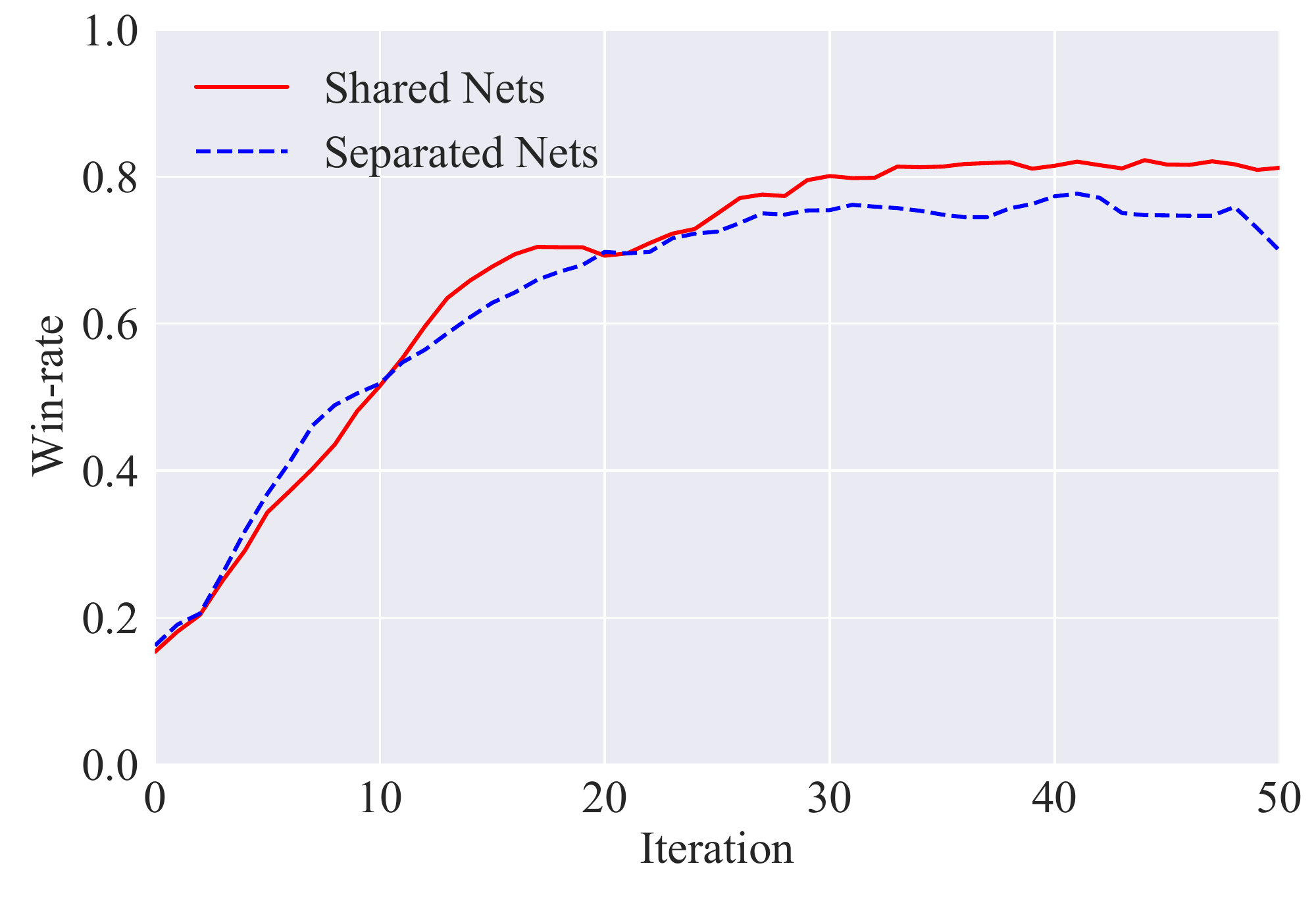}
		\label{fig:s.4.9.2}
    }
    \subfloat[Episode num]{
        \centering
        \includegraphics[width=0.230\columnwidth]{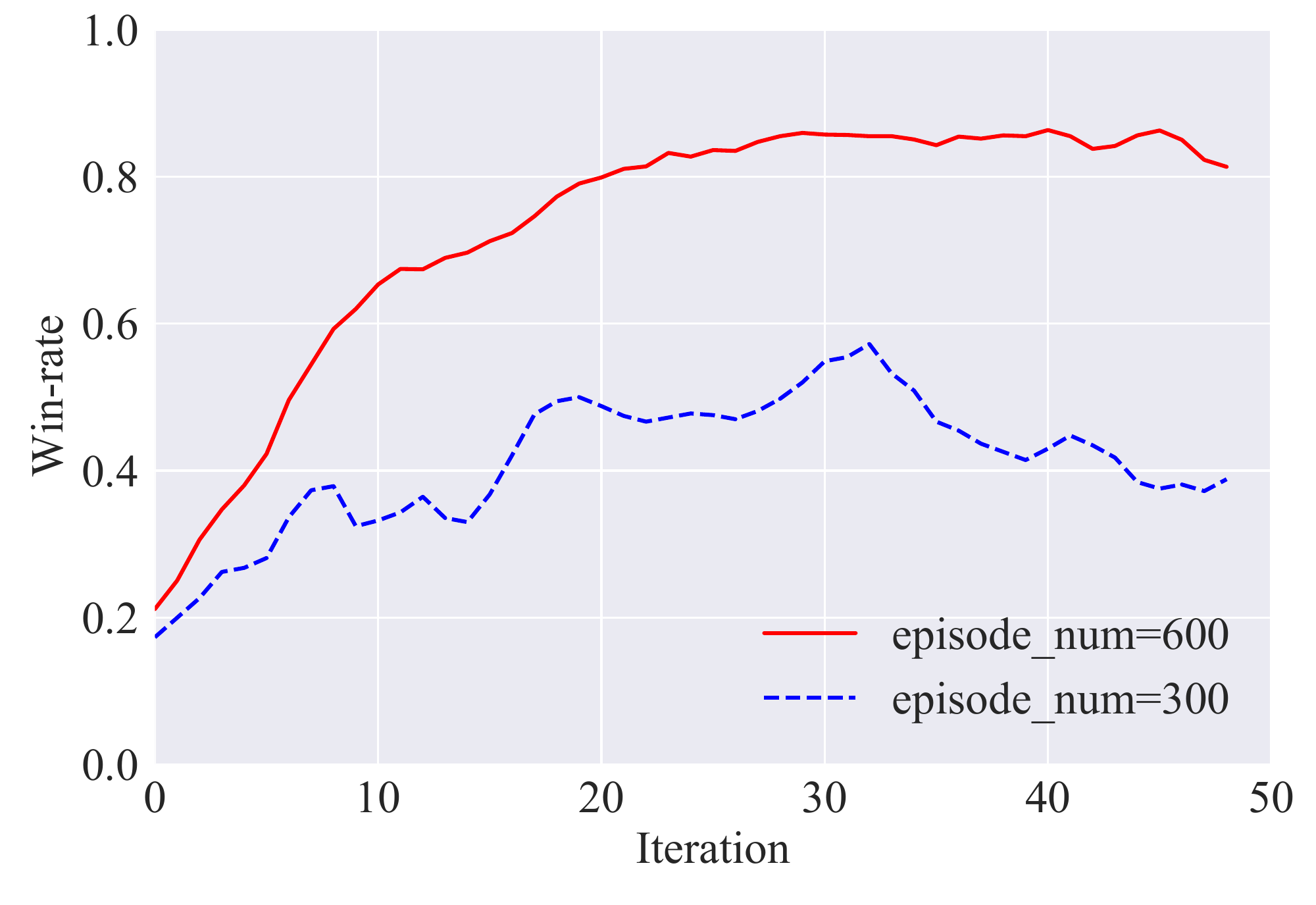}
		\label{fig:s.4.9.3}
   }	
   \subfloat[Batch size]{
		\centering
		\includegraphics[width=0.230\columnwidth]{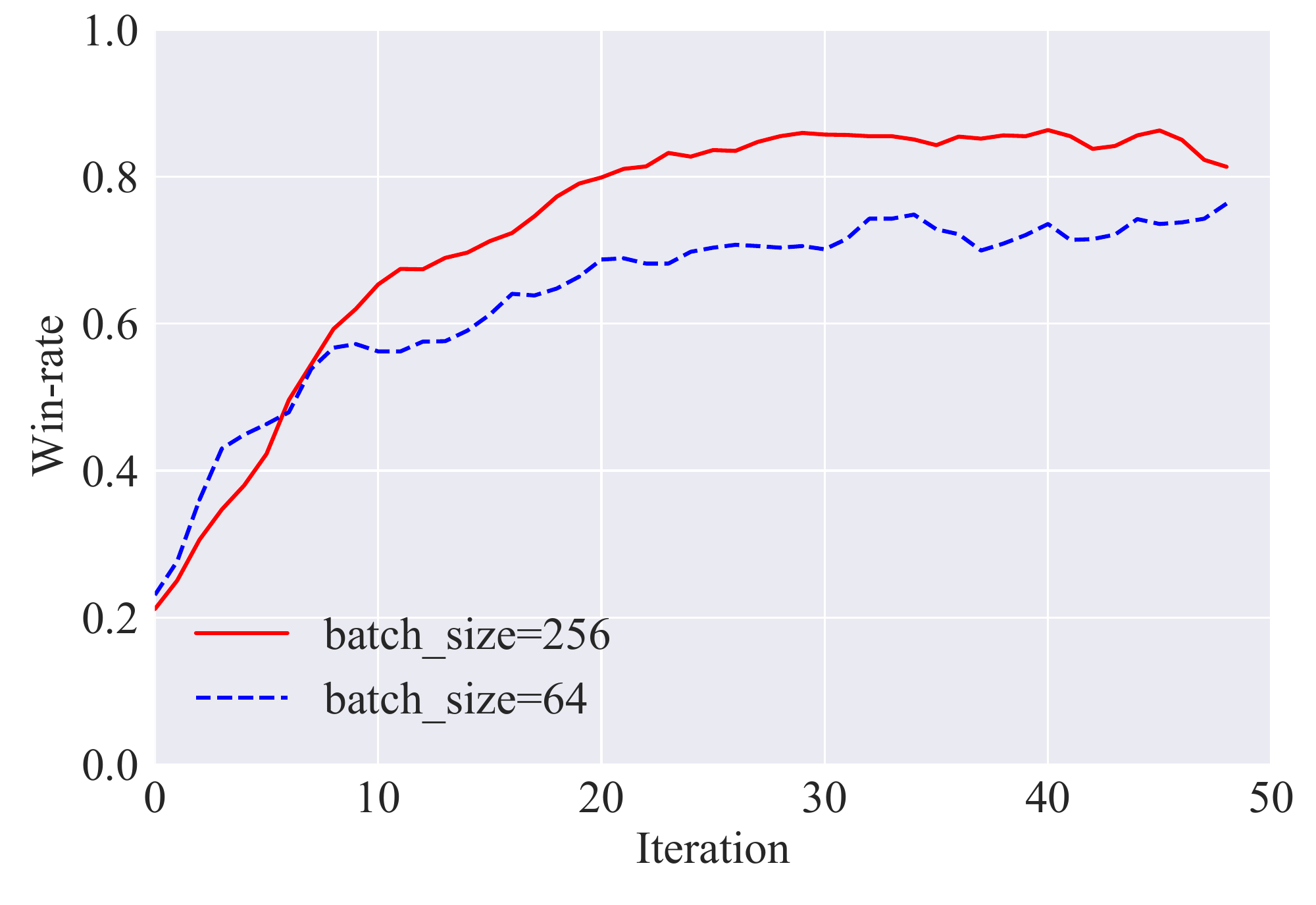}
		\label{fig:s.4.9.4}
	}	
    \caption{Best practices for experiments on training SC2 agents. (a): the impact of the final 3-layer architecture. (b): the impact of the shared parameters of networks. (c): the impact of the number of episodes in each updating iteration. (d): the impact of the batch size in each updating iteration.}
    \label{fig:s.a.9.1}
\end{figure*}

\begin{figure*}[t]
    \centering
    \subfloat[Epoch nums]{
        \centering
        \includegraphics[width=0.230\columnwidth]{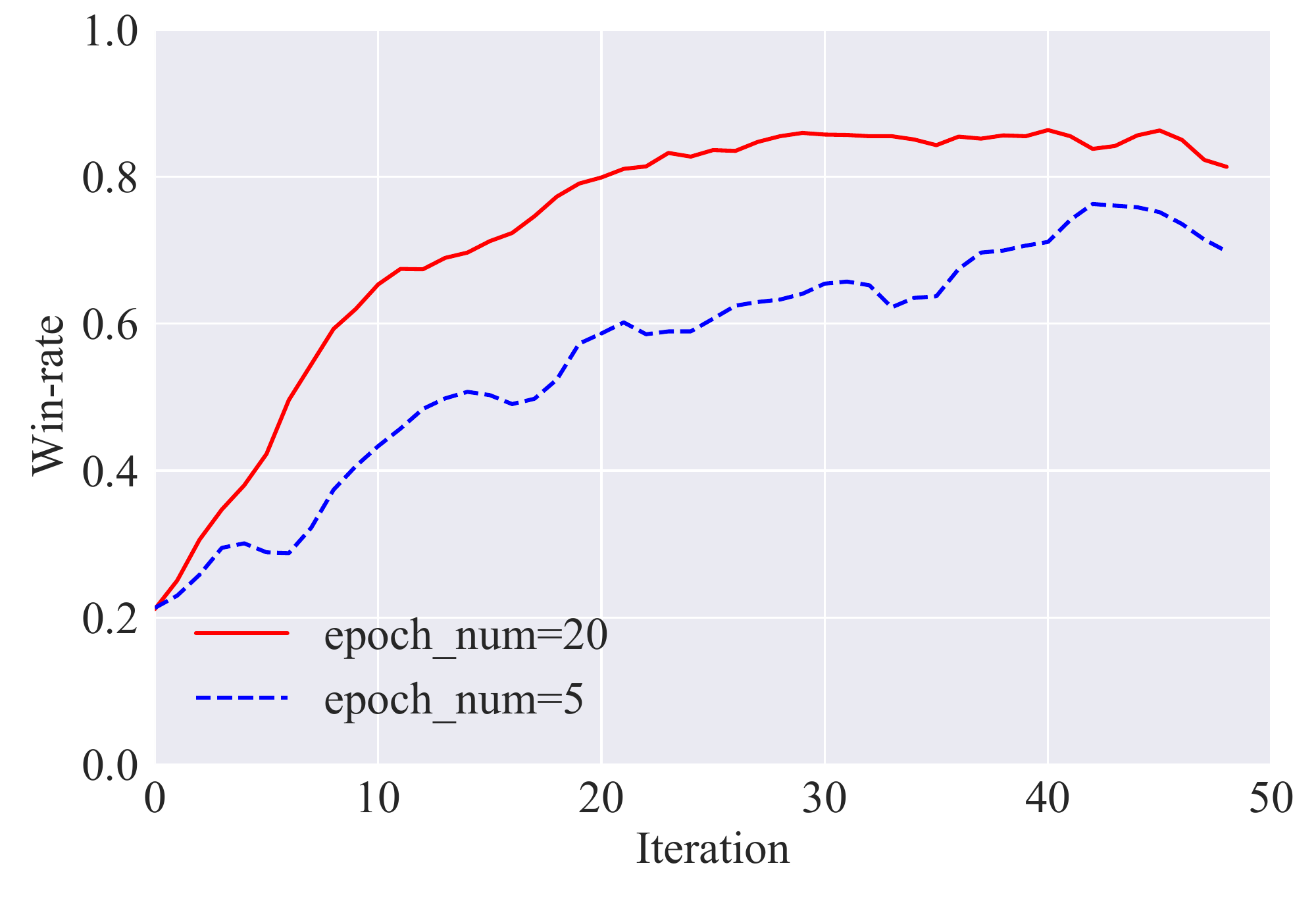}
		\label{fig:s.4.9.5}
   }
    \subfloat[c\_1 in PPO loss]{
        \centering
        \includegraphics[width=0.230\columnwidth]{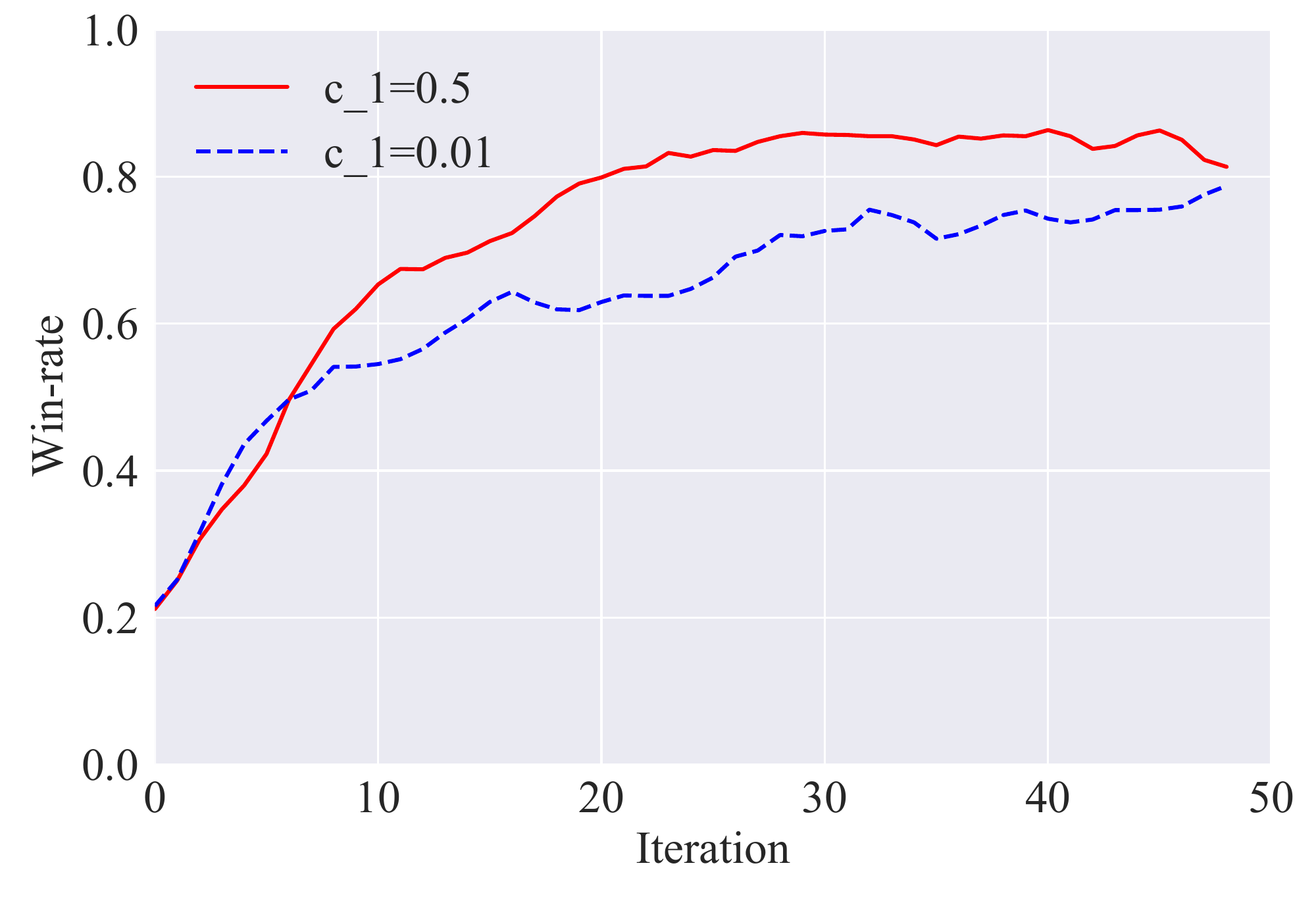}
		\label{fig:s.4.9.6}
    }
    \subfloat[Queued attack]{
        \centering
        \includegraphics[width=0.230\columnwidth]{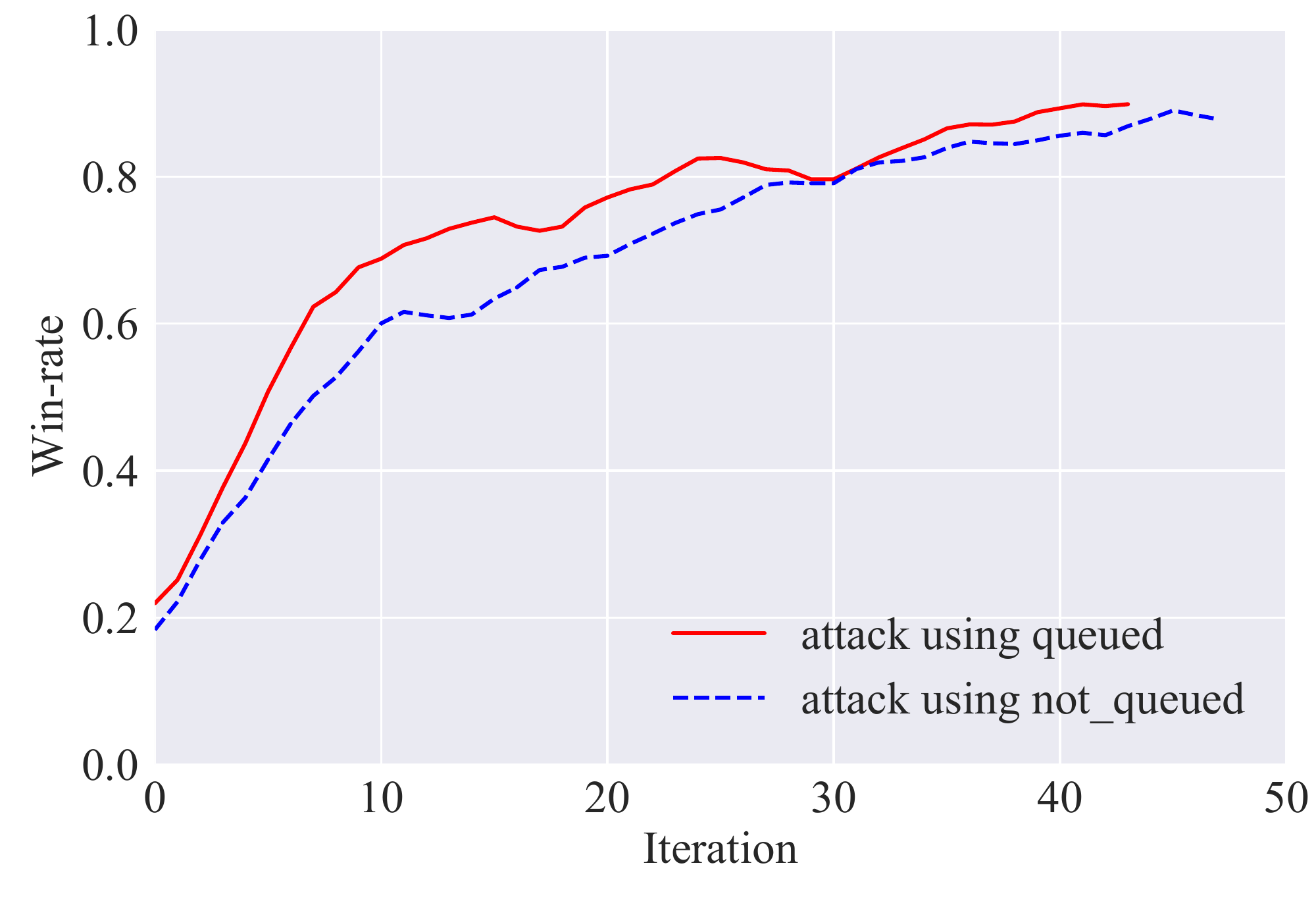}
		\label{fig:s.4.9.7}
   }	
   \subfloat[With no\_op]{
		\centering
		\includegraphics[width=0.230\columnwidth]{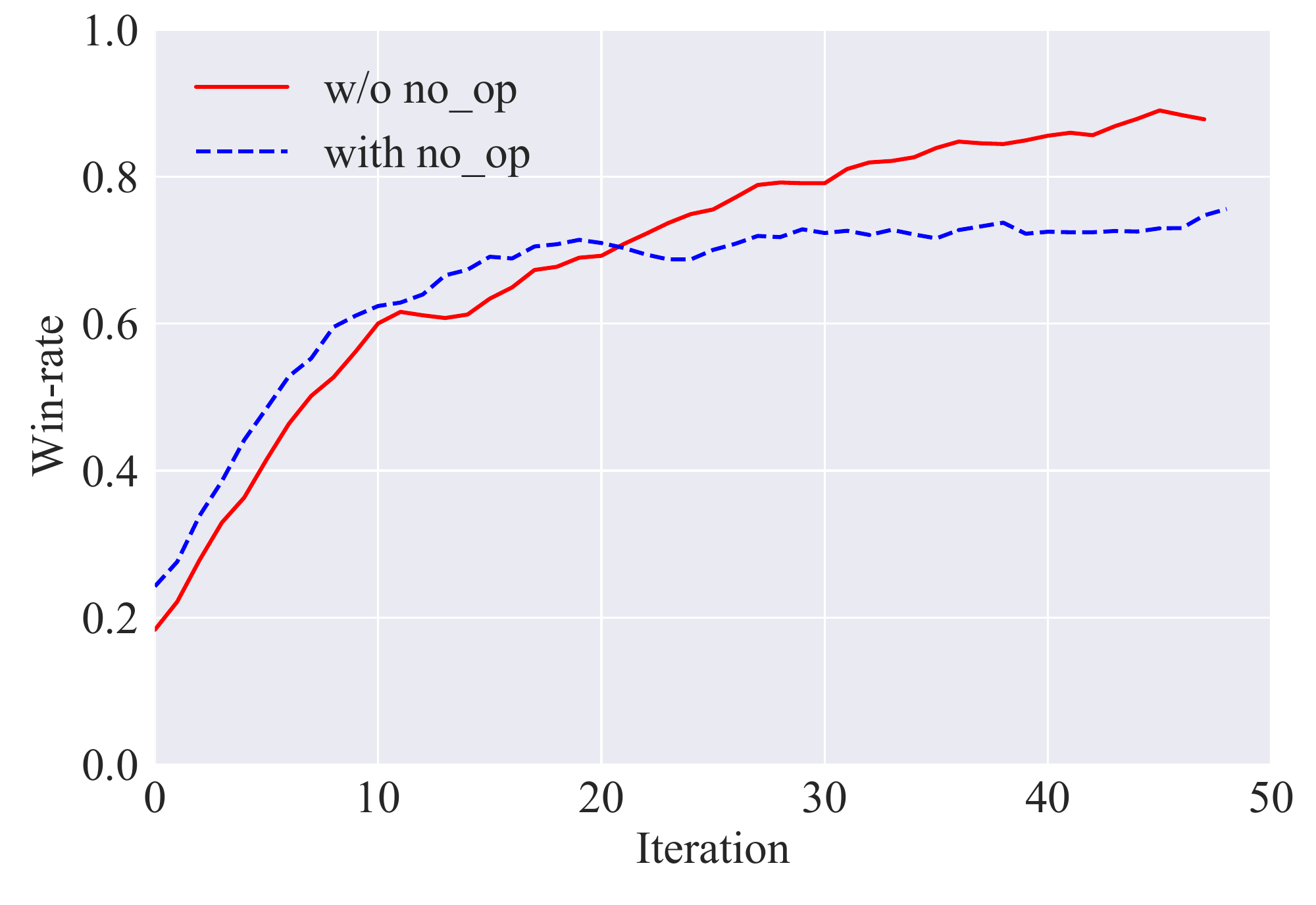}
		\label{fig:s.4.9.8}
	}	
    \caption{More best practices for training SC2 agents. (a): the impact of the number of epochs in each update of training. (b): the impact of the c\_1 hyper-parameter in the PPO algorithm. (c): the impact of the queued attack setting in the macro actions. (d): the impact of the with/without no\_op operation in the action space of each leaf sub-network.}
    \label{fig:s.a.9.2}
\end{figure*}

Fig.~\ref{fig:s.a.9.1} (a) shows the improvements of the final 3-layer architecture. Besides this, we also find other useful tricks that can increase the performance of the trained SC2 agents. We conclude these as the best practices for training, which are as follows: (1) letting the policy and value network in PPO share the same previous layers can increase the performance, which is shown in Fig.~\ref{fig:s.a.9.1} (b); (2) more episodes of games in one updating can increase the win rate (see Fig.~\ref{fig:s.a.9.1} (c)); (3) setting batch size to a larger value improves the performance (see Fig.~\ref{fig:s.a.9.1} (d)); (4) more epochs in each updating makes training better, which is shown in Fig.~\ref{fig:s.a.9.2} (a); (5) setting c\_1 to a proper value gets better performance (see Fig.~\ref{fig:s.a.9.2} (b)); (6) the attack macro actions can be set to be queued or not\_queued, and setting it to queued is better (see Fig.~\ref{fig:s.a.9.2} (c)); (7) the hierarchical architecture has some leaf sub-networks. Should the leaf sub-networks contain no\_op in its action space? The setting without no\_op gets better results (see Fig.~\ref{fig:s.a.9.2} (d)), which may be due to that the higher layer could decide which leaf sub-networks to get control of, thus achieving the effect of doing no\_op for the leaf sub-networks.


\subsection{Final Results against Cheating Level AIs}
Combining the final 3-layer hierarchy and the best practices for training agents, we achieve the best results of 0.96, 0.97, and 0.94 win rates against the level-8, level-9, and level-10 built-in AIs, respectively, shown in Table~\ref{tab: final against cheating}. Note these results have already been better than another work of ours~\cite{liu2021ThoughtGame}. Compared to that method~\cite{liu2021ThoughtGame}, the work in this paper also has another advantage which is there is no need to manually design an abstract model to help RL training.


\begin{table}[h]
    \centering
    \scalebox{1.0}{
    \begin{tabular}{l | c c c }
    \toprule
    Difficult  & Level-8 & Level-9 & Level-10 \\
	\midrule
	TG~\cite{liu2021ThoughtGame}  &   0.95   &  0.94   &  0.90 \\
	\midrule
    3-layer  &    0.79   &  0.87   &  0.86 \\
    Final 3-layer  &   \textbf{0.96}   &  \textbf{0.97}   &  \textbf{0.94} \\
	\bottomrule
    \end{tabular}
    }
    \caption{Training results against the highest cheating-level AIs.}
    \label{tab: final against cheating}
\end{table}


\subsection{Discussion}
Strategies used by built-in AIs are highly optimized by game developers and therefore pose significant challenges to learning algorithms. Our method is characterized by the abstraction and reduction of StarCraft II from multiple levels. For example, by learning the macro-actions, the action space in StarCraft II is greatly reduced. The long horizon time step problem is mitigated through the hierarchical approach. The combination of these techniques makes our framework suitable for solving large-scale RL problems such as StarCraft II. Our current approach still has some shortcomings. For example, the 64x64 map we are currently testing is small, and we only use the first two combat unit types.

\section{Conclusion} \label{section: Conclusion}
In this paper, we investigate a set of techniques of reinforcement learning for the full-length games of StarCraft II, including hierarchical policy training with extracted macro-actions, reward design, and curriculum learning. We also provide an open-source implementation of the state-of-the-art work to be compared with our approach. With limited computation resources, we show that our approach achieves better results fighting against the built-in AIs. So far, StarCraft II is still the most complex game domain for RL research, and it deserves more attention from the RL community. Fast and efficient RL training on StarCraft II with less human knowledge is still a challenge for the RL research. In the future, we will continue to explore more efficient RL algorithms that can better solve problems in StarCraft.

\acks{We kindly thank all the reviewers of this paper for their suggestions and advice, which helped the paper be improved a lot.} This work was supported by Natural Science Foundation of China under Grant (61921006, 61832008).



\appendix

\begin{appendices}

\section{State Space} \label{append:State Space}
In this section, we present the design of the state space and the macro actions learned. The design of the reward function and specific training settings are given later. Finally, we introduce how to deal with problems encountered in the selection of building locations.

In our hierarchical architecture, the state features have two types: non-spatial features and spatial features. The list of the non-spatial state features is shown in Table~\ref{tab: scalar states}. The content of the spatial state features is shown in Table~\ref{tab: spatial states}. The controller only uses some of the non-spatial features as its global state. The base sub-policy uses all the non-spatial features as its local state. The battle sub-policy uses both the non-spatial and spatial features as its local state.

\begin{table*}[t]
	\centering
	\scalebox{1.0}{
		\begin{tabular}{ l | l }
			\hline
			features   &  remarks  \\
			\hline
			opponent.difficulty  & from 1 to 10 \\
			observation.game-loop & game time in frames  \\
			observation.player-common.minerals & minerals  \\
			observation.player-common.vespene & gas \\
			observation.score.score-details.spent-minerals & mineral cost \\
			observation.score.score-details.spent-vespene  & gas cost \\
			player-common.food-cap & max population \\
			player-common.food-used & used population \\
			player-common.food-army  & population of army \\
			player-common.food-workers  & population of workers \\
			player-common.army-count & counts of army \\
			food-army / food-workers & rate of army on workers \\
			num of army & the number of our army \\
			* num of probe, zealot, stalker, etc & multi-features \\
			* num of pylon, assimilator, gateway, cyber, etc &  multi-features \\
			* score\_cumulative & blizzard detailed score \\
			\hline
			* cost of pylon, assimilator, gateway, cyber, etc &  multi-features \\
			* num of probe which are doing building actions &  multi-features \\
			\hline
			* num of probe for mineral and the ideal num & multi-features \\
			* num of probe for gas and the ideal num & multi-features \\
			* num of the training probe, zealot, stalker &  multi-features \\
		\end{tabular}
		}
		\caption{Non-spatial Features}
		\label{tab: scalar states}	
\end{table*}

\begin{table*}[t]
	\centering
	\scalebox{1.0}{
		\begin{tabular}{ l | l }
			\hline
			features   &  post processing (map-width is set to 64)  \\
			\hline
			observation(minimap)[height] & reshape(-1, map-width, map-width) / 255 \\
			observation(minimap)[visibility] & reshape(-1, map-width, map-width) / 2 \\
			observation(minimap)[camera]& reshape(-1, map-width, map-width) \\
			observation(minimap)[relative]& reshape(-1, map-width, map-width) / 4 \\
			observation(minimap)[selected]& reshape(-1, map-width, map-width)\\
			
			observation(screen)[relative]& reshape(-1, map-width, map-width) / 4 \\
			observation(screen)[selected]& reshape(-1, map-width, map-width) \\
			observation(screen)[hitpoint-r]& reshape(-1, map-width, map-width) / 255 \\
			observation(screen)[shield-r]& reshape(-1, map-width, map-width) / 255 \\
			observation(screen)[density-a]& reshape(-1, map-width, map-width) / 255 \\
		\end{tabular}
		}
		\caption{Spatial Features}
		\label{tab: spatial states}
\end{table*}

\section{Macro Actions} \label{append:Macro Actions}
Here is how we generate macro actions. First, we let the experts play 30 games including difficult level-1 to level-3. In these games, we use the Protoss race against the Terran race and only use the combat units of Zealot and Stalker. After that, we saved the replays of the experts and analyzed these replays. The sequence of actions we dig through the PrefixSpan algorithm is as follows. The top 30 action sequences with the highest frequency of occurrence are listed in Table~\ref{tab: macro action}. It is worth noting that since all StarCraft 2 operations follow a form similar to English grammar, that is, the form of the subject plus verbs, the first action of all action sequences is necessarily the selection action. In addition, many actions are smart screen operations in the action sequence. This operation produces different effects depending on the target of execution. Therefore, it does not help build macro actions and can be discarded. When we have dropped some duplicate action sequences, we can construct a collection of macro actions. It should be noted that since macro actions are not codes, we need to ``translate" them into specific codes like Python. These codes need to be carefully written to ensure that the atomic actions in macro actions perform in the right order.

\begin{table*}[t]
	\centering
	\scalebox{0.75}{
		\begin{tabular}{ l  l  c }
			\hline
			action sequences   & frequency  \\
			\hline
			select-point(unit name: Protoss.Probe) $\to$ Harvest-Gather-Probe-screen & 2711 \\
			select-point(unit name: Protoss.Probe) $\to$ Smart-screen & 2253  \\
			select-point(unit name: Protoss.Nexus) $\to$ Train-Probe-quick & 1421  \\
			select-point(unit name: Protoss.Probe) $\to$ Build-Pylon-screen & 1298 \\
			select-point(unit name: Protoss.Nexus) $\to$ Smart-screen & 1030 \\
			select-point(unit name: Protoss.Nexus) $\to$ select-army & 968 \\
			select-point(unit name: Protoss.Probe) $\to$ Build-Gateway-screen & 814 \\
			select-point(unit name: Protoss.Gateway) $\to$ Train-Zealot-quick & 762 \\
			select-point(unit name: Protoss.Probe) $\to$ Build-Pylon-screen $\to$ Harvest-Gather-Probe-screen & 659 \\
			select-point(unit name: Protoss.Gateway) $\to$ Train-Stalker-quick & 621 \\
			select-point(unit name: Protoss.Nexus) $\to$ select-army $\to$ Attack-Attack-screen & 581 \\
			select-point(unit name: Protoss.Probe) $\to$ select-army & 574 \\
			select-point(unit name: Protoss.Nexus) $\to$ Train-Zealot-quick & 536 \\
			select-point(unit name: Protoss.Gateway) $\to$ select-army & 467 \\
			select-point(unit name: Protoss.Probe) $\to$ Build-Gateway-screen $\to$ Harvest-Gather-Probe-screen & 466 \\
			select-point(unit name: Protoss.Nexus) $\to$ Train-Stalker-quick & 325 \\
			select-point(unit name: Protoss.Gateway) $\to$ Smart-screen & 300 \\
			select-point(unit name: Protoss.Gateway) $\to$ select-army $\to$ Attack-Attack-screen & 284 \\
			select-point(unit name: Protoss.Probe) $\to$ Attack-Attack-screen & 260 \\
			select-point(unit name: Protoss.Probe) $\to$ Build-Pylon-screen $\to$ Smart-screen & 258 \\
			select-point(unit name: Protoss.Probe) $\to$ select-army $\to$ Attack-Attack-screen & 254 \\
			select-point(unit name: Protoss.Probe) $\to$ Build-Assimilator-screen & 253 \\
			select-point(unit name: Protoss.Probe) $\to$ Train-Zealot-quick & 243 \\
			select-point(unit name: Protoss.Gateway) $\to$ Cancel-BuildInProgress-quick & 209 \\
			select-point(unit name: Protoss.Probe) $\to$ Smart-screen $\to$ Build-Pylon-screen & 194 \\
			select-point(unit name: Protoss.Probe) $\to$ Build-CyberneticsCore-screen & 182 \\
			select-point(unit name: Protoss.Probe) $\to$ Smart-screen $\to$ select-army & 166 \\
			select-point(unit name: Protoss.Probe) $\to$ Smart-screen $\to$ Harvest-Gather-Probe-screen & 164 \\
			select-point(unit name: Protoss.Probe) $\to$ Smart-screen $\to$ Build-Gateway-screen & 147 \\
			select-point(unit name: Protoss.Probe) $\to$ Harvest-Gather-Probe-screen $\to$ select-army & 147 \\
		\end{tabular}
		}
		\caption{PrefixSpan Results (30 most frequently occurring)}
		\label{tab: macro action}
\end{table*}

\section{Network Structure} \label{append:Network Structure}
The network for the controller and sub-policies contains two parts: a policy net and a value net. The policy net has 1 input layer, 3 hidden layers, and 1 output layer and the number of units for each hidden layer is 128. The activation functions are Relu for all hidden layers. For the output layer, the activation function is Softmax. The value net contains 1 input layer, 3 hidden layers, and 1 output layer. Value net directly accepts the observation input, meaning that the policy net and value net have no intersecting paths. The number of units for each hidden layer is 128. The activation functions are Relu for all hidden layers. For the output layer, the number of units is 1 and has no activation function. The CNN layer is similar as in~\cite{mnih2015human}. The number of channels for each Conv2d layer is 32, 64, 64, and 3. The strides for the Conv2d layer are all 3. 

In the final 3-layer hierarchical architecture, the policy net and the value net share the same previous layers, which is that they use the same input layer and 3 hidden layers. The data flow from the final hidden layer will go to the output layer of the policy net and the output layer of the value net.

\section{PPO Settings} \label{append:PPO Settings}
We present the settings for PPO here. The $ \gamma $ value is $ 1 $. The $ \lambda $ in the generalized advantage estimation is set to $ 1 $. The clip value $ \epsilon $ in the PPO is $ 0.1 $. For the coefficient of the value network, the $c_1$ value is $ 0.01 $. For the coefficient of the entropy, the $c_2$ value is set to $ 10^{-5} $. The learning rate is set to $ 10^{-4} $. The batch size of PPO is $ 64 $. We run the batch 20 epochs in each update of PPO.

In the final 3-layer hierarchical architecture, most values of these parameters have been optimized. The episodes in one updating is $1000$. The $ \gamma $ value is $ 0.9995 $. The $ \lambda $ in the generalized advantage estimation is set to $ 0.9995 $. The clip value $ \epsilon $ in the PPO is $ 0.2 $. For the coefficient of the value network, the $c_1$ value is $ 0.5 $. For the coefficient of the entropy, the $c_2$ value is set to $ 10^{-3} $. The batch size of PPO is $ 512 $.

\section{Reward Functions} \label{append:Reward Functions}
The designed reward functions have two types. One is the instant reward, which can be got after every action. The other is the final reward, which can only be got at the end state. The controller and sub-policies use the same final reward: win/loss and a time penalty. For each sub-policy, the instant reward is different. In our version, instant rewards depend on the number of units and structures, e.g., the number of probes, pylons, and so on. The number collected statistically from expert replays for each unit and building is in Table~\ref{tab: reward weight}. The reward weight $ \alpha $ for calculation of $ (-\alpha * M) $ is set to 10. The reward weight $ \beta $ for multiplying the outcome (-1, 0, 1) is $50$.

\begin{table}[h]
    \centering
    \scalebox{1.0}{
    \begin{tabular}{c | c c c c | c c}
    \toprule
    Unit type & pylon & gas & gateway & cyber & probe & army \\
    \midrule
    Number  &   8 & 2 & 5 & 2 & 16 & 40   \\
    \bottomrule
    \end{tabular}
    }
    \caption{The collected value from expert replays.}
    \label{tab: reward weight}
\end{table}

We use experiments to show that the Blizzard score is not a good reward function in the preceding sections. If we only choose the part of the Blizzard scores, the agent can also learn a policy. Specifically, the base sub-policy can use the worker idle time, the production idle time, the total value unit, and the total value structures of the Blizzard scores as its instant reward. The battle sub-policy can use the killed value units $ kill\_unit $ and killed value structures $ kill\_struc $ of the Blizzard scores as its instant reward. If we use these scores with the hand-designed reward function, we need to give these scores a weight to balance, which is that $ (kill\_unit / \omega) $ and $ (kill\_struc / \rho)  $. The $ \omega $ and $ \rho $ are set to 100 and 50 respectively. The instant reward for the controller is the cumulative sum of the chosen policy's instant reward. 

In the final 3-layer hierarchical architecture, we use the win/loss (outcome) reward as the reward for RL training.

\begin{figure}[t]
	\begin{minipage}[t]{\linewidth}
		\centering
		\includegraphics[width=0.8\textwidth]{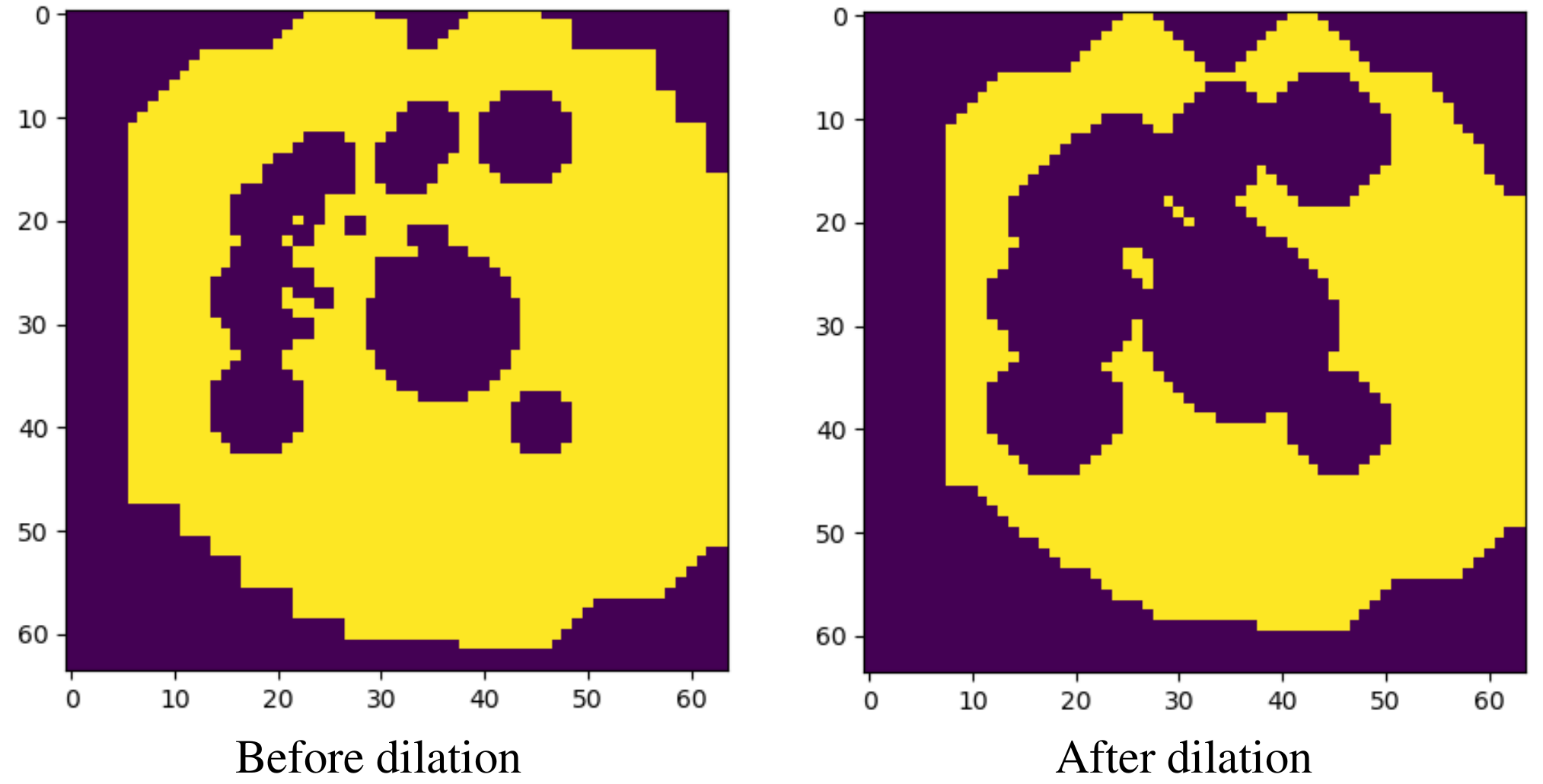}
		\caption{Effect of dilation algorithm. Purple is 1, and yellow is 0.}
		\label{fig:build}
	\end{minipage}
\end{figure}

\section{Selection of Building Locations} \label{append:Selection of Building Locations}
It is worth noting that the construction of a building needs to specify a location. There are three ways to get the location: learning, heuristic ways by human knowledge, and random method. The first two ways may be somewhat costly in applications, and most locations have little influence on the game results. Therefore, after some trials, we choose the random method. However, random selection may have a serious problem. As the name implies, random selection is randomly selecting a location on the game screen. If we specify any position without restriction, since all buildings and units have collision volumes, it will cause conflicts with the existing buildings or units. The solution we take is to read the feature map of the current screen, mark all the existing units and buildings, and the rest is the space where the new building can be built. Consider a 64x64 map, all pixels with units and buildings are labeled $1$, and the rest are labeled $0$, which are buildable are. This map is called $ M $. Direct use of $ M $ is not applicable. Because every building in SC2 has a size of roughly between 2 and 5 in pixels, when the position we specify is at the edge of the buildable area, the building may collide with other existing ones, causing the action to fail.

We use the dilation algorithm, common in image processing, to handle this problem. That is, instead of using the original $ M $ feature-map, we use the $ \operatorname*{dilat}(M) $ feature-map. Dilation reduces the area of the $0$ areas, but the success rate for constructing buildings is improved. In Fig.~\ref{fig:build}, the left figure is a feature-map before the dilation operator, and the right one is a feature-map after the dilation operator. It can be seen that the buildable area (yellow) becomes smaller after the dilation algorithm. Protoss is a special race that requires the supply of power. Therefore, when calculating the buildable area, the feature-map should be ``AND'' operated with the feature-map of the power mask. Note the design of the selection module of building location needs knowing some basic rules of the SC2, but it doesn't rely on much specific human knowledge.

\end{appendices}

\bibliography{reference}
\bibliographystyle{theapa}

\end{document}